\documentclass[a4paper,12pt]{article}

\usepackage{wrapfig}
\usepackage{etoc}
\usepackage[dvipsnames]{xcolor}
\usepackage[title]{appendix}
\usepackage{tikz}
\usetikzlibrary{arrows.meta, positioning, shapes.geometric, calc}
\usepackage{comment}
\usepackage[authoryear]{natbib}
\usepackage{amsmath}
\usepackage{amsfonts}
\usepackage{amsthm}
\usepackage{amssymb, bm}
\usepackage{enumerate}
\usepackage{setspace}
\usepackage[colorlinks=true, allcolors=blue]{hyperref}
\usepackage{algorithm}
\usepackage{algpseudocode}
\allowdisplaybreaks
\usepackage{graphicx}
\usepackage{booktabs}
\usepackage{subcaption}
\usepackage{xr-hyper}

\usepackage{nicematrix}
\NiceMatrixOptions{
code-for-first-row = \color{blue} ,
code-for-last-row = \color{blue} ,
code-for-first-col = \color{blue} ,
code-for-last-col = \color{blue}
}

\usepackage[customcolors]{hf-tikz}

\tikzset{style green/.style={
    set fill color=green!50!lime!60,
    set border color=white,
  },
  style cyan/.style={
    set fill color=cyan!90!blue!60,
    set border color=white,
  },
  style orange/.style={
    set fill color=orange!80!red!60,
    set border color=white,
  },
  hor/.style={
    above left offset={-0.15,0.31},
    below right offset={0.15,-0.125},
    #1
  },
  ver/.style={
    above left offset={-0.1,0.3},
    below right offset={0.15,-0.15},
    #1
  }
}

\newcommand{\anon}{1}

\begin{document}
\addtocontents{toc}{\protect\setcounter{tocdepth}{-1}}

\newcommand{\bb}[1]{\mathbb{#1}}
\newcommand{\op}[1]{\operatorname{#1}}
\newcommand\given[1][]{\:#1\vert\:}
\newtheorem{theorem}{Theorem}[section]
\newtheorem{proposition}{Proposition}[section]
\newtheorem{corollary}{Corollary}[section]
\newtheorem{lemma}{Lemma}[section]
\newtheorem{definition}{Definition}
\newtheorem{remark}{Remark}
\newtheorem{assumption}{Assumption}[section]

\newcommand{\nrm}[2]{\|#1\|_{#2}}
\newcommand{\lrnrm}[2]{\left\|#1\right\|_{#2}}
\newcommand{\lrp}[1]{\left(#1 \right)}
\newcommand{\brcs}[1]{\left\{#1 \right\}}
\newcommand{\bxs}[1]{\left[#1 \right]}
\newcommand{\vrt}[1]{\left|#1 \right|}

\newcommand{\ba}{\mathbf{a}}
\newcommand{\balp}{\bm\alpha}
\newcommand{\bdb}{\mathbf{b}}
\newcommand{\bc}{\mathbf{c}}
\newcommand{\bd}{\mathbf{d}}
\newcommand{\be}{\mathbf{e}}
\newcommand{\bdf}{\mathbf{f}}
\newcommand{\bg}{\mathbf{g}}
\newcommand{\bh}{\mathbf{h}}
\newcommand{\bi}{\mathbf{i}}
\newcommand{\bj}{\mathbf{j}}
\newcommand{\bk}{\mathbf{k}}
\newcommand{\bl}{\mathbf{l}}
\newcommand{\bmm}{\mathbf{m}}
\newcommand{\bn}{\mathbf{n}}
\newcommand{\bo}{\mathbf{o}}
\newcommand{\bp}{\mathbf{p}}
\newcommand{\bq}{\mathbf{q}}
\newcommand{\br}{\mathbf{r}}
\newcommand{\bs}{\mathbf{s}}
\newcommand{\bt}{\mathbf{t}}
\newcommand{\btht}{\bm{\tht}}
\newcommand{\bu}{\mathbf{u}}
\newcommand{\bv}{\mathbf{v}}
\newcommand{\bw}{\mathbf{w}}
\newcommand{\bx}{\mathbf{x}}
\newcommand{\by}{\mathbf{y}}
\newcommand{\bz}{\mathbf{z}}

\newcommand{\bA}{\mathbf{A}}
\newcommand{\bB}{\mathbf{B}}
\newcommand{\bC}{\mathbf{C}}
\newcommand{\bD}{\mathbf{D}}
\newcommand{\bE}{\mathbf{E}}
\newcommand{\bF}{\mathbf{F}}
\newcommand{\bG}{\mathbf{G}}
\newcommand{\bH}{\mathbf{H}}
\newcommand{\bI}{\mathbf{I}}
\newcommand{\bJ}{\mathbf{J}}
\newcommand{\bK}{\mathbf{K}}
\newcommand{\bL}{\mathbf{L}}
\newcommand{\bM}{\mathbf{M}}
\newcommand{\bN}{\mathbf{N}}
\newcommand{\bO}{\mathbf{O}}
\newcommand{\bP}{\mathbf{P}}
\newcommand{\bQ}{\mathbf{Q}}
\newcommand{\bR}{\mathbf{R}}
\newcommand{\bS}{\mathbf{S}}
\newcommand{\bT}{\mathbf{T}}
\newcommand{\bU}{\mathbf{U}}
\newcommand{\bV}{\mathbf{V}}
\newcommand{\bW}{\mathbf{W}}
\newcommand{\bX}{\mathbf{X}}
\newcommand{\bY}{\mathbf{Y}}
\newcommand{\bZ}{\mathbf{Z}}

\newcommand{\cA}{{\cal A}}
\newcommand{\cB}{{\cal B}}
\newcommand{\cC}{{\cal C}}
\newcommand{\cD}{{\cal D}}
\newcommand{\cE}{{\cal E}}
\newcommand{\bcE}{\mathbf{\cal E}}
\newcommand{\cF}{{\cal F}}
\newcommand{\cG}{{\cal G}}
\newcommand{\cH}{{\cal H}}
\newcommand{\cI}{{\cal I}}
\newcommand{\cK}{{\cal K}}
\newcommand{\cL}{{\cal L}}
\newcommand{\cT}{\mathcal{T}}
\newcommand{\cM}{{\cal M}}
\newcommand{\cN}{{\cal N}}
\newcommand{\cO}{{\cal O}}
\newcommand{\cP}{{\cal P}}
\newcommand{\cR}{{\cal R}}
\newcommand{\cs}{{\cal s}}
\newcommand{\cS}{{\cal S}}
\newcommand{\cU}{{\cal U}}
\newcommand{\cV}{{\cal V}}
\newcommand{\cW}{{\cal W}}
\newcommand{\cX}{{\cal X}}
\newcommand{\cY}{{\cal Y}}

\newcommand{\eps}{\epsilon}
\newcommand{\lmd}{\lambda}
\newcommand{\omg}{\omega}
\newcommand{\sgm}{\sigma}
\newcommand{\Sgm}{\Sigma}

\newcommand{\tht}{\theta}
\newcommand{\Tht}{\Theta}
\newcommand{\Gmm}{\Gamma}

\newcommand{\fm}{\mathfrak{m}}
\newcommand{\fn}{\mathfrak{(n)}}

\def\bon{\mathbf{1}}
\def\R{\mathbb{R}}
\def\pr{\mathbb{P}}
\def\b{\boldsymbol}
\def\h{\widehat}
\def\t{\tilde}
\def\E{\mathbb E}
\def\Var{\operatorname{Var}}
\def\off{\operatorname{off}}
\def\diag{\operatorname{diag}}
\def\tr{\operatorname{tr}}
\def\sk{^{\scriptscriptstyle(k)}}
\def\so{^{\scriptscriptstyle(1)}}
\def\st{^{\scriptstyle\star}}
\def\sK{^{\scriptscriptstyle(K)}}
\def\skt{^{\scriptscriptstyle\star(k)}}
\def\sKt{^{\scriptscriptstyle\star(K)}}
\def\sht{^{\scriptscriptstyle\rm ht}}
\def\shm{^{\scriptscriptstyle\rm hm}}

\def\dlt{\delta}
\def\Dlt{\Delta}
\def\dht{\dlt}
\def\hdlt{\h\dlt}
\def\dhm{\dlt_{1\fm}}
\def\hdhm{\h\dlt_{1\fm}}
\def\dst{\dlt\st}

\def\Ep{\bcE_s}
\def\Epm{\bcE_{s\fm}}
\def\Et{\bcE_t}
\def\Etm{\bcE_{t\fm}}

\def\Pp{\bP_s}
\def\Pt{\bP_t}
\def\Dt{\bD_t}
\def\Xt{\bX_t}
\def\xt{\bx_t}
\def\Zt{\bZ_t}
\def\hZt{\h\bZ_t}
\def\tZt{\t\bZ_t}
\def\zt{\bz_t}

\def\Dp{\bD_s}
\def\Xp{\bX_s}
\def\xp{\bx_s}
\def\Zp{\bZ_s}
\def\zp{\bz_s}
\def\tDt{\t\bD_t}

\def\Xip{\Xi_s}
\def\Xit{\Xi_t}

\def\Sgot{\Sgm_{x_t}}
\def\Sgxp{\Sgm_{\xi_s}}
\def\Sgxt{\Sgm_{\xi_t}}

\def\np{n_s}
\def\nt{n_t}
\def\post{p_1\st}

\def\sep{\sgm_\eps}
\def\set{\tau_\eps}
\def\sp{\sgm}
\def\sop{\sgm_x}
\def\stp{\sgm_z}
\def\sxp{\sgm_\xi}
\def\sot{\tau_x}
\def\stt{\tau_z}
\def\sxt{\tau_\xi}

\def\rp{\rho_s}
\def\rt{\rho_t}
\def\pp{\pi_s}
\def\pt{\pi_t}

\def\spacingset#1{\renewcommand{\baselinestretch}%
{#1}\small\normalsize} \spacingset{1}


\if1\anon
{
  \title{\bf Heterogeneous transfer learning for high-dimensional regression with feature mismatch}
  \author{Jae Ho Chang, Massimiliano Russo, and Subhadeep Paul\\
  \emph{Department of Statistics, The Ohio State University}}
  \date{}
  \maketitle
}
\fi

\if0\anon
{
  \bigskip
  \bigskip
  \bigskip
  \begin{center}
    {\LARGE\bf Heterogeneous transfer learning for high-dimensional regression with feature mismatch}
  \end{center}
  \medskip
}
\fi

\bigskip
\begin{abstract}
 We study Heterogeneous Transfer Learning (HTL) for high-dimensional regression with differing feature sets. Such feature mismatch arises when some variables available in a data-rich source domain are unavailable in a data-poor target domain. Yet most homogeneous TL methods require the same feature space in both the source and target domains, limiting their practical applicability. Conversely, existing HTL methods lack statistical error guarantees, limiting their utility for scientific discovery. We propose an HTL method that first learns a feature map between the missing and observed features leveraging the vast source data, imputes the unavailable features in the target, and then performs a two-step TL for penalized regression. We consider both the linear and the nonparametric feature maps. We develop upper bounds on the estimation and prediction errors of HTL, assuming that the source and target parameters differ sparsely, without requiring the target model itself to be sparse.  We also establish matching minimax lower bounds, showing that the proposed procedures achieve optimal rates.  Our results elucidate the effects of model complexity, sample size, the quality and differences in feature maps, and differences in the models across domains. We also derive minimax rates for the misspecified homogeneous TL model that discards unavailable features and show that our HTL procedure can attain a smaller error rate than homogeneous TL. We further extend the framework to multiple source domains and develop a negative-transfer defense that provably excludes adversarial sources from transfer with high probability.
\end{abstract}

\noindent%

\spacingset{1.8} 
\setdisplayskipstretch{1}


\section{Introduction}
The Transfer Learning (TL) problem concerns estimating a statistical model for a target population from which we have only limited data, using information from a source domain with abundant data. The model we wish to infer may have many more (even exponentially more) parameters in comparison to the sample size available in the target domain. Therefore, it may be infeasible to train the model using target data alone. On the other hand, we have a much \textit{larger dataset} from a source population, sufficient to train a complex statistical or machine learning model. The characteristics of the source domain are expected to be similar to those of the target population, but some parameters might differ between the populations. TL, therefore, aims to \textit{transfer} the knowledge from a \textit{pre-trained model} in the source population to improve estimation and prediction on the target population. 

In practical applications of TL, several additional challenges arise, including differing sets of features or covariates between target and source data. For example, in a data-poor new environment, researchers may be \textit{limited} not only by sample size but also by the \textit{features} on which they can collect data.
Several methods have been proposed for transfer learning in the last two decades (see \cite{zhuang2020comprehensive,weiss2016survey} for comprehensive reviews). However, most TL methods are developed for the \textit{homogeneous problem} (HmTL), in which the target and source domains share the same feature space, thereby limiting their practical applicability \citep{day2017survey}. 
As surveyed in \citet{day2017survey}, heterogeneous transfer learning (HTL) methods have been developed to bridge this gap. 
However, these HTL methods typically lack statistical guarantees on estimation and prediction errors, limiting their utility for scientific discovery. In this article, we develop an HTL method for high-dimensional linear models for improved estimation and prediction in the target population. We then study the statistical properties of the method and show that our method is minimax optimal for this problem. 

Recently, several authors have studied the problem of TL in various statistical problems, including high-dimensional linear models \citep{bastani2021predicting,li2022transfer,takada2020transfer}, generalized linear models (GLMs) \citep{li2023estimation,tian2022transfer}, graphical models \citep{li2023transfer}, nonparametric classification and regression problems with differential privacy \citep{auddy2024minimax,cai2024transfer} and regression under covariate shift \citep{he2024transfusion}. In the homogeneous linear model setting, i.e., when the same set of covariates is available in both source and target data, \cite{bastani2021predicting,takada2020transfer} proposed an estimation strategy based on estimating a sparse discrepancy between source and target parameters with $\ell_1$ penalized regression. 
Later, \citet{li2022transfer} extended this idea to multiple sources, obtained minimax-optimal rates of estimation error, and showed that their ``translasso'' algorithm achieves the optimal error rate. 

A key limitation of the above approaches is that they assume that the same set of covariates or features is available in both the target and source populations. However, this assumption is unrealistic in several applied settings. For example, in biomedical studies, the target domain could be a smaller dataset from a specific patient population measuring a few key covariates, while the source population could be obtained from Electronic Health Records (EHRs), where more information is available. The problem of learning with heterogeneous feature spaces in the source datasets has been explored in the literature \citep{wiens2014study,yoon2018radialgan}, but these methods with different features lack statistical error guarantees. Our work bridges this gap in the literature. 

We develop an HTL method for learning a target high-dimensional linear regression model with both matched and unmatched features, where the unmatched features are observed in the source domain but unavailable in the target domain.  We further assume that the missing features are related to the observed features, allowing us to estimate their values in the target data using a \textit{feature map} learned from the vast source data. We model the feature map both as a linear map as well as \textit{nonparametrically as unknown functions} to be estimated from data. 
We consider the challenging scenario where both matched and \textit{missing features are high-dimensional} relative to the target sample size. The source and target regression coefficients are allowed to differ, but their difference is assumed to be sparse. Importantly, unlike \cite{li2022transfer} and \cite{tian2022transfer}, we \textit{do not require the target coefficient vector to be sparse}. Instead, we assume that we have a vast amount of data available in the source domain, where data-intensive model fitting (``pre-training'') is feasible. This setup allows a complex model to be pretrained on a large source sample and then calibrated on a much smaller target sample. The problem setup we study is depicted succinctly in the schematic diagram in Figure \ref{fig:htl_schematic}. 

\begin{figure}[h]
\centering
\begin{tikzpicture}[
arrow/.style={-Stealth, thick, color=green!60!black},
]
\node[rectangle, draw, thick, minimum width=5cm, minimum height=5cm, rounded corners,
fill=blue!10, draw=blue!60] (source) at (-1.2,0) {};
\node[above=0.1cm of source.north, font=\bfseries\large] {Source Domain};
\node[below=0.2cm of source.north, font=\small] {$n_s = 10{,}000$};
\foreach \x in {-3.2, -2.8,-2.4, -2, -1.6, -1.2, -0.8, -0.4, 0, 0.4} {
\draw[fill=blue!30, draw=blue!60] (\x, -1.6) rectangle (\x+0.3, 1.7);
}
\node[below, font=\scriptsize] at (-1.2, -1.6) {$\mathbf{D}_s = (X_1, \ldots, X_{p_1}, Z_1, \ldots, Z_{p_2})$};
\node[below, font=\scriptsize] at (-1.2, -2.0) {$p=p_1+p_2$ features available};
\node[rectangle, draw, thick, minimum width=3.5cm, minimum height=4cm, rounded corners,
fill=orange!10, draw=orange!60] (target) at (6,0) {};
\node[above=0.1cm of target.north, font=\bfseries\large] {Target Domain};
\node[below=0.2cm of target.north, font=\small] {$n_t = 200$};
\foreach \x in { -0.4, 0, 0.4} {
\draw[fill=orange!30, draw=orange!60] (6+\x, -0.8) rectangle (6+\x+0.3, 0.9);
}
\node[below, font=\scriptsize] at (6, -1) {$\Xt = (X_1, \ldots, X_{p_1})$};
\node[below, font=\scriptsize] at (6, -1.4) {$p_1$ features available};
\node[rectangle, draw, fill=blue!20, thick, minimum width=2.5cm, minimum height=0.6cm, rounded corners]
at (-1.2, -3.0) {$\omega$};
\node[rectangle, draw, fill=orange!40, thick, minimum width=2.5cm, minimum height=0.6cm, rounded corners]
at (6, -2.5) {$\beta$};
\draw[arrow, line width=2pt] (1.9, 0.5) -- node[above, font=\bfseries] {Transfer} (3.8, 0.5);
\end{tikzpicture}
\caption{\textbf{HTL challenges:} (1) $n_t \ll p$ (limited target data), (2) $\omega \neq \beta$ (target and source models differ), and (3) $p_2$ features are missing in the target domain.}
\label{fig:htl_schematic}
\end{figure}

We derive high-probability upper bounds for estimation and prediction error under both linear and nonparametric feature maps.  Our results suggest that the estimation error is related to the target and source domain sample sizes, the model's complexity, the strength and cross-domain differences of the feature map, and the discrepancy between the target and source model parameters. The results indicate that the estimation error increases with larger parameter differences between the two domains and with a larger number of model parameters, while it decreases with increasing sample size in the target domain. Further, the estimation error increases as the association between matched and mismatched features decreases. We allow the feature maps to differ somewhat between the source and target domains. However, intuitively, the maps cannot differ much since we do not observe the missing features in the target domain and hence cannot calibrate the feature map to the new domain. The estimation error increases as the discrepancy between the feature maps increases. The results also indicate that it is possible to estimate models with an \textit{exponentially large} number of parameters compared to the \textit{target sample size} provided the source sample size is sufficiently large. Since \textit{we do not require the target model to be sparse}, the HTL procedure represents substantial benefits over training a model with target data only. We further derive matching minimax lower bounds within a model subclass, showing that our methods achieve the \textit{minimax-optimal} rates under the assumed model class.
We then extend this method to the case when multiple source datasets are available. For this multi-source setting, we develop a \textit{negative transfer defense} strategy that \textit{provably excludes} the adversarial or harmful sources with high probability. 

To quantify the value of modeling feature mismatch, we analyze the misspecified model obtained by discarding the unmatched features and applying a homogeneous TL (HmTL) method of \citet{bastani2021predicting} and \citet{li2022transfer} to the common covariates. We also derive minimax rates for this reduced model class and show that the error achieved by our HTL procedure (which operates in a richer model class) can be smaller than the minimax lower bound in both the linear and nonparametric feature-map cases. This comparison formalizes the cost of ignoring unavailable features and establishes HTL's superiority over HmTL in the presence of feature mismatch. Our simulations support the theoretical conclusions and show improved estimation and prediction compared with HmTL and the target-only lasso.

Two papers related to ours are the recent works by \cite{zhao2023heterogeneous} and \cite{zhang2024concert}, but the problem settings differ.
While \cite{zhao2023heterogeneous} focuses on the case when ``summary-level" information from an external study is available in addition to individual-level data from a main study, our setup poses individual-level data availability in both domains, allowing us to introduce and learn a \textit{feature map}. Importantly, unlike \cite{zhao2023heterogeneous}, we do not assume that parameters corresponding to common features are identical across the source and target domains; instead, we assume only that their difference is sparse. 
On the other hand, \cite{zhang2024concert} considers the possibility that some features in a source dataset are relevant to the target domain problem and transferable, while others are not. 

However, our framework differs from this work in that we assume some required covariates are \textit{unavailable} in the target domain. In contrast, in the framework of \cite{zhang2024concert}, all covariates are available in all domains (as for HmTL), but the effect of some covariates is different across domains, making them \textit{non-transferable}. In particular, \cite{zhang2024concert} 
needs to assume that transferable and non-transferable features are orthogonal, as their method requires that adding non-transferable covariates to the model does not change the coefficients associated with transferable covariates. This assumption is difficult to check in practice and might be unrealistic in many settings. In contrast, we provide an end-to-end solution that includes estimation in the source domain, giving practitioners tools to transfer knowledge when some covariates are unavailable. Our approach to imputing missing covariates is also related to the problem of integrating block-wise missing data from multiple sources \citep{xue2021integrating}. A key difference is that \cite{xue2021integrating} assumes that
the coefficients of the linear model relating $Y$ to $\bX$ are the same across all sources; we allow sparse differences between the target and source domains.

\section{Heterogeneous transfer learning method}\label{sec:HTL}
We first consider transferring knowledge from a single-source dataset and will later generalize to the case of multiple source datasets. Accordingly, suppose we have a single-source study that is informative for a predictive task in the target domain.
Let $\by=(Y_1,\dots,Y_n)'\in\R^n$ denote a response vector. Denote matched features by $\bx\in\R^{p_1}$, assumed to be common to both the source and target populations, and mismatched features by $\bz\in\R^{p_2}$, which are not observed in the target domain. We assume that $\bx$ and $\bz$ are random vectors (i.e., random designs).
We write $\bX\in\R^{n\times{p_1}}$ and $\bZ\in\R^{n\times{p_2}}$ to denote matrices whose rows are independent copies of $\bx$ and $\bz$, respectively. Then, our complete design matrix becomes $\bD=\bxs{\bX,\,\bZ}$.
From now on, let us use subscripts $s$ and $t$ to indicate the source and target domains, respectively.
Then, consider the following data-generating processes for a statistical unit from the target domain $\brcs{\xt,\zt,\by_t}$ and source domain $\brcs{\xp,\zp,\by_s}$, respectively:
\begin{align}
    Y_{t} = {\xt}'\beta\st_1 + {\zt}'\beta\st_2 + \eps_t,\qquad
    Y_{s} = {\xp}'\omg\st_1 + {\zp}'\omg\st_2 + \eps_s,
    \label{eq:dgp}
\end{align}
for zero-mean sub-Gaussian errors  $\eps_t$ and $\eps_s$ (see Definition 2 in Appendix \ref{sec:design}).
We observe $\nt$ sample points from the target domain and $\np$ from the source domain.
Our objective is to leverage information from the source data, which is transferred via the regression coefficients $\omg_1\st,\omg_2\st$, to enhance our estimator for the parameters in the data-generating process of the target data ($\beta\st_1, \beta\st_2$).
Only the covariates $\bx$ are accessible in the target domain, whereas the covariates $\bz$ are not. However, 
both $\bx$ and $\bz$
are observed in the source population.

We posit that the differences in covariate effects are sparse in the sense that $\dlt_l\st = \beta\st_l - \omg\st_l$ is a sparse vector for $l=1,2$, either in the $\ell_0$ or $\ell_1$ sense.
Our focus in this paper is both an improved estimation of target domain parameters $(\beta\st_1,\beta\st_2)$, and improved prediction accuracy in the target domain. We emphasize that the true data-generating process in the target population requires both $\bx$ and $\bz$, but only $\bx$ is available to us in the training samples and any future prediction tasks. Therefore, when comparing the methods' prediction errors, we assume the test data is generated according to the target data's data-generation process, yet only $\bx$ will be available in the test data. Note that we do not assume that the regression model parameters, for either the matched or the mismatched cases, are the same across domains. In the subsequent methods and theory sections, we allow both matched and mismatched features to be high-dimensional. A typical high-dimensional asymptotic setup in which our results hold is $p_1 \asymp p_2$, $ p \gg n_t$, and $\np \gg \nt$. We do not make the restrictive assumption of sparsity in the target domain. This is because, intuitively, through transfer learning, we can ``transfer'' information contained in the vast sample $\np$. Provided $\np$ is sufficiently large, this allows for the number of parameters to be exponentially larger than the sample size in the target model without sparsity assumptions. 
We use standard notation for vector and matrix norms, and standard definitions of sub-Gaussian random variables, vectors, and ensembles. A complete list of notations and definitions is in Appendix \ref{sec:design}.

Our methodology involves estimating the missing covariate matrix $\bZ$ in the target data with the help of a feature map learned from the source data. 
Specifically, we assume the existence of relationships on the source and target populations as follows:
\[\zp=\bh_s(\xp)+\xi_s,\quad \zt=\bh_t(\bx_t)+\xi_t,\]
for some function class $\bh:\R^{p_1}\to\R^{p_2}$ to be specified later and an independent random perturbation $\xi\in\R^{p_2}$. We will call the functions $\bh_s(\bx):=\E(\zp|\bx),\bh_t(\bx):=\E(\zt|\bx)$ feature maps in source and target domains, respectively. We will let the \emph{feature maps between the domains differ}, and the \emph{extent of their differences} will be part of our theoretical upper bounds on error rates. The error terms $\xi_s,\xi_t$ are $\sxp,\sxt$-sub-Gaussian random vectors, respectively.
If $\np,\nt$ independent observations are available for the source and target domains, respectively. i.e., $(\bx_{is},Y_{is})$ for $i=1,\dots,n_s$ and $(\bx_{jt},Y_{jt})$ for $j=1,\dots,\nt$,
then the relationship between feature sets will be written as $\Zp=\bH_s(\Xp)+\Xi_s$, and $\Zt=\bH_t(\Xt)+\Xi_t.$
Also, the source sample and target sample are assumed to be independent.
\begin{assumption}\label{asm:dsgn}
The minimum eigenvalues of $\Var(\xp)=:\Sgm_{x_s}$,$\Var(\xt)=:\Sgot$ and $\Var(\xi_s)=:\Sgxp,$$\Var(\xi_t)=:\Sgxt$ are bounded below by some positive absolute constant, and the maximum eigenvalues of $\Var(\eps_s)=:\Sgm_{\eps_s},\Sgm_{x_s}$, and $\Sgm_{\xi_s}$ are bounded above by some positive absolute constant.
\end{assumption}

\begin{remark}
    \textbf{Strength of feature map:} Assumption \ref{asm:dsgn} rules out the possibility of perfectly predicting the missing features from the matched features. However, there is no assumption on the minimum strength of the feature map. The sub-Gaussian variance proxies $\sxp$ and $\sxt$ control the \emph{strength of the association in the feature map}.  
\end{remark}

\subsection{Feature maps}
\subsubsection{Linear feature map}\label{sec:lfm}
First, we consider the case when $\bh$ is a linear map of $\bx\in\R^{p_1}$ to $\bz\in\R^{p_2}$. i.e., $\bh(\bx)=\bP'\bx$ for some fixed matrix $\bP\in\R^{{p_1}\times{p_2}}$.
We specify the following linear process for both the target and source data: $
    \Zp = \Xp\Pp + \Xip,$ and $ 
    \Zt = \Xt\Pt + \Xit,$
where $ \Xp\sim SG(\sop)\ \bot\ (\Xip)_{n\times {p_2}} \sim SG(\sxp)$ and $ \Xt\sim SG(\sot)\ \bot\ (\Xit)_{n\times {p_2}} \sim SG(\sxt)$. Therefore, the matrices $\bX$ and $ \Xi$ are two independent sub-Gaussian ensembles. This assumption is equivalent to assuming that a linear map of matched covariates can approximately recover the mismatched covariates.  
Then, we have the covariance matrix $\Sgm$ of a single (complete) feature vector $\bd=(\bx',\bz')'$ as
$\Sgm:=\Var(\bd)=\begin{bmatrix}
    \Sgm_{x} & \Sgm_{x}\bP \\
	& \bP'\Sgm_{x}\bP+\Sgm_\xi
\end{bmatrix}$, 
by construction.
We denote the maximum singular value of $\bP$ by $\rho:=\sgm_{\max}(\bP)$.
We consider the feature map to be sparse and denote the number of nonzero elements in the source feature map by $ s_\bP:= \sum_{i=1}^{p_1}\sum_{j=1}^{p_2}\bb I((\Pp)_{ij}\ne0)$, and its maximum column sparsity by $m_\bP:=\max_{j\le p_2}\sum_{i=1}^{p_1}\bb I((\Pp)_{ij}\ne0)$. 
\begin{assumption}\label{asm:lfm-sp}
    There exists a constant $C_\bP>0$ such that $\max_{(i,j)}|(\Pp)_{ij}|\le C_\bP$
    .
\end{assumption}
The above assumption that the feature map magnitudes are bounded uniformly implies $\nrm{\bP_s}{1}=\cO(m_\bP)$ (which we will assume to be $\cO(1)$ in the theoretical results). 
The utility of our proposal depends on the discrepancy between the feature mappings, $\Dlt_\bP:=\Pp-\Pt$. As mentioned before, we allow the feature maps to differ across domains, and $\nrm{\Dlt_\bP}{}$ \emph{measures the extent of their differences}. Write the maximum singular values of $\Pp,\Pt\in\R^{{p_1}\times {p_2}}$ as $\rp$, $\rt$ respectively.
Since $\Xp$ and $\Xip$ are assumed to be independent, we have that $\Zp\sim SG(\sgm_z)$, where $\sgm_z:=\sgm_x\rho_s+\sgm_\xi$, by the convenient results we prove in \ref{lem:sg-add} in Appendix E. Therefore, $\Zp$ is also a sub-Gaussian ensemble.
We have the random design matrix as $\Dp=\bxs{\bX,\,\bZ}\in\R^{n\times p}$, and hence $\Dp\sim SG\lrp{\sgm_x+\sgm_z}$.
Then, we can denote $\sp:=\sop+\stp=\sop(1+\rp)+\sxp$ (and $\tau:=\sot+\stt=\sot(1+\rt)+\sxt$ for the target domain).

\subsubsection{Nonparametric modeling of feature map}\label{sec:nlfm}
We now relax the linearity assumption and allow the feature map to be nonparametric. Let the matched features $\bx\in\R^{{p_1}}$, which are common for both domains, be drawn from the uniform distribution on $[-a,a]^{p_1}:=\cD_x$ for some fixed $a>0$.
For the mismatched features $\bz \in \mathbb{R}^{p_2}$, we consider the nonparametric regression model
$z_j = h_j(\bx) + \xi_j,$ $j=1,\dots,p_2,$  $h_j(\bx):=\E(z_j|\bx),$
where $\xi\in\mathbb{R}^{p_2}$ is a $\sgm_\xi$-sub-Gaussian perturbation with $\E(\xi)=\b0_{p_2}$, independent of $\bx$ and~$\eps$.
We write $\bz = \bh(\bx)+\xi$ with $\bh(\bx):=\E(\bz|\bx)\in\mathbb{R}^{p_2}$. 
Therefore, we assume the mismatched features are noisy versions of some \emph{unknown functions} of the matched features.
Stacking $n$ independent copies of $\bx$ row-wise, $\bX$ denotes an $n\times{p_1}$ design matrix and $\bH(\bX):=[\E(\bz_1|\bx_1),\dots,\E(\bz_n|\bx_n)]'\in\R^{n\times{p_2}}$ contains their feature maps.

Our construction of non-linear feature mappings is motivated by \cite{zhang2023regression}, with technical details relegated to Section \ref{nlfm_app} of the Appendix.  
Briefly, we put the following assumptions and Assumption \ref{asm:senergy} on the class of functions for $\bh$, denoted by $\cH$. 

\begin{assumption}[active feature set]\label{asm:active-dimension}
    Let $\cD\st_x:=[-a,a]^{\post}$ for some fixed $\post\le{p_1}$. There exists a $\post$-variate vector function $\bh\st:\cD_x\st\to\R^{p_2}$, and a set of indices $S_1\subset[{p_1}]$ with $|S_1|=\post$ such that for any $\bu\in\cD_x$, we have $\bh(\bu)=\bh\st(\bu_{S_1})$.
\end{assumption}
This assumption states that only a subset of the coordinates of $\bx$ has a meaningful relationship with $\bz$.
Therefore, we can focus on estimating $\bh$ over $\post$ active covariates, rather than the full $p_1$-dimensional input, thereby avoiding the curse of dimensionality in nonparametric regression. In the following assumption, ${\rm D}^\ba f$ denotes $\frac{\partial^{\nrm{\ba}{1}}}{\partial x_1^{a_1}\dots\partial x_d^{a_d}}f$ for $f:\R^d\to\R$.
\begin{assumption}[Feature map class]\label{asm:smooth-h}  
    For $\cI\st:=\brcs{\ba\in\bb N^{\post};\nrm{\ba}{\infty}\le1}$, the active feature map $\bh\st$ satisfies a dominating mixed smoothness 1, i.e., 
    $\max_{j\le p_2}\sum_{\ba\in\cI\st}\|{\rm D}^\ba h_j\st\|_{L_2(\cD_x\st)}^2<\infty$
\end{assumption}

Under our assumptions, $\bh$ can be written as $\bh(\bx)=\b\psi_\infty(\bx)'\btht_\infty$ where $\b\psi_\infty\in\R^\infty(\bx)$ is a series expansion of $\bx$ and $\btht_\infty\in\R^\infty$ the corresponding coefficients. We use the compact notation $\bH(\bX)=\Psi_\infty(\bX)\Tht_\infty$ where $\Psi_\infty$'s rows are $\psi_\infty(\bx_1),...,\psi_\infty(\bx_n)$ and $\Tht_\infty\in\R^{\infty\times p_2}$ collects corresponding coefficients mapping $\bx$ to $\bz$. If the infinite-order basis expansion is truncated at $m=M$, then, for the source domain, 
$\Psi_M(\Xp):=[\b\psi_M(\bx_{1,s}),\dots,\b\psi_M(\bx_{\np,s})]'\in\R^{\np\times M},$ contains the unraveled basis functions evaluated at each data point $\bx_{i,s}$.

\begin{remark}
    \textbf{Identifiability with linear feature map:} 
    The parameter vectors $\beta\st$ and $\omega\st$ in the original data-generating models remain identifiable because $\bz$ is not an exact linear function of $\bx$. The model parameters are not identifiable from the reduced model $
        Y_{t} = {\xt}'\beta\st_1 + {\xt}'\bP_{t}\beta\st_2 + \xi_{t} '\beta\st_2 +  \eps_t,$
    with access to $\xt$ and ${\bP_{t}}'{\xt}$. However, with access to the full set of covariates, $\xt$ and $\zt$, the original model parameters are identifiable. 
\end{remark}

\subsection{Two-stage HTL Estimation of $\beta\st$}
Now, we propose heterogeneous transfer learning (HTL), which begins by imputing mismatched features in the source domain and then employs a two-stage estimation approach.
Imputation of missing features can be conducted either with a penalized linear map estimation or a penalized sieve estimation for a nonparametric feature map.
The latter is useful to address unknown nonlinear associations between feature sets $\bx$ and $\bz$.
Specifically, under the assumption of an unknown non-linear relationship among features, we employ the following approximation chain: $\Psi_\infty(\Xt)\Tht_{t,\infty}\approx\Psi_M(\Xt)\Tht_{t,M}\approx\Psi_M(\Xt)\Tht_{s,M},$
where $\Tht_{t,M},\Tht_{s,M}\in\R^{M\times{p_2}}$ are matrices of Fourier coefficients for the target and source domains, respectively, truncated at order $m=M$.
The truncation $M \equiv M(n_s, p)$ is a deterministic sequence scaling with the data dimension to balance the sieve approximation error.

The quality of the first approximation is determined by the sieve approximation error in the target domain, which can be made arbitrarily small by choosing $M$ sufficiently large under standard regularity conditions. The second approximation depends on the discrepancy between the finite-order feature representations in the target and source domains that we cannot control. The impact of this discrepancy on HTL's performance will be quantified in the theoretical analysis. We denote the spectral norm of this discrepancy by
\(
\nrm{\Dlt_\Tht}{}:=\nrm{\Tht_{t,M}-\Tht_{s,M}}{}
\)
(an analogue in the linear feature mapping case$\nrm{\Dlt_\bP}{}=\nrm{\Pt-\Pp}{}$).

The sieve method would approach the non-parametric regression problem by truncating an infinite polynomial expansion with basis functions to estimate each coordinate $h_j$ for $j\in[p_2]$.
First, we truncate the expansion series \eqref{eq:tht-exp} at $m=M$, i.e., obtain $\h h_{s,j}:=\sum_{m=1}^M\h\tht_{mj,s}\psi_m(\xp)$.
We solve $\ell_1$-penalized sieve estimator \citep{zhang2023regression} ${p_2}$ times to obtain an estimator $\h\Tht_{s,M}:=[\h\btht_s^1,...,\h\btht_s^{p_2}]\in\R^{M\times{p_2}}$ on the source domain where
\[\h\Tht_{s, M}:={\arg\min}_{\Tht=[\btht^1,...,\btht^{p_2}]\in\R^{M\times{p_2}}}\brcs{\frac{1}{\np}\nrm{\bZ_s-\Psi_M(\Xp)\Tht}{F}^2+\sum_{j=1}^{p_2}\gamma_j\nrm{\btht^j}{1}}\]
for tuning parameters $\gamma_j>0$ and $\bZ_{s,j}$ the $j$-th column of $\Zp$.

When we specify the feature mapping to be linear, we can instead obtain the estimator $\h\Pp$ by solving the penalized multi-response linear regression problem
\[\h\Pp:={\arg\min}_{\bP\in\R^{{p_1}\times{p_2}}}\brcs{\frac{1}{\np}\nrm{\Zp-\Xp\bP}{F}^2+\sum_{j=1}^{p_2}\gamma_j\nrm{\bp_{s,j}}{1}},\]
which is equivalent to running $p_2$ lasso regressions.
A two-stage estimation follows this feature map estimation step.
\begin{enumerate}[\textbf{Step} \bf I.]
    \item Imputation: for $\Psi_M(\Xt):=[\b\psi(\bx_{1t}),\dots,\b\psi(\bx_{\nt,t})]'\in\R^{\nt\times M}$,
    \begin{align*}
        \h\bZ_t=\begin{cases}
            \Xt\h\Pp,&\text{ when }\bh\text{ is linear} \\
            \Psi_M(\Xt)\h\Tht_{s,M},&\text{ when }\bh\text{ is non-linear}
        \end{cases}
    \end{align*}
    \item Estimation:
    \begin{enumerate}[1)]
        \item $\h\omg=\arg\min_{\omg\in\R^p}\frac{1}{n_{s}}\lrnrm{\by_{s}-\Xp\omg_1-\Zp\omg_2}{}^2$
        \item $\h\beta = \arg\min_{\beta\in\R^p}\brcs{\frac{1}{\nt}\nrm{\by_{t}-\Xt\beta_1-\h\bZ_t{\beta}_{2}}{}^2 + \lmd\nrm{\beta - \h\omg}{1}}$
    \end{enumerate}
\end{enumerate}
In the first stage, which we call the imputation stage, we use the estimated feature map to approximate $\Zt$ with the target data $\Xt$ in the linear case and with the expansion of the target data $\Psi_M(\Xt)$ in the nonparametric case. Note that this expansion of the nonparametric feature map is learned by transferring knowledge about the truncation parameter $M$ from the source domain. Therefore, we transfer both $M$ and $\Tht_{s,M}$ from the source domain.

We assume the source domain sample size is vast. The multi-response lasso in the linear case and penalized sieve estimation for the nonparametric case are the most data-intensive tasks in the source domain. Under our assumptions on the sparsity of the feature maps, these procedures require sample sizes of at least $O(p)$, since they are multivariate procedures that require imputation of $p_2$ vectors. We remind readers that we assume the missingness is also high-dimensional, such that $ p_1 \asymp p_2 \asymp p$. Therefore, in the estimation stage, we use OLS to estimate $\h{\omega}$ since the sample size in the source domain is large enough to enable OLS.
However, we note that we allow for high-dimensional target model coefficients without any sparsity assumption, since the target sample size is expected to be much smaller than $p$. 

Throughout the paper, we refer to the above two-stage procedure as \emph{heterogeneous transfer learning} (HTL). In simulations and the case study, we write \texttt{HTL.LN} for HTL with a linear feature map ($\h\Pp$) and \texttt{HTL.NL} for HTL with a sieve-based nonlinear map ($\h\Tht_{s,M}$); the homogeneous baseline of Section \ref{sec:HmTL} is denoted \texttt{HmTL}.

\subsection{Baseline Homogeneous Transfer Learning} \label{sec:HmTL}
We can compare this new method with the baseline approach of homogeneous transfer learning on the \emph{misspecified model} using only matched features $\bX$ (e.g., \citealp{bastani2021predicting,li2022transfer}).
This is equivalent to ignoring the mismatched features (and their associations with the matched ones) during the modeling process, thereby transferring knowledge only for estimating $\beta_1\st$.
The joint estimation method using our notation is,
\begin{align*}
    \h\omg_{1\fm}&={\arg\min}_{\omg_1\in\R^{p_1}}\frac{1}{n_{s}}\lrnrm{\by_{s}-\Xp\omg_1}{}^2 \\
    \h\beta_{1\fm}&={\arg\min}_{\beta_1\in\R^{p_1}}\brcs{\frac{1}{\nt}\nrm{\by_{t}-\Xt\beta_1}{}^2 + \lmd\nrm{\beta_1 - \h\omg_{1\fm}}{1}}
\end{align*}

However, under our true data-generating model that includes both $\bX$ and $\bZ$, this method will fit a misspecified model. Let $\Et := (\eps_{1,t}, \ldots, \eps_{\nt,t})' \in \R^{\nt}$ denote the correctly specified target model's error vector. 
The baseline homogeneous transfer learning (HmTL) estimator under a linear feature map targets the misspecified model's parameter $\beta_{1\fm}\st := \beta_1\st + \Pt\beta_2\st$, and not the true parameter $\beta_1\st$. The same marginalized parameter is targeted by target domain-only penalized regression such as lasso. Hence, both methods incur heavy estimation error for estimating $\beta_1\st$ due to this bias and provide no estimate for $\beta_2\st$. However, even this misspecified model may have predictive utility; hence, we will use it as a baseline against which to compare our HTL method, both theoretically and through simulation. To assess prediction accuracy, we need to understand how the marginalized parameter $\beta_{1\fm}\st$ is estimated. We note that this parameter target admits the decomposition $\beta_{1\fm}\st = \t\omg_{1\fm}\st + \dlt_{1\fm}\st$, where $\dlt_{1\fm}\st := \dlt_1\st + \Pt\dlt_2\st$ and $\t\omg_{1\fm}\st := \omg_1\st + \Pt\omg_2\st$. The HmTL estimator, by construction, also decomposes into $\h\beta_{1\fm} = \h\omg_{1\fm} + \hdhm$, where $\h\omg_{1\fm}$ estimates the source marginal parameter $\omg_{1\fm}\st := \omg_1\st + \Pp\omg_2\st$, and the correction term $\hdhm$ approximately optimizes the misspecified objective: $
\frac{1}{\nt}\lrnrm{\Etm + \Xt(\dlt_{1\fm}\st - \dlt_{1\fm} + \Dlt_\bP\omg_2\st)}{}^2 + \lmd_{1\fm}\nrm{\dlt_{1\fm}}{1}$,
where $\Etm := \Et + \Xit\beta_2\st$ denotes the composite error vector, which is higher due to omitted variable. 

\begin{remark}[Predictive utility of HTL with linear feature map]
The proposed HTL objective trades the omitted variable error for imputation error. HmTL misspecifies the target model as $Y_t = {\xt}'\beta^*_{1\fm}+  \eps_{t\fm},$ while the true target domain model with the feature map is $Y_{t} = {\xt}'\beta\st_1 + \bh_t({\xt})'\beta\st_2 + \xi_{t} '\beta\st_2 +  \eps_t$. 
In general, HTL can provide more accurate predictions since this misspecified model cannot learn the true signal $\bh_t(\xt)'\beta_2\st$.
In the case of a linear feature map, $\hZt=\Xt\h\Pp$ lies in the column space of $\Xt$, which might seem to offer no additional predictive power beyond $\h\beta_{1\fm}$.
However, the target-side imputation of HTL is statistically necessary to transfer the source knowledge for $\beta_2\st$ into the target domain.
Our procedure transfers the knowledge for estimating $\beta_1\st, \beta_2\st, \Pt$ separately, as opposed to transferring information for estimating $\beta_{1\fm}\st$ marginally.  This loss of information from not using $Z$ in the source domain, despite having access to it, is an important reason for HTL's outperformance over homogeneous TL in prediction error. A second reason of HmTL's underperformance is that even though $\delta_1^*$ is a sparse vector, that does not automatically imply $\delta_{1m}^*$ will also be a sparse vector.  We analyze in detail how this difference can lead to better predictions for HTL in section \ref{sec:theory}. In the case of a non-linear feature map, in addition to lower estimation error for $\beta_1\st$, we expect HTL to have superior predictive performance, since the misspecified model with $\bX$ cannot compensate for the effect of missing $\bZ$.
\end{remark}

\section{Theoretical properties of the HTL estimator} \label{sec:theory}
Before stating our main results, we clarify the notion of prediction error in our setup. 
We assess out-of-sample prediction error in the target domain considering
$n_t^\fn$ new observations from the target domain. We remind the reader that the data-generating process for this new data from the target domain is $Y_t^\fn={\bd_t^\fn}\,'\beta\st+\eps_t^\fn$ while only $\xt^\fn$ is available to regress on.
In general, systematic error $\eps_t^\fn$ is difficult to reduce, and the signal $\bd_t^\fn\,'\beta\st$ contains an imputation noise $\xi_t^\fn\,'\beta_2\st$.
Regardless of the choice of model for feature mapping, these components are not learnable from the available data.
Instead, we analyze the prediction errors of the signals that we can learn from the data: $\E(Y_t^\fn|\xt^\fn)=\E(\bd_t^\fn|\xt^\fn)'\beta\st$.

\subsection{Estimation under linear feature map}
Our main result in the linear feature map case provides, under certain assumptions, a high-probability bound on the estimation and prediction errors of the HTL estimator.

\begin{theorem}[Non-asymptotic HTL error rate under linear feature map]\label{thm:lfm-sp-err}
    Assume that $\max(\rp,\sop,\sxp,\sep,\sot) = \cO(1)$, $\log p\lesssim\nt$, and $\np\gtrsim\rp^2p+s_\bP\log{p_1}$. Further assume that
    $m_\bP^2\nrm{\dst}{1}\sqrt{\frac{\log p_1}{\nt}}=\cO\bxs{\set+\nrm{\beta_2\st}{}(\nrm{\Dlt_\bP}{}+\sxt)}.$
    Then, under Assumption \ref{asm:re-mp} with $K=1$, with probability at least $1-c_1\exp(-c_1'\log p)-c_2{p_2}\exp(-c_2'\log{p_1})$,
    \begin{align*}
        \nrm{\h\beta-\beta\st}{}^2&\lesssim\frac{p}{\np}\bxs{1\vee\frac{\frac{m_\bP^2p}{\np} \lrp{\rp+\sqrt{\np^{-1}s_\bP\log{p_1}}}^{-2}}{\brcs{\set+\nrm{\beta_2\st}{}\lrp{\nrm{\Dlt_\bP}{}+\sxt}}^2}} \\
        &\qquad+\nrm{\dst}{1}\bxs{\nrm{\dst}{1}\wedge\lrp{1+\sqrt{\np^{-1}s_\bP\log{p_1}}}\brcs{\set+\nrm{\beta_2\st}{}\lrp{\nrm{\Dlt_\bP}{}+\sxt}}\sqrt{\frac{\log p_1}{\nt}}}.
    \end{align*}
    For $\tDt^\fn:=\Xt^\fn[\bI_{p_1},\h\Pp]$, with probability at least $1-c\exp(-c'n_t^\fn)$ we have the out-of-sample prediction error of
    \begin{align*}
        \frac{1}{n_t^\fn}\lrnrm{\E(\by_t^\fn|\Xt^\fn)-\tDt^\fn\h\beta}{}^2\lesssim\nrm{\Dlt_\bP}{}^2\nrm{\beta_2\st}{}^2+\nrm{\h\beta-\beta\st}{}^2.
    \end{align*}
\end{theorem}
The following result is a direct corollary of the preceding theorem, established under additional structural assumptions introduced to streamline the analysis and elucidate the roles of the parameters of primary interest.
\begin{corollary}\label{cor:lfm-sp-err}
    Under the assumptions in Theorem \ref{thm:lfm-sp-err}, further assume that $\np\gg(\rp^2+m_\bP^2)p+s_\bP\log{p_1}$.
    Then, with the same probability bounds,
    \begin{align}
        \nrm{\h\beta-\beta\st}{}^2&\lesssim
        \frac{p}{\np} + \nrm{\dst}{1}\bxs{\nrm{\dst}{1}\wedge\brcs{\set+\nrm{\beta_2\st}{}\lrp{\nrm{\Dlt_\bP}{}+\sxt}}\sqrt{\frac{\log p_1}{\nt}}}
        \label{eq:lfm-sp-err}
    \end{align}
\end{corollary}

The theorem and the corollary, in particular, bring out the dependence of the upper bound on key quantities associated with the model. First, we note that the estimation error increases with increasing $\nrm{\dst}{1}$, indicating that as the \emph{domains differ more}, the effectiveness of HTL will decrease. Second, $\sxt$ is the sub-Gaussian variance proxy of the error term in the feature map, which can be thought of as a \emph{measure of the (inverse) strength of association}. Therefore, the larger the value of $\sxt$, the weaker the association strength in the feature map. The estimation error increases with $\sxt$, indicating a weaker feature map leads to performance deterioration. Third, as stated earlier, we allow the feature maps to differ between domains, and the extent of the difference is quantified by the spectral norm of the feature map matrices, $\nrm{\Dlt_\bP}{}$. Since we do not observe any data points with both $\bX$ and $\bZ$ in the target domain, we cannot calibrate the feature map to the target domain. Therefore, the \emph{discrepancy between the feature maps} appears in our error bound. The greater this discrepancy, the higher the HTL estimation error.

We note from the above corollary that the error of estimating $\beta\st$ converges to 0 (or equivalently, the parameter estimation loss with high probability converges to 0) as long as $(\rp^2 + m_\bP) p\ll\np$, $s_\bP\log{p_1}\ll\np$, and
$\nrm{\dst}{1}^2\brcs{\set^2+\nrm{\beta_2\st}{}^2(\nrm{\Dlt_\bP}{}^2+\sxt^2)}\log p_1\ll\nt.$
Therefore, we can have an even \emph{exponentially larger number of parameters }in the target model than the sample size and still estimate the parameters consistently. Therefore, the key assumptions we need to make are that the source database has a large enough sample, that the difference between the two domains is sparse, and that the coefficients corresponding to the unobserved features in the target domain are small. We make no assumptions about the sparsity of the matched or observed features in the target domain. Therefore, the successful transfer of information enables estimation of a much larger and more complex model than the available sample size in the target population, by borrowing information from a vast source dataset.

\begin{remark}
    \textbf{Minimax optimal:} The rate in Corollary \ref{cor:lfm-sp-err} above matches the minimax lower bound in Theorem \ref{thm:htl-lb} up to constants and ignoring the domain shift in the feature map. Therefore, our HTL procedure is minimax optimal within our problem setup.
\end{remark}
\begin{remark}
    \textbf{Comparison with results in the literature:} 
    The first part of the estimation error rate is driven by a combination of OLS estimation error and feature map imputation error. It scales down as the source sample size $n_s$ increases. 
    As for the assumption on $n_s$, we remind the reader that for sparse feature map matrix  $\bP_s$, both the quantities $\rho_s$ and $m_\bP$ can be $\cO(1)$, while $s_\bP$ which controls the sparsity of the matrix is $\cO(p)$, since there are $p_2 \asymp p$ missing features. Therefore the condition boils down to $n_s \gg p$. The second part of the estimation error is similar to the rate in \cite{li2022transfer} and can be compared to their error rate (in our notation) of $\nrm{\dst}{1} \lrp{\nrm{\dst}{1} \wedge\sqrt{{\log p_1}/{\nt}}}$. Compared to the second term in \cite{li2022transfer}'s minimum, we have a multiplication by terms related to the imputation of missing features. We also do not have the additional term inside the minimum ${\nrm{\beta\st}{0}\log p}/{\nt}$  because we do not assume sparsity in the target model and lose the additional downside protection of the lasso rate.
    The prediction error has two components. The second component is the estimation error of the parameter $\beta\st$.
    The first term $\nrm{\Dlt_\bP}{}^2\nrm{\beta_2\st}{}^2$ arises from the imputation of unobserved features. While \cite{bastani2021predicting} did not analyze prediction error under fixed designs, our results can be compared with those of \cite{li2022transfer}, who studied prediction error under a random design similar to ours. The second term is identical to the result in \cite{li2022transfer}, while the first term is an additional additive term due to the imputation of missing features. The result follows the intuition as  we expected higher prediction errors in our setup due to the lack of access to certain covariates.
\end{remark}

For the baseline homogeneous transfer learning strategy, we derive high-probability bounds on the estimation and prediction errors. The HmTL algorithm is mathematically forced to target the marginalized parameter $\beta_{1\fm}\st$ due to fitting a model with only matched features. Our bound explicitly captures this structural misspecification inherent in homogeneous transfer learning.  Hence, the following error bound is qualitatively different from the results in \cite{bastani2021predicting,li2022transfer} and is specific to the case of feature mismatch.

In Theorem \ref{thm:lfm-sp-hm-err}, we obtain two estimation error bounds for estimating the parameters $\beta\st_1$ and $\beta_{1\fm}\st$, which correspond to the coefficient of $\bX$, in the true model and in the misspecified model, respectively.  The error rate for estimating $\beta\st_1$ is relevant for parameter estimation of the true data-generating model and includes the error due to omitted variable bias. The error rate for $\beta_{1\fm}\st$ is related to the prediction error, since it is the parameter of the best misspecified model and serves as a fair baseline for comparing the HTL's predictive performance. We have the following corollary under the simplified settings.
\begin{corollary}\label{cor:lfm-sp-err-hm}
    Assume that $\np\gg{p_1}$. Then (1) In Theorem \ref{thm:lfm-sp-hm-err}-1., with the same probability bounds,
    \begin{align}
        \nrm{\h\beta_{1\fm}-\beta_1\st}{}^2&\lesssim\nrm{\omg_2\st}{}^2\lrp{1\vee\frac{\nrm{\omg_2\st}{}^2}{\brcs{\set+(\sxt+\rt)\nrm{\beta_2\st}{}}^2}}\nonumber\\
        &\qquad+\nrm{\dlt_1\st}{1}\bxs{\nrm{\dlt_1\st}{1}\wedge\brcs{\set+(\sxt+\rt)\nrm{\beta_2\st}{}}\sqrt{\frac{\log{p_1}}{\nt}}}. 
        \label{eq:lfm-sp-hm-err}
    \end{align}
    (2) In Theorem \ref{thm:lfm-sp-hm-err}-2., further assume that $\nrm{\omg_2\st}{}\vee1=\cO\lrp{\nrm{\Sgot\Dlt_\bP\omg_2\st}{\infty}+\set+\sxt\nrm{\beta_2\st}{}+\nrm{\Dlt_\bP\omg_2\st}{}}.$ Then, with the same probability bounds,
    \begin{align*}
        \nrm{\h\beta_{1\fm}-\beta_{1\fm}\st}{}^2& \lesssim \lrp{\nrm{\omg_2\st}{}\vee1}^2\frac{{p_1}}{\np}\\&
        +\nrm{\dlt\st_{1\fm}}{1}\bxs{\nrm{\dlt\st_{1\fm}}{1}\wedge\brcs{\nrm{\Sgot\Dlt_\bP\omg_2\st}{\infty}+\lrp{\set+\sxt\nrm{\beta_2\st}{}+\nrm{\Dlt_\bP\omg_2\st}{}}\sqrt{\frac{\log{p_1}}{\nt}}}}
    \end{align*}
    and $\pr\Bigl(\frac{1}{n_t^\fn}\lrnrm{\E(\by_t^\fn|\Xt^\fn)-\Xt^\fn\h\beta_{1\fm}}{}^2\lesssim\rt^2\nrm{\beta_2\st}{}^2+\nrm{\h\beta_{1\fm}-\beta_{1\fm}\st}{}^2\Bigr)>1-c\exp(-cn_t^\fn).$
\end{corollary}
The key quantities to compare the HTL estimator $\h\beta$ and the homogeneous TL estimator $\h\beta_{1\fm}$ in terms of prediction error rates are $\omg_2\st,\dst,\dlt_{1\fm}\st,$ and $\Dlt_\bP$. 
\begin{remark}
\label{delta1m}
    Importantly, even though we assume the true model differs sparsely, i.e., $\dlt\st$ is small, the marginalized bias term $\dlt_{1\fm}\st$ need not be small and can exceed it, depending on $\Pt$. To see this, for $\dst\ne\b0_p$ and $\Pt^+:=[\bI_{p_1},\Pt]$ we can examine the inequality $$\nrm{\dst}{1}\le\nrm{\dst_{1\fm}}{1},~~i.e.,~~1\le\frac{\nrm{\Pt^+\dst}{1}}{\nrm{\dst}{1}}.$$
    This holds whenever the columns of $\Pt$ have $\ell_1$-norms greater than one.
\end{remark}
As stated in Remark  \ref{delta1m}, $\dlt_{1\fm}\st$ maybe much larger than the sparse discrepancy $\dlt\st$ making the target debiasing rate higher for HmTL compared to HTL. Moreover, the source estimation error rate for $\h\beta_{1\fm}$ is inflated by a factor of $\nrm{\omg_2\st}{}^2$ compared to $\h\beta$. More critically, the presence of the irreducible bias term $\nrm{\Sgot\Dlt_\bP\omg_2\st}{\infty}$ forces the estimation error rate to remain large regardless of the target sample size $\nt$. This structural bottleneck prevents the prediction error from improving even when target data is abundant. For heavy domain shifts (i.e., large $\nrm{\Dlt_\bP}{}$), the predictions of HTL estimator $\h\beta$ can also degrade due to amplified estimation variance, bounded by $\nrm{\Dlt_\bP}{}^2\nrm{\beta_2\st}{}^2$; however, this error decreases as $\nt$ increases. Therefore, even if the feature map shift $\Dlt_\bP$ is relatively small, the $\h\beta_{1\fm}$'s prediction significantly degrades compared to the HTL estimator $\h\beta$ due to this persistent bias induced by the marginalized estimation. 

The estimator
$\h\beta_{1\fm}$ also has a larger error than $\h\beta$ when estimating the true parameter $\beta_1\st$. In particular, unless the influence of the mismatched source features is negligible (specifically $\nrm{\omg_2\st}{} = o(1)$), the bound in \eqref{eq:lfm-sp-hm-err} can be larger than that in \eqref{eq:lfm-sp-err}. Theorem \ref{thm:lfm-sp-hm-err}-2 characterizes the prediction error, which scales in the omitted variable bias $\nrm{\omg_2\st}{}\vee1$ and $\rt\nrm{\beta_2\st}{}$. This additional error component is absent in \cite{li2022transfer}, as $\omg_2\st=\beta_2\st=(0)$ in their setting.

\subsection{Estimation under an unknown map $\bh\in\cH$}
The next theorem and corollary establish high-probability bounds on the estimation and prediction error when the feature map is non-parametric, with $\bz$ a nonlinear mapping of $\bx$.

\begin{theorem}\label{thm:sp-err}
    Assume that $\pp+\sxp + \sep = \cO(1)$, $\np\gtrsim \pp^2p$, $M\asymp\sqrt{{p_2}^3(\log{p_2})^{2\post+1}}$, and $\log p\lesssim\nt$. 
    Further assume that
    $\nrm{\dst}{1}\sqrt{\frac{\log p}{\nt}}=\cO\bxs{\set+(\sxt+\nrm{\Dlt_\Tht}{}+\lmd_{R_\Tht})\nrm{\beta_2\st}{}}.$
    Then, with probability at least $1-c\exp(-c'\log p)-c\log^{-1/2}p_2$,
    \begin{align*}
        \nrm{\h\beta-\beta\st}{}^2&\lesssim~
        \frac{p}{\np}\bxs{1\vee\frac{p/n_s}{\brcs{\set+(\sxt+\nrm{\Dlt_\Tht}{}+\lmd_{R_\Tht})\nrm{\beta_2\st}{}}^2}} \\
        &\qquad+\nrm{\dst}{1}\bxs{\nrm{\dst}{1}\wedge\brcs{\set+(\sxt+\nrm{\Dlt_\Tht}{}+\lmd_{R_\Tht})\nrm{\beta_2\st}{}}\sqrt{\frac{\log p}{\nt}}}.
    \end{align*}
    Further assume that we observe $n_t^\fn<\infty$ new observations from the target domain, i.e., $\Dt^\fn:=[\Xt^\fn,\,\Zt^\fn]$. Then, for $\tDt^\fn:=[\Xt^\fn,\,\Psi_M(\Xt^\fn)\h\Tht_{s,M}]$, with probability at least $1-c\exp(-c'n_t^\fn)-c\log^{-1/2}p_2$,
    \begin{align}\label{eq:sp-err2}
        \frac{1}{n_t^\fn}\lrnrm{\E(\by_t^\fn|\Xt^\fn)-\tDt^\fn\h\beta}{}^2\lesssim\nrm{\Dlt_\Tht}{}^2\nrm{\beta_2\st}{}^2+\nrm{\h\beta-\beta\st}{}^2.
    \end{align}
\end{theorem}
\begin{corollary}\label{cor:sp-err}
    Further assume that $\np\gg \pp^2p$.
    Then, with the same probability bounds,
    \begin{align}
        \nrm{\h\beta-\beta\st}{}^2&\lesssim \frac{p}{\np}+\nrm{\dst}{1}\bxs{\nrm{\dst}{1}\wedge\brcs{\set+(\sxt+\nrm{\Dlt_\Tht}{})\nrm{\beta_2\st}{}}\sqrt{\frac{\log p}{\nt}}}.
        \label{eq:sp-err}
    \end{align}
\end{corollary}
Compared to linear feature mapping, Sieve estimation of the source feature map converges at the slower rate of $\lmd_{R_\Tht}:=\lrp{{p_2}^{3/4}\log^{3/2}{p_2}\frac{\log^{\post-1}\np}{{\np}}}^{2/3}$.
The error of estimating $\beta\st$ converges to 0 as long as $\pp^2p \ll\np$ and
$\nrm{\dst}{1}^2\brcs{\set^2+\nrm{\beta_2\st}{}^2(\nrm{\Dlt_\Tht}{}^2+\sxt^2)}\log  p\ll\nt.$
There are technical distinctions between the linear mapping scale $\rp=\nrm{\Pp}{}$ and its non-parametric counterpart $\pp:= \|\Tht_{s,\infty}\|$. 
Since the linear mapping spans a low-dimensional subspace of the broader multivariate Sobolev space $\mathcal{H}$, the non-parametric parameter space naturally has significantly greater capacity, typically yielding $\pi \gg \rho$. 
This establishes a statistical trade-off: the sieve approximation offers flexibility to mitigate non-linear model misspecification, but it comes at the cost of the source domain sample size.

Despite this, the proposed framework continues to accommodate target models with exponentially larger parameters than the target domain sample size.  As in the linear case, the upper bound is influenced by the difference in feature maps (through the parameter $\nrm{\Dlt_\Tht}{}$), the strength of association between observed and missing features, and the sparsity of parameter differences across domains. Since the upper bound in Corollary \ref{cor:sp-err} matches the minimax lower bound in Theorem \ref{thm:htl-lb} with the condition on $\np$, our HTL procedure with a nonparametric feature map is \textit{minimax optimal}.

Next, we derive the following bounds for homogeneous transfer learning under the assumption of a non-linear feature map. Here, for theoretical convenience in deriving the results under a misspecified model, we assume the missing features $\bZ$ are linearly independent of the matched features $\bX$, but are nonlinearly related. In particular, we assume that there exists no $\bP\in\R^{{p_1}\times{p_2}}$
 such that $\bH(\bX)=\bX\bP$  in both target and source domain. This is reasonable since if they were linearly related, then we could use the linear feature map itself. The full theorem is stated in Appendix \ref{sec:prf-thm-err} as Theorem \ref{thm:sp-hm-err}.
\begin{corollary}
\label{cor:sp-hm-err}
    In addition to the conditions in Theorem \ref{thm:sp-hm-err}, assume that $\np\gg{p_1}$.
    Then, with the same probability bounds,
    \begin{align}\label{eq:sp-hm-err}
        \nrm{\h\beta_{1\fm}-\beta_1\st}{}^2 \lesssim&~ \nrm{\omg_2\st}{}^2 \bxs{1\vee\frac{\nrm{\omg_2\st}{}^2}{\brcs{\set+\nrm{\beta_2\st}{}(\sxt+\pt)}^2}}\nonumber\\
        &+\nrm{\dlt_1\st}{1}\bxs{\nrm{\dlt_1\st}{1}\wedge\brcs{\set+(\sxt+\pt)\nrm{\beta_2\st}{}}\sqrt{\frac{\log{p_1}}{\nt}}}.\nonumber
    \end{align}
    and $
        \frac{1}{n_t^\fn}\lrnrm{\E(\by_t^\fn|\Xt^\fn)-\Xt^\fn\h\beta_{1\fm}}{}^2
        \lesssim\pt^2\nrm{\beta_2\st}{}^2 + \nrm{\h\beta_{1\fm}-\beta_1\st}{}^2.$
\end{corollary}

Under a nonlinear feature relationship, HmTL exhibits error rates that scale with $\pt=\nrm{\Tht_{t,\infty}}{}$ for estimation and with $\pt^2$ for prediction. Corollary \ref{cor:sp-hm-err} yields a higher estimation error bound than the corresponding result for HTL in Corollary \ref{cor:sp-err}. The first term does not converge to 0 unless the influence of the mismatched source features satisfies $\nrm{\omg_2\st}{}^2 = o(1)$. Meanwhile, our proposed HTL effectively reduces the OVB by transferring $\h\omg_2$.
In addition, the second term now scales with the spectral energy of the target feature map, whereas the HTL result depends on $\nrm{\Dlt_\Tht}{}$.
Further, unlike the linear case, HmTL's prediction error is much higher due to $\pt$ being larger than $\rho_t$. The misspecified model can no longer effectively approximate the effect of the missing features; hence, HmTL severely underperforms HTL.

\subsection{Minimax Optimality}
The high-probability upper bounds in the previous section share a common structure. There is a \emph{source} error rate that decays with $n_s$ and a \emph{debiasing} rate governed by the contrast $\dst$ and the target sample size $n_t$. We now show that these error rates match the minimax lower bound and hence the procedures are minimax optimal. 
The source lower bounds are established under both the linear map $\bh(\bx)=\bP'\bx$ (Section \ref{sec:lfm}) and the sieve-based nonlinear map $\bh(\bx)=\Tht_\infty'\b\psi_\infty(\bx)$ (Section \ref{sec:nlfm}); the debiasing floor is stated in terms of the composite-noise scale and applies to both specifications. We prove minimax lower bounds under the following class of models that assumes the feature map remains the same across domains (Hence $\Dlt_P$ and $\Dlt_\Tht$ are 0).

\begin{definition}[Distribution class]\label{def:htl-class}
    Let $\Pi\sht$ denote the collection of all joint laws of the source sample $\Bigl\{(\xp\sk,\zp\sk,Y_s\sk)\Bigr\}_{k=1}^K$ and the target sample $\brcs{(\xt,Y_t)}$ generated under either the linear feature map $\bh(\bx)=\bP'\bx$ (Section \ref{sec:lfm}) or the sieve-based nonlinear map $\bh(\bx)=\Tht_\infty'\b\psi_\infty(\bx)$ (Section \ref{sec:nlfm}), satisfying Assumption \ref{asm:dsgn} with $\max(\sop,\sxp,\sep,\sot)=\cO(1)$.
    The homogeneous class is a subclass $\Pi\shm\subseteq\Pi\sht$, and is the same collection of laws: under $\bh(\bx)=\bP'\bx$ we index them by the marginalized target parameters $\beta_{1\fm}\st$ and $\dlt_{1\fm}\st$, and under $\bh(\bx)=\Tht_\infty'\b\psi_\infty(\bx)$ by the matched coefficients $\beta_1\st$ and $\dlt_1\st$.
\end{definition}
 Define the composite noise in the marginalized target regression model as $\eps_{t\fm}=\eps_t+\xi_t'\beta_2\st$. Write $\tau_{\eps_\fm}:=\set+\sxt\nrm{\beta_2\st}{}$ to denote its variance proxy parameter. For convenience of display we use the notation $\zeta(h):=h\cdot\min\lrp{h,\tau_{\eps_\fm}\sqrt{{\log p_1}/{\nt}}}$.
\begin{theorem}[Minimax lower bound for HTL]\label{thm:htl-lb}
    Assume that $\nrm{\dlt\st}{1}\tau_{\eps_\fm}\sqrt{{\log p_1}/{n_t}}\lesssim(\nrm{\dlt\st}{1}^2\vee1)$.
    Then, there exist an absolute constant $c>0$ (depending only on the design constants in Assumption \ref{asm:dsgn}) such that
    \begin{equation}\label{eq:htl-lb}
        \inf_{\h\beta}\sup_{\cP\in\Pi\sht}\cP\bxs{\nrm{\h\beta-\beta\st}{}^2\ge c\brcs{\frac{p_1}{\np} \vee \zeta(\nrm{\dst}{1})}}\ge\frac12,
    \end{equation}
\end{theorem}
\begin{theorem}[Homogeneous TL]\label{thm:hmtl-lb}
    There exists some absolute constant $c>0$ such that the following hold.
    \begin{enumerate}
        \item Suppose $\bh(\bx)=\bP'\bx$ and assume that $\nrm{\dlt_{1\fm}\st}{1}\tau_{\eps_\fm}\sqrt{\frac{\log{p_1}}{n_t}}\lesssim\nrm{\dlt_{1\fm}\st}{1}^2\vee1$. Then,
        \[\inf_{\h\beta_{1\fm}}\,\sup_{\cP\in\Pi\shm}\cP\bxs{\nrm{\h\beta_{1\fm}-\beta_{1\fm}\st}{}^2\ge c\brcs{\nrm{\omg_2\st}{}^2\frac{{p_1}}{\np}\vee \zeta(\nrm{\dlt_{1\fm}\st}{1})}}\ge\frac12.\]
        \item Suppose $\bh(\bx)=\Tht_\infty'\b\psi_\infty(\bx)$ and assume that $\nrm{\dlt_1\st}{1}\tau_{\eps_\fm}\sqrt{\log{p_1}/\nt}\lesssim\bxs{\nrm{\dlt_1\st}{1}^2\vee1}$. Then,
        \[\inf_{\h\beta_{1\fm}}\,\sup_{\cP\in\Pi\sht}\cP\bxs{\nrm{\h\beta_{1\fm}-\beta_1\st}{}^2\ge c\brcs{\nrm{\omg_2\st}{}^2\frac{{p_1}}{\np}\vee \zeta(\nrm{\dlt_1\st}{1})}}\ge\frac12\]
    \end{enumerate}
\end{theorem}
The minimax rates for HTL provided over Theorem \ref{thm:htl-lb} and Corollaries \ref{cor:lfm-sp-err}, \ref{cor:sp-err} show that the rate is lower than the minimax lower bound in Theorem \ref{thm:hmtl-lb} for the class of models homogeneous TL is restricted to by a factor of $\nrm{\omg_2\st}{}^2$ in the source error rate. This clearly establishes the superiority of HTL over HmTL for both parameter estimation and prediction. As expected, the minimax lower bounds for homogeneous TL vary by feature map type. In the case of the nonparametric feature map, there are no marginal parameters because we assume there is no linear association between $\bx$ and $\bz$; the relationship is nonlinear. Therefore, we state the lower bound in terms of the true parameter $\beta_1\st$. 

\section{Multiple source domains and negative transfer}\label{sec:ntd}
In a multi-source environment where information is aggregated across $K$ distinct source domains, transfer learning faces a fundamental challenge of performance degradation due to uninformative or adversarial sources, a phenomenon widely recognized as negative transfer. 
The extension of our single-source HTL framework to multi-source settings involves pooling of source coefficients and feature maps, and the comprehensive formulation is deferred to Appendix \ref{sec:mp}. 
This section addresses the negative transfer defense and introduces a Signal-based Adaptive Negative transfer Defense (SAND) method, formalized in Algorithm \ref{alg:sand} in the Appendix. Throughout the subsequent development, we operate under a linear feature map, but note that the algorithm extends naturally to non-linear feature map. 
We establish rigorous finite-sample control of false positives that will filter out adversarial sources.

For any candidate estimator $\bdb\in\R^{p_1}$, define its out-of-sample prediction risk in the target domain (where only $\bx$ is available) by
\begin{align}\label{eq:risk-fn}
    \cR(\bdb)
    :=\E\bxs{\lrp{Y_t - {\xt}'\bdb}^2}
    =\Var(\eps_{t\fm}) + \nrm{\Sgot^{1/2}\lrp{\bdb - \beta_{1\fm}\st}}{}^2
\end{align}
Hence $\cR(\beta_{1\fm}\st)=\Var(\eps_{t\fm})=\Var(\eps_t)+\nrm{\Sgxt^{1/2}\beta_2\st}{}^2$, is the oracle risk that establishes the fundamental lower bound for any target-only estimation strategy.
Because the oracle marginalized parameter $\beta_{1\fm}\st$ is inaccessible, we establish an empirical benchmark based on the cross-validated predictive risk of the target-only Lasso model.

Following Algorithm \ref{alg:sand} in Appendix, partition the target index set into $\kappa$ folds $\brcs{\cV_j}_{j=1}^\kappa$. 
Letting $\t{\beta}_{1\fm}^{(j)}(\lmd)$ denote the Lasso estimator fitted on the subset $\{i : i \notin \mathcal{V}_j\}$ with penalty parameter $\lambda$, the achievable baseline risk at a selected penalty $\lmd$ is defined as:
$    \bar{\cR}(\lmd) := \frac{1}{\kappa} \sum_{j=1}^\kappa \cR\left[ \t{\beta}_{1\fm}^{(j)}(\lmd) \right] \ge \cR(\beta_{1\fm}\st).$
The primary objective of the SAND procedure is to exclude source domains that may degrade predictive performance relative to this achievable baseline. We can compare the in-sample residual $\t{S}\sk:= \frac{1}{\nt}\lrnrm{\by_t- \Xt\h{\omg}_1\sk - \h\bH_s\sk(\Xt)\h{\omg}_2\sk}{}^2$ for each candidate source $k=1, \dots, K$ against the out-of-sample target excess risk $\bar{\cR}(\lmd_{\text{1SE}})$.
Since the risk $\bar\cR$ is unknown, the SAND selection rule instead includes $k$ in the active source set $\h{A}$ whenever the out-of-sample target residual sum $\bar R$ exceeds $\t{S}\sk$ by a prespecified safety threshold $\t{\eta} > 0$.

Defining the marginalized source parameter as $\omg_{1\fm}\skt := \omg_1\skt + \Pp\sk\omg_2\skt$, the conditional expectation $\mathbb{E}(Y_t | \xt) = \xt'\beta_{1\fm}\st$ implies the decomposition:
\begin{equation} \label{eq:excess-risk}
\beta_{1\fm}\st - \omg_{1\fm}\skt = \delta_1\skt + \Pp\sk\delta_2\skt - \Dlt_\bP\sk\beta_2\st,
\end{equation}
where $\Dlt_\bP\sk := \Pp\sk - \Pt$ captures the structural feature map shift. The risk of the source domain parameter $\mathcal{R}(\omg_{1\fm}\skt) = \text{Var}(\eps_{t\fm}) + \|\Sgm_{x_t}^{1/2}(\omg_{1\fm}\skt - \beta_{1\fm}\st)\|^2$ thus measures both the outcome model discrepancy through $(\delta_1\skt, \delta_2\skt)$ and the feature space mismatch through $\Dlt_\bP\sk$.

To ensure that the excess risk of an informative source is detectable, we posit that there exists a fixed tolerance that defines an informative source set
\begin{equation}\label{eq:A-true}
    A_{\eta_0}
    :=
    \brcs{k\in[K];\,\cR(\omg_{1\fm}\skt) \le \bar\cR(\lmd_{\rm1SE}) + \eta_0 }.
\end{equation}
A source $k\in A_{\eta_0}$ predicts the target response, in the population, at least as well as the fitted target-only baseline up to tolerance $\eta_0$.
Sources in $A_{\eta_0}^c$ (i.e., a set of $k$'s such that $\cR(\omg_{1\fm}\skt)>\bar\cR(\lmd_{\rm 1SE})+\eta_0$) are \emph{adversarial sources} that can incur negative transfer if selected.
Further, we need a separation level $\eta_\star>\eta_0$ such that every adversarial source is bounded away from the baseline,
\begin{equation}\label{eq:sep}
    \cR(\omg_{1\fm}\skt) \ge \bar\cR(\lmd_{\rm1SE}) + \eta_\star
    ~ \text{for all } k\in A_{\eta_0}^c .
\end{equation}
We define the risk-adaptive source selection margin as
$
    \t\eta:=c_{\t\eta}\brcs{\tau_{\eps_\fm}^2+\bar\cR(\lmd_{\rm1SE})}\sqrt{\frac{\log{p_1}}{\nt}},~~c_{\t\eta}>0 \text{ fixed}.$
\begin{theorem}\label{thm:ntd}
    Suppose Assumptions \ref{asm:ntd-desgn}, \ref{asm:ntd-cv}, and \ref{asm:ntd-margin} hold.
    Then, for every source index $k$,
    \begin{equation}\label{eq:fp-bound}
        \pr\lrp{k\in\h A,\, k\in A_{\eta_0}^c}
        \le
        c_1p_2\exp\lrp{-c_2\log{p_1}},
    \end{equation}
    where $c_1,c_2>0$ depend only on $(\set,\sxt,\sot,\lmd_{\min}(\Sgot),\lmd_{\max}(\Sgot),\kappa)$ but not on $K$ or $\nrm{\beta_2\st}{}$.
    Consequently,
    $
        \pr\lrp{A_{\eta_0}^c\cap\h A=\emptyset}
        \ge
        1-Kc_1p_2\exp\lrp{-c_2\log{p_1}}.
    $
\end{theorem}
\begin{remark}[Interpretation and scope]
    Theorem \ref{thm:ntd} shows that SAND effectively excludes sources whose oracle target risk exceeds the achievable baseline $\bar\cR(\lmd_{\rm1SE})$ by at least $\eta_\star$.
    Following the decomposition in \eqref{eq:excess-risk}, this separation can emerge from domain shifts such as $(\dlt_1\skt,\dlt_2\skt)$ or $\Dlt_\bP\sk$ (or a confluence of both).
    While SAND provides stringent false-positive control, it does not mandate the recovery of \emph{all} informative sources, i.e., $\brcs{\h A=A}$.
\end{remark}
The theoretical risk margin $\t{\eta}$ is approximated via the empirical plug-in estimator:
$$\h\eta = \brcs{\h\tau_{\eps_\fm}^2+\bar R(\lmd_{\rm1SE})}\sqrt{\frac{\log{p_1}}{\nt}},\quad\h\tau_{\eps_\fm}^2:=\frac{\nrm{\by_t-\Xt\t\beta_{1\fm}(\lmd_{\rm min})}{}^2}{\nt-\nrm{\t\beta_{1\fm}(\lmd_{\rm min})}{0}}$$ 
where the marginalized noise variance $\h{\tau}_{\eps_\fm}^2$ is estimated using the degrees-of-freedom adjusted residual sum of squares \citep{reid2016study}.

\section{Simulation}
\label{sec:sim}

We evaluate the finite-sample performance of the proposed HTL estimators. We refer to Appendix \ref{sec:sim-set} for a detailed description of the simulation setup.
\paragraph{Linear feature mapping}\label{sec:sim-lfm}
\begin{figure}[h]
    \centering
    \includegraphics[width=0.45 \textwidth]{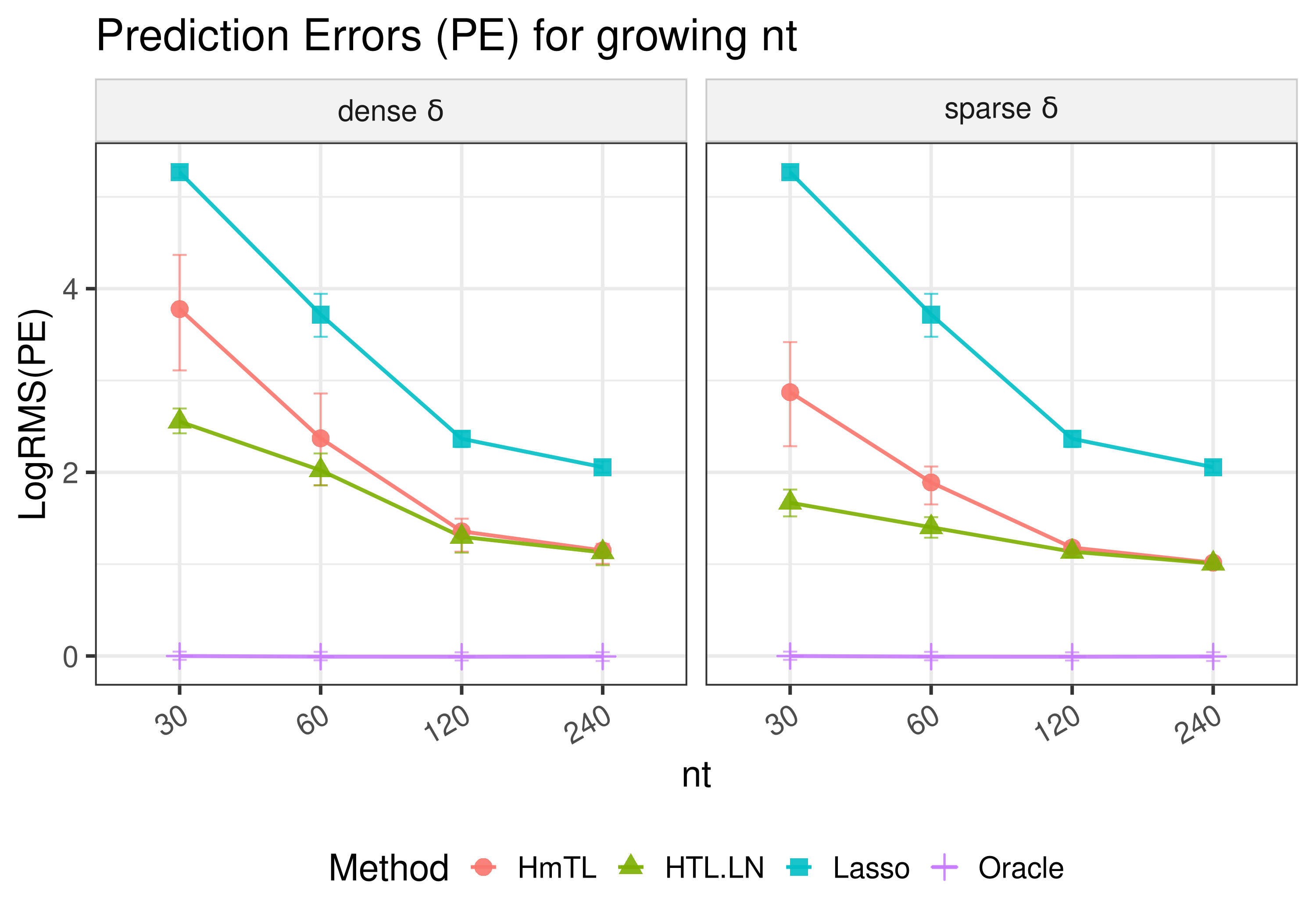}
    \includegraphics[width=.45\textwidth]{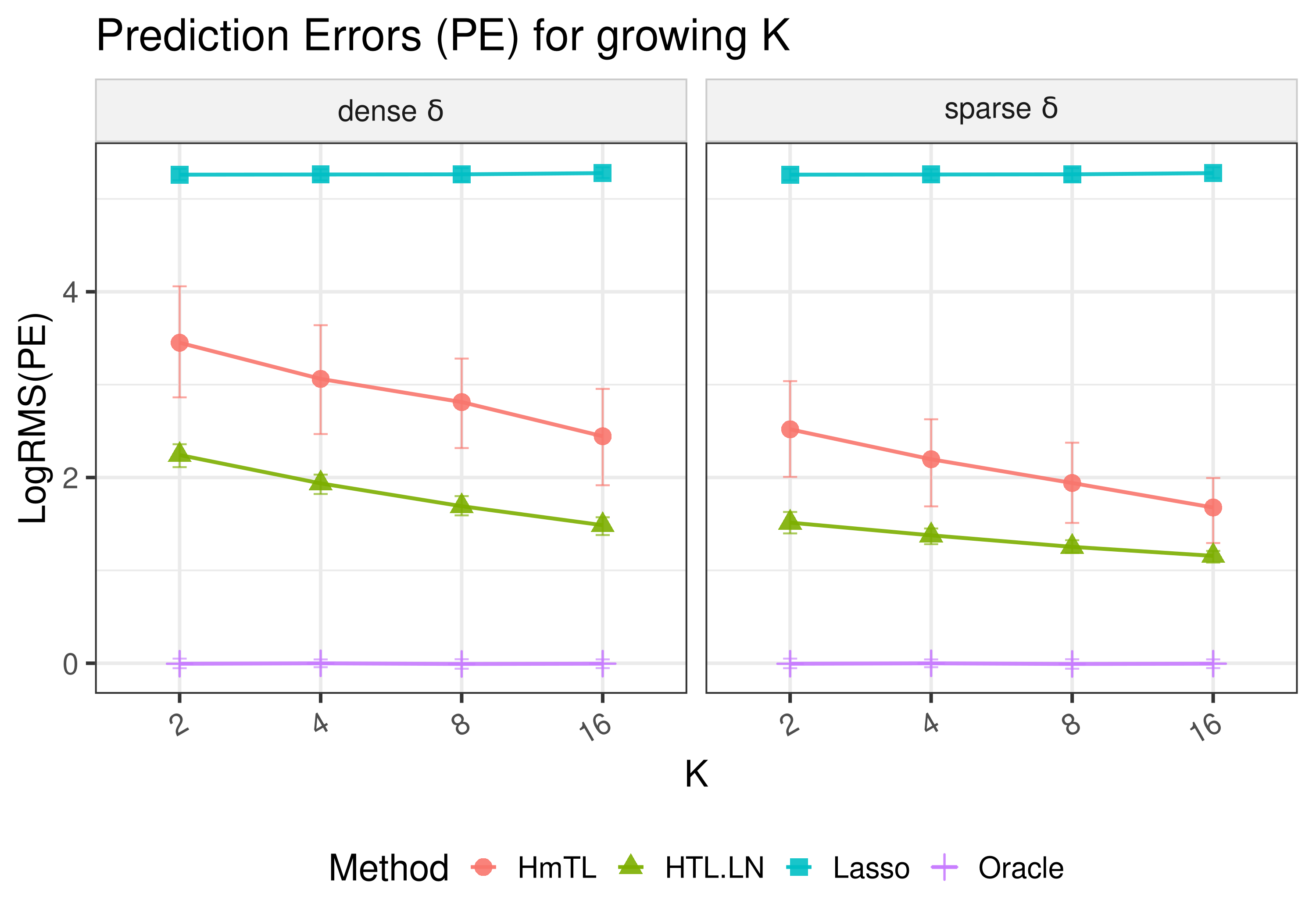}

    \caption{Prediction Errors of considered methodologies with sparse $\dst$ (\texttt{sparse $\delta$}) and non-sparse $\dst$ (\texttt{dense $\delta$}). In the left figure $n_t$ is increased keeping $K,n_s,n_t^\fn,p_1,p_2$ fixed, while in the right figure $K$ is increased keeping others fixed. 
    The upper and lower bars for each point indicate the 85th and 15th percentiles, respectively.}
    \label{fig:htl-lfm-sim}
\end{figure}
\begin{figure}[h]
    \centering
      \includegraphics[width=.48\textwidth]{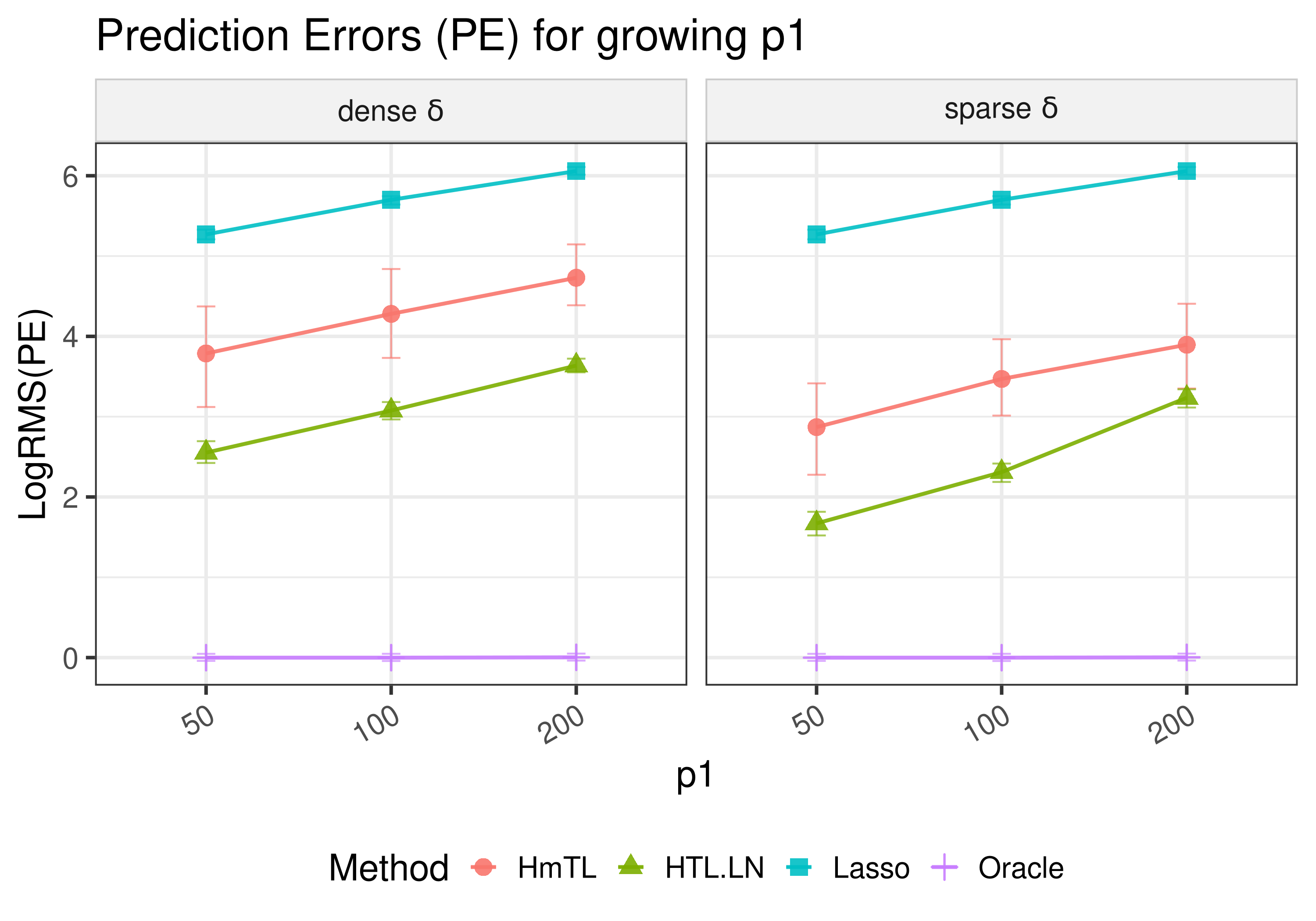}
    \includegraphics[width=.48\textwidth]{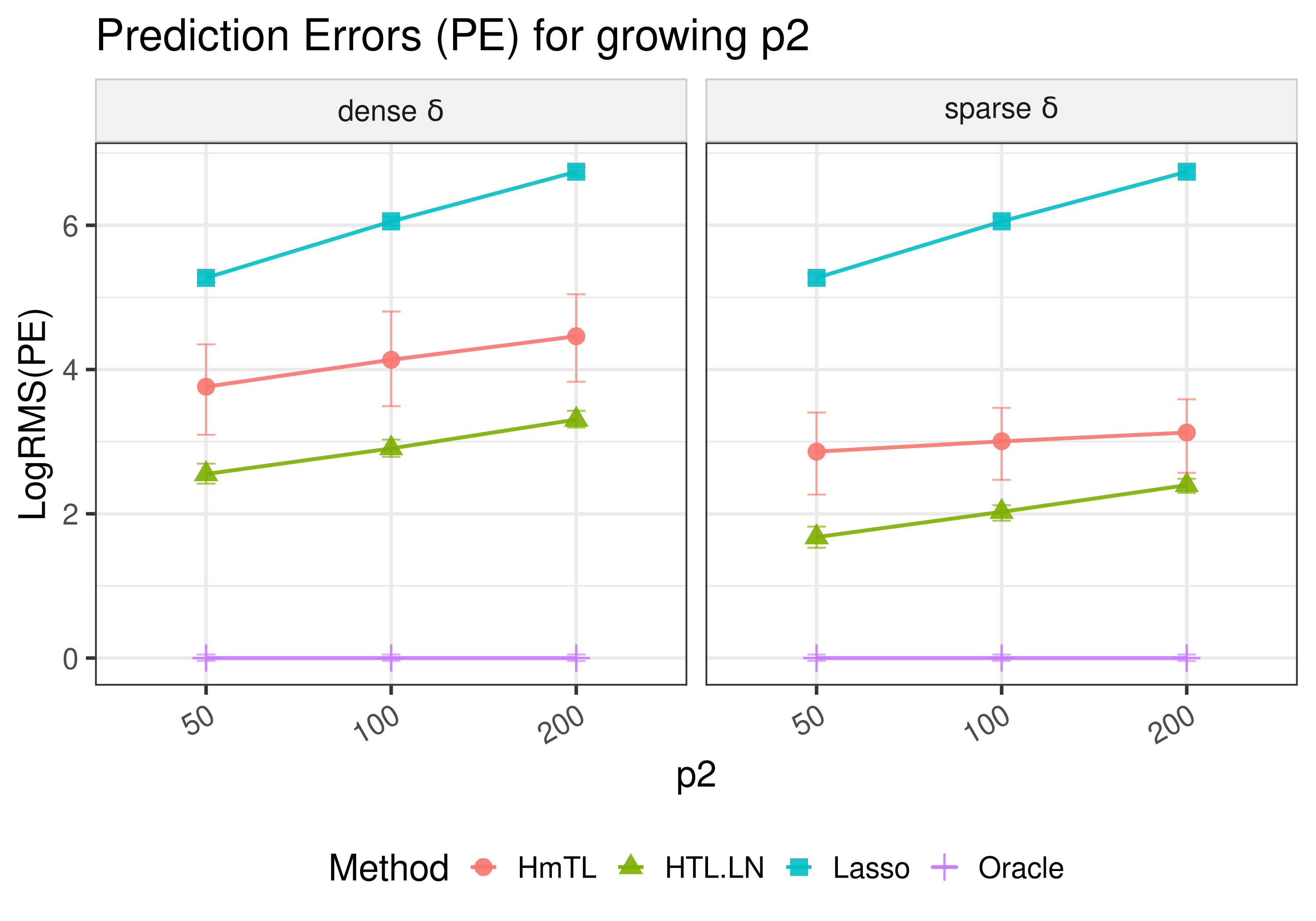}
    \caption{Prediction errors with (left) increasing $p_1$ and (right) increasing $p_2$.
    \vspace{-10pt}}
    \label{fig:htl-lfm-sim2}
\end{figure}
Across 200 Monte Carlo replications, we evaluate the out-of-sample predictive performance of competing methods on $n_t^\fn = 200$ test observations. We compare the following four procedures.
\texttt{Oracle} prediction errors serve as the irreducible test-domain noise $\epsilon_t^\fn$.
\texttt{HTL.LN} is the proposed heterogeneous transfer learning method with the linear feature map,
\texttt{HmTL} implements the homogeneous transfer learning baseline \citep{bastani2021predicting}, which naively restricts estimation to the matched (marginal) coefficients $\beta_{1\fm}\st$ while ignoring the unobserved features.
\texttt{Lasso} fits an $\ell_1$-penalized regression exclusively on the matched target sample.
We compute the predictive accuracy as the logarithm of the Root Mean Squared Prediction Error, defined as $\texttt{LogRMS(PE)} := \log \left\{ \frac{1}{n_t^\fn} \sum_{i=1}^{n_t^\fn} (y_{i, nt} - \widehat{y}_{i, nt})^2 \right\}^{1/2}$. The corresponding error rates across varying configurations of $(K, \np, \nt, p_1, p_2)$ are summarized in Figures \ref{fig:htl-lfm-sim} and \ref{fig:htl-lfm-sim2}.
Unless specified otherwise on the horizontal axis of each figure, the baseline configuration is set to $K=1$, $\np=400$, $\nt=30$, and ${p_1}={p_2}=50$.

In the first two plots of Figure \ref{fig:htl-lfm-sim}, the logarithm of the root-mean-squared prediction errors is reported as the target sample size $n_t$ increases. As expected, all the considered methodologies show a significant decrease in prediction error as $n_t$ increases. 
However, the proposed \texttt{HTL.LN} procedure dominates both the \texttt{HmTL} baseline and the target-only \texttt{Lasso} across all target sample sizes $n_t$, under both sparse and dense $\dst$ regimes.
When $n_t$ is smallest ($n_t=30$), the performance gap is biggest, and our method is most beneficial, empirically validating Theorems \ref{thm:lfm-sp-err} and \ref{thm:lfm-sp-hm-err}.
The next two plots in Figure \ref{fig:htl-lfm-sim} consider multiple source studies, demonstrating the benefits of aggregating multiple informative knowledge sources for the proposed HTL method. The performance of both TL methods improves with increasing number of source studies, with \texttt{HTL.LN} outperforming \texttt{HmTL} throughout. As expected, the performance of the target \texttt{Lasso} does not improve.
The effect of increasing model dimension on the prediction errors in Figure \ref{fig:htl-lfm-sim2} reveals the benefits of the HTL as well.
While the predictive performance of all methods deteriorates as the number of features $p_1$ and $p_2$ grows, \texttt{Lasso} and \texttt{HmTL} show a clear performance gap compared to \texttt{HTL.LN}.

\begin{figure}[h]
    \centering
      \includegraphics[width=.49\textwidth]{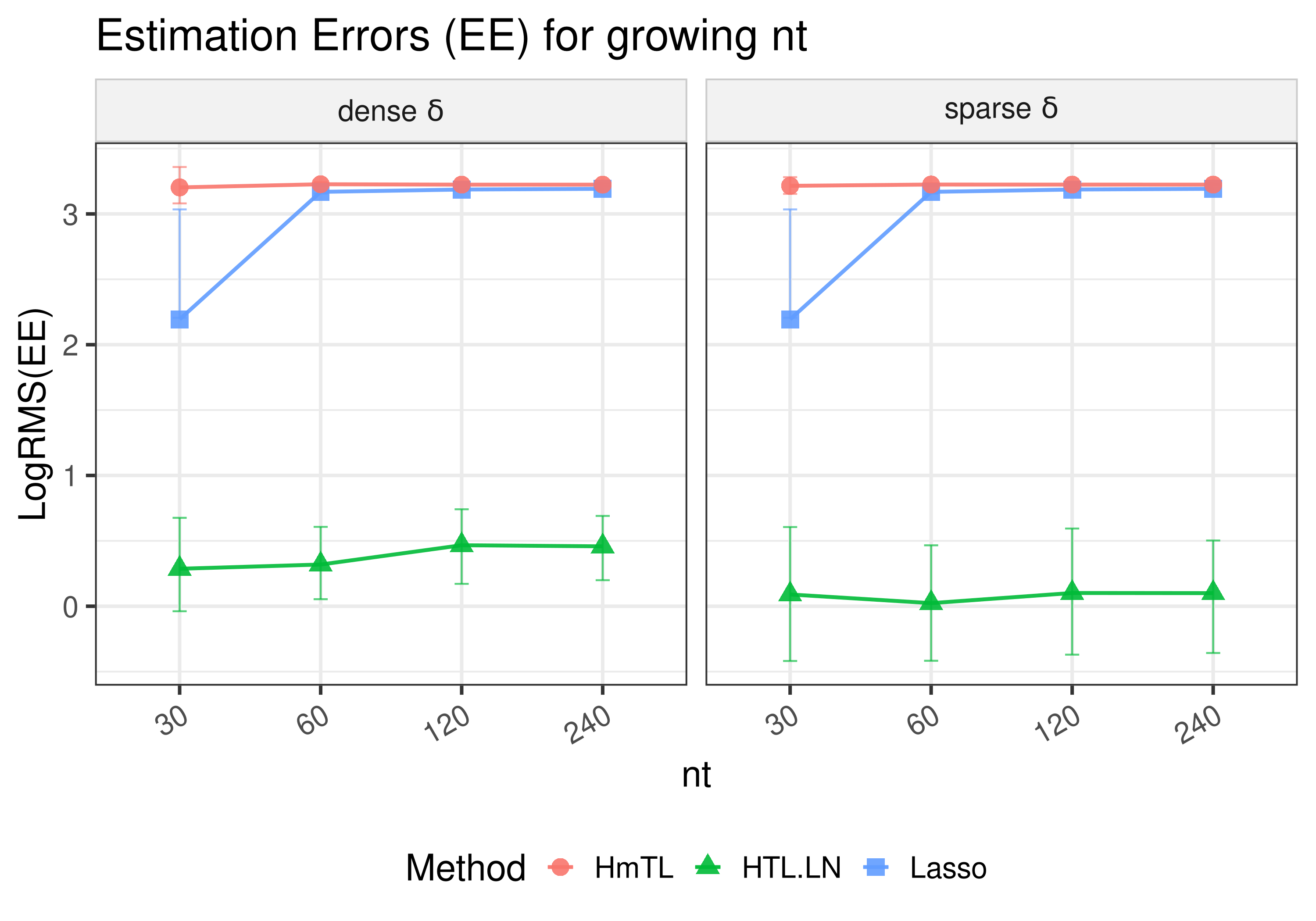}
      \includegraphics[width=.49\textwidth]{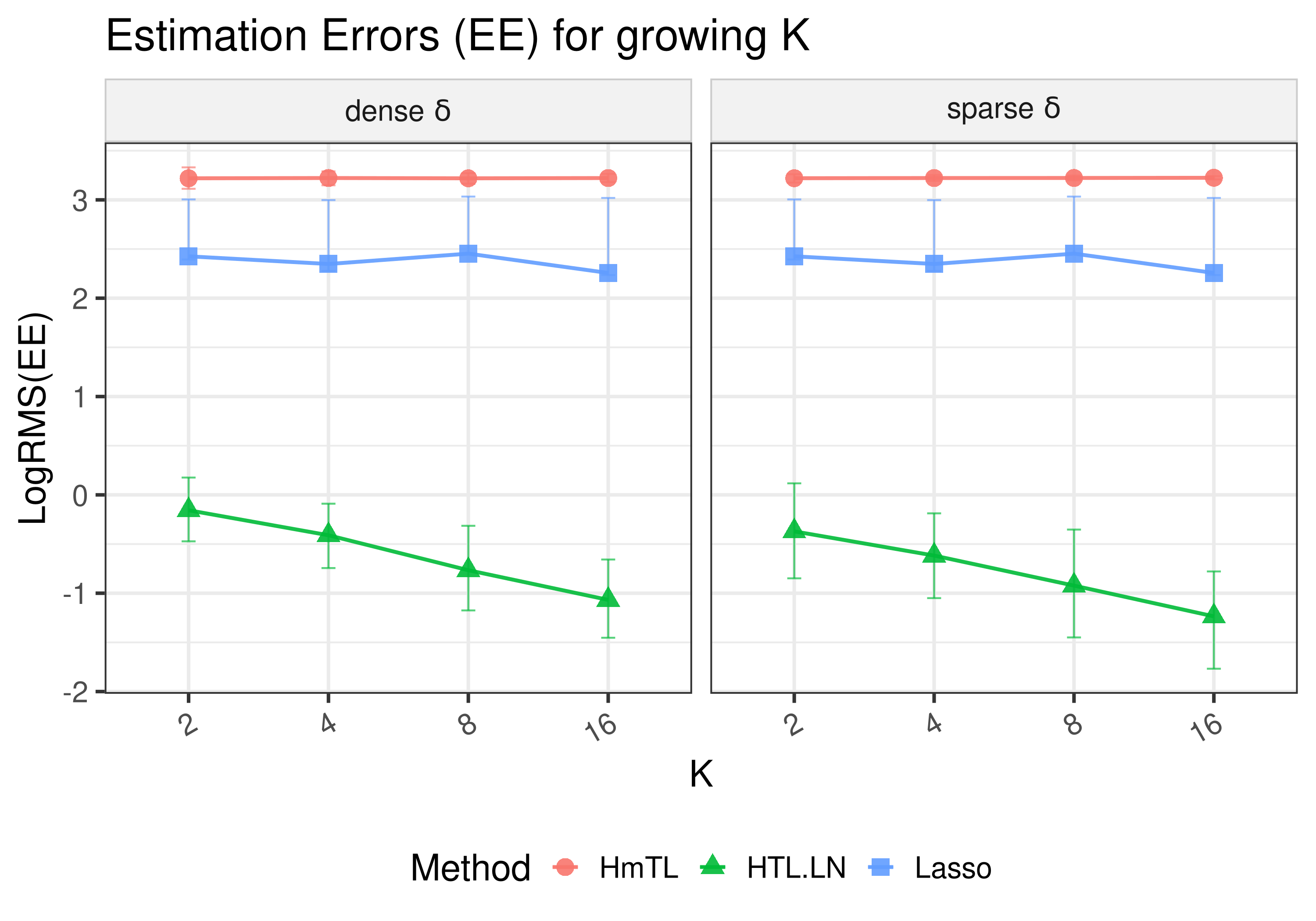}
    \caption{Estimation error of considered methodologies with sparse $\dst$ and non-sparse $\dst$, with increasing $n_t$ and  $K$ respectively.
    \vspace{-10pt}}
    \label{fig:htl-lfm-sim3_short}
\end{figure}
We also compute estimation errors of the matched target feature effect $\beta_1\st$ in Figure \ref{fig:htl-lfm-sim3_short} and additional figures in \ref{fig:htl-lfm-sim3}, measured in $\ell_1$-norm as $\nrm{\h\beta_1-\beta_1\st}{1}$ for each estimator.
Both \texttt{HmTL} and \texttt{Lasso} exhibit substantial estimation bias that does not decrease with increasing $n_t$. The estimation errors from those two methods remain elevated across all dimensionalities ($p_1, p_2$) and number of source domains ($K$). This is consistent with the analysis in Theorem \ref{thm:lfm-sp-hm-err}, which demonstrates the large omitted variable bias when mismatched covariates $\bZ$ are ignored.
Conversely, \texttt{HTL.LN} yields the most accurate estimation under all scenarios.

\paragraph{Non-linear feature mapping}\label{sec:sim-nlfm}
\begin{figure}[h]
    \centering
    \includegraphics[width=0.48 \textwidth]{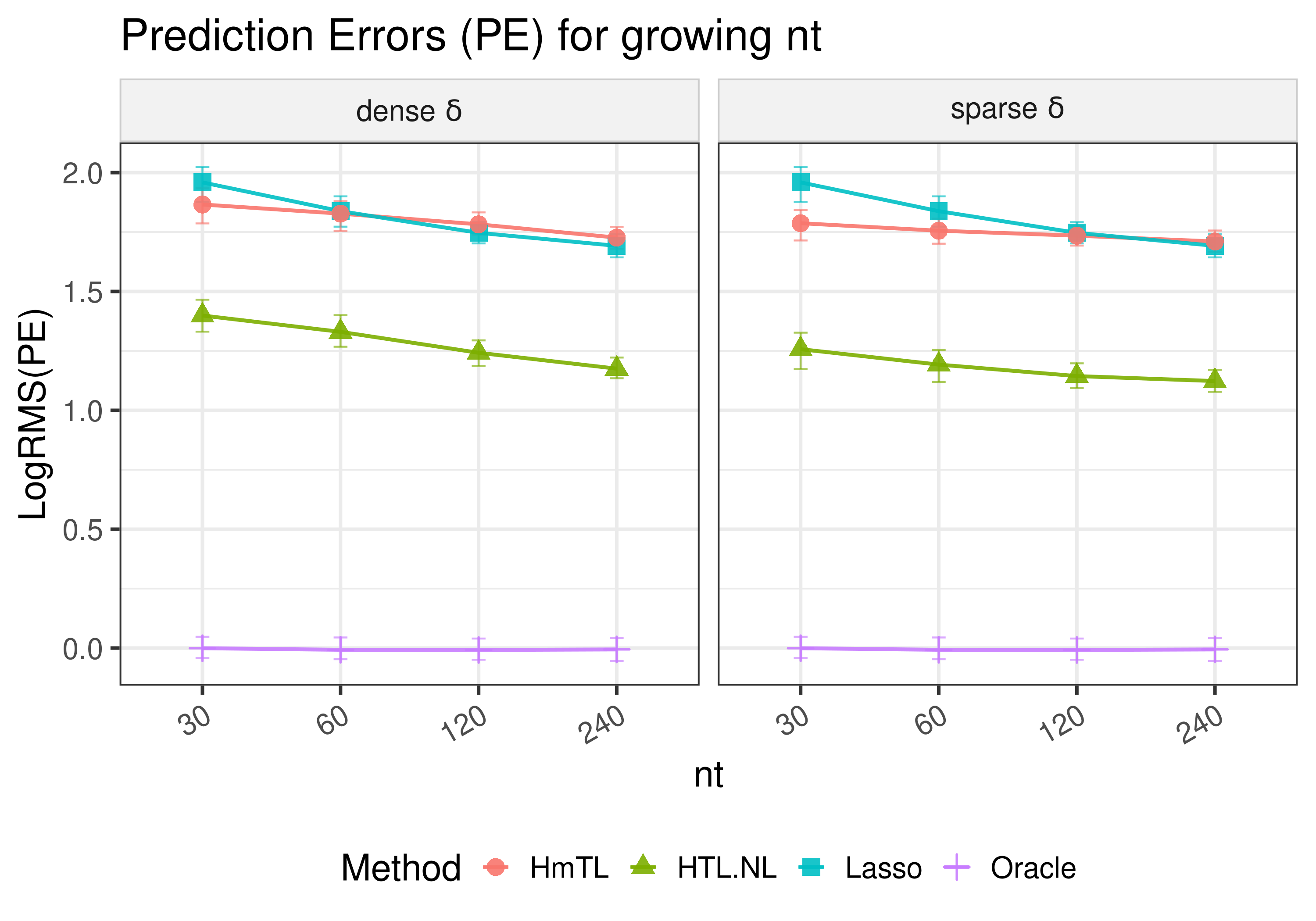}
    \includegraphics[width=.48\textwidth]{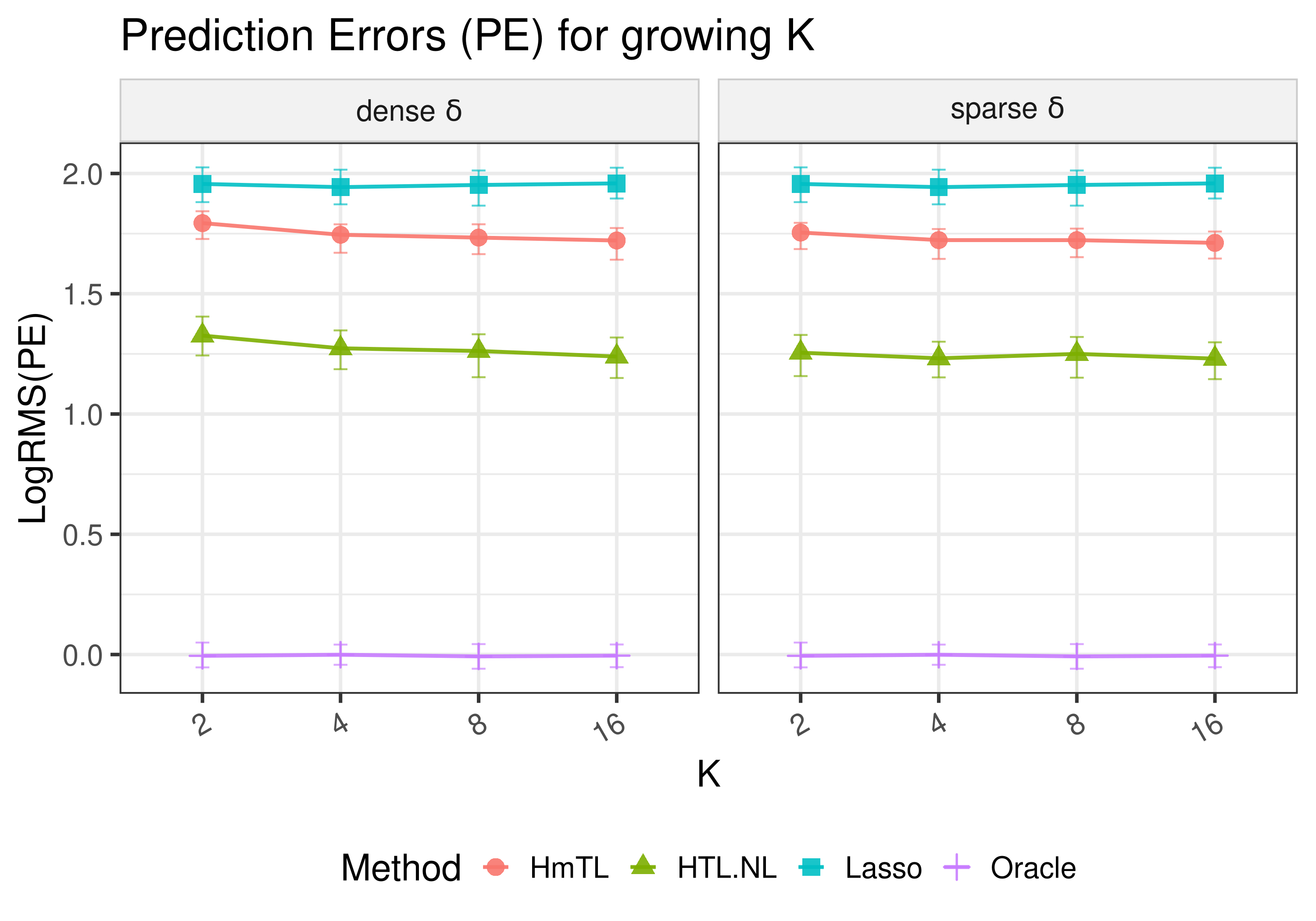}
    \caption{Prediction errors of considered methodologies for non-linear feature map in sparse $\dst$ and non-sparse $\dst$ cases, with increasing $n_t$ and increasing $K$, respectively.}
    \label{fig:htl-sim}
\end{figure}

\begin{figure}[h]
    \centering
    \includegraphics[width=.48\textwidth]{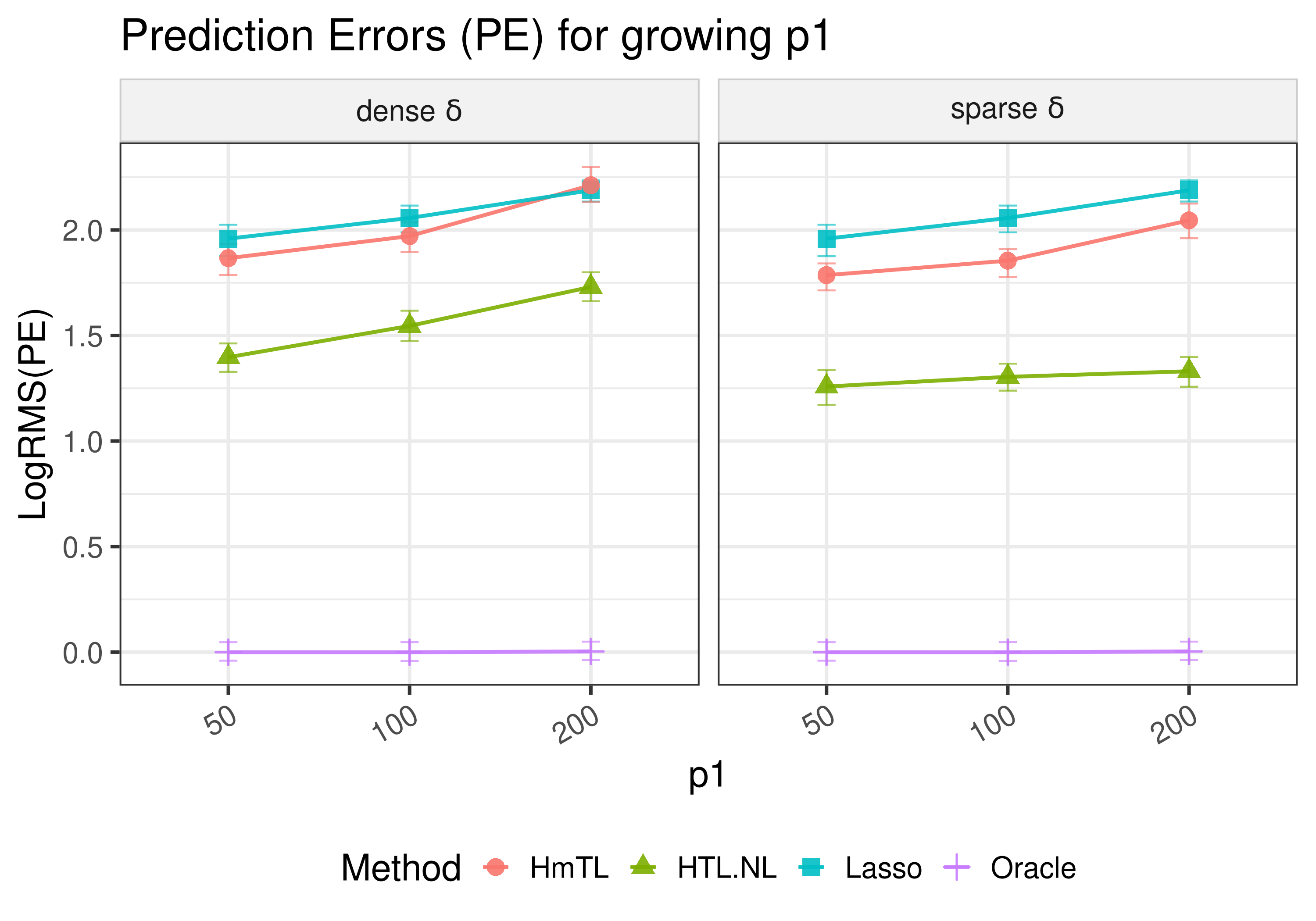}
    \includegraphics[width=.48\textwidth]{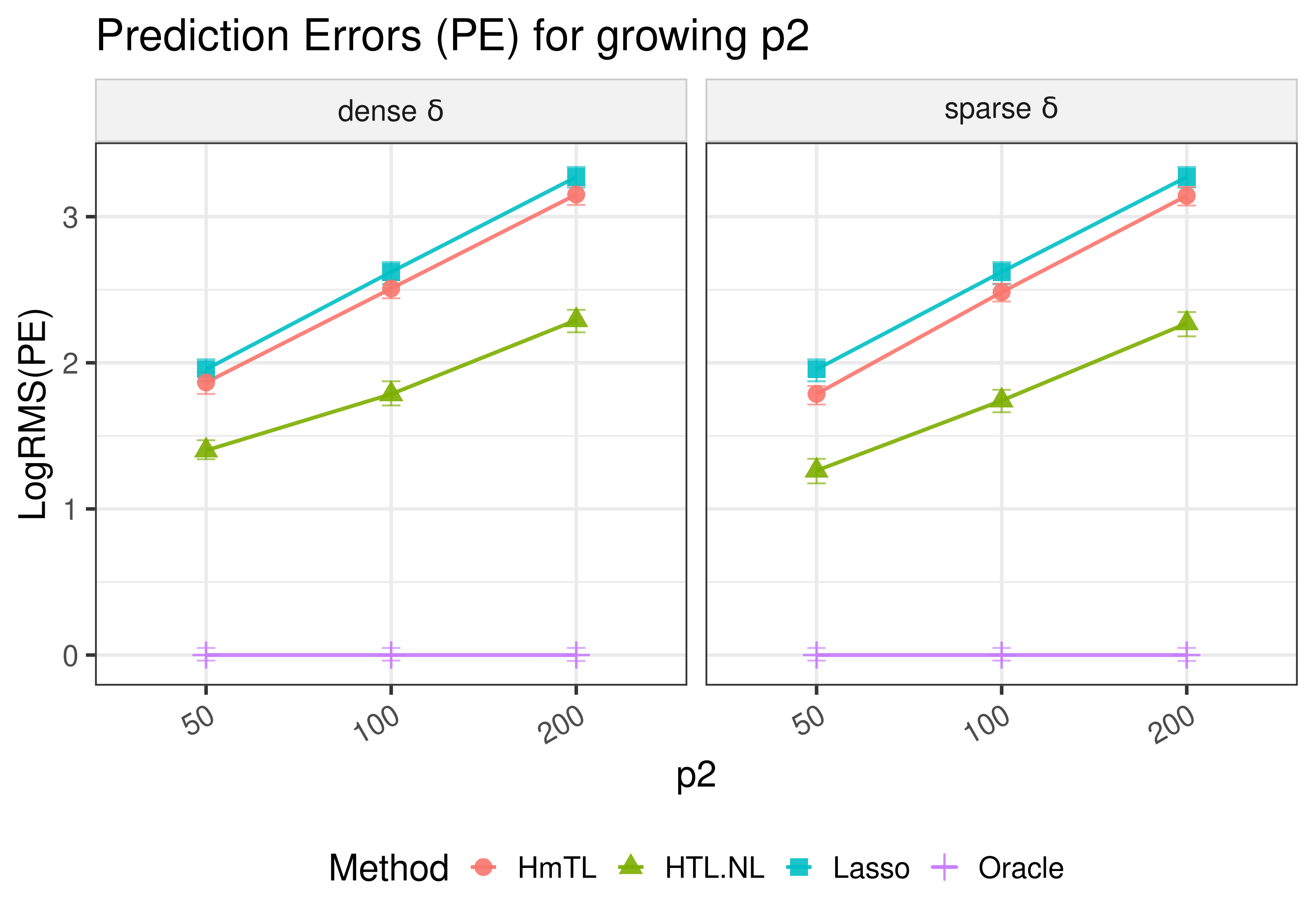}
    \caption{Prediction error of considered methodologies for non-linear feature map with increasing $p_1$ and $p_2$, respectively.
    \vspace{-10pt}}
    \label{fig:htl-sim2}
\end{figure}
When the feature mapping is assumed to be an unknown function, \texttt{HTL.NL} uses sieve estimation for imputing mismatched features $\bz$ in the target and test datasets. For each source $k$, we obtain $\h\Theta_s\sk$ with cross-validation using \texttt{Sieve} package in R \citep{zhang2023sieve}, and then obtain $\h\Theta_s=\frac{1}{K}\sum_{k\le K}\h\Theta_s\sk$ and the number of bases as $M=\max_{k\le K}M_k$. We calculate $\t\bD_t^\fn\h\beta$, with $\t\bD_t^\fn$ obtained via imputation of $\bZ_t^\fn$ as $\h\bZ_t^\fn=\Psi_M(\Xt^\fn)\h\Theta_s$. The other three methods for comparison, namely, \texttt{HmTL}, \texttt{Lasso}, and \texttt{Oracle}, remain the same as before.

In Figure \ref{fig:htl-sim}, we see that the prediction error in \texttt{HmTL} remains much higher than that of HTL even as $n_t$ and $K$ increase. \texttt{HTL.NL}, on the other hand, strongly outperforms both \texttt{HmTL} and target \texttt{Lasso} as it effectively reduces the omitted variable bias.
In Figure \ref{fig:htl-sim2} again, the prediction error of all methods grows in model complexity, but \texttt{HTL} continues to produce the most accurate predictions. Overall, in the case of a non-linear feature map, the prediction performance gap between \texttt{HTL.NL} and \texttt{HmTL} remains high in all scenarios. This is expected intuitively and from our theoretical results, since the HmTL fails to model both $\omg_2\st$ and the feature mapping $\bh$.
Our proposed \texttt{HTL.NL} also produces the most accurate estimations of $\h\beta_1$ in Figure \ref{fig:htl-sim3} in the Appendix, agreeing with its theoretical comparisons with \texttt{HmTL}. In all cases, the HmTL method performs poorly and often fails to outperform even the baseline \texttt{Lasso} in the target domain.

\section{Case Study: Ovarian Cancer Gene Expression Data}\label{sec:app}
We apply our transfer learning approach to two microarray gene expression datasets available from a curated data collection on ovarian cancer (see  \citet{curatedovarian} for details on data collection).
We are interested in predicting the survival time of study participants from the day of the microarray test. Besides guiding clinical decision-making, predicting a patient's prognosis can inform
and improve design, conduct, and data analysis of clinical trials and real-world studies~\citep[e.g.][]{loureiro:2021}.
Studies typically focus on a subset of genes that are correlated with or predictive of a specific outcome variable (e.g., overall or progression-free survival). However, other data sources targeting different populations are also available and can be used to improve predictions \citep{zhao:2022}. Approaches incorporating transfer learning ideas have been considered for these prediction tasks and have shown potential in empirical analyses \citep[e.g.,][]{zhao:2022,jeng:2023}.  To our knowledge, most of these methods use heuristics to transfer knowledge across datasets without proper statistical formulations.

In this section, we focus on patients with ovarian cancer and show that our transfer learning approach can improve prediction compared to discarding or partially using source data. 
We use two public datasets from the curated ovarian cancer collection (described below).
In our example, we consider microarray data, but the same approach can be applied to other features, such as demographics or blood samples, which are typically available in hospital and clinic data \citep [e.g.,][]{becker:2020}.
In such contexts, the target data may be clinical trials, which typically have small sample sizes and target specific populations.

Two datasets from the curated ovarian cancer collection are selected: The one presented in \citet{tothill:2008} (as source) and the one 
in \citet{bonome:2008} (as target). From now on, we will refer to these datasets only as target and source to simplify the exposition.  From both datasets, we select patients with information on survival status: $113$ for the source and $129$ for the target. 
We note that in our example, the sample sizes for the target and source data are similar; nevertheless, HTL methods can still provide meaningful improvements in estimation and prediction tasks. 
As a response variable, we consider the logarithm of the number of days from enrollment in the study until death.
Before the analysis, we select a subset of genes correlated with survival. 
In our application, we base this feature screening on the target and source data, but more general contextual knowledge or other external data should be used for this task.  
Specifically, we fit an elastic net model to the target datasets and select the smallest available tuning parameter. Of these variables, we keep the $p_1=393$ that are common to both the target and the source. We use the same screening method for the resulting $p_2=335$ variables as our $\Zp$ matrix.
 
We compare the predictive performance of four models: Lasso on the target data (\texttt{Lasso}), homogeneous transfer learning (\texttt{HmTL}), and our approach with linear feature map estimation (\texttt{HTL.LN}) and sieve method for modeling non-linear feature mapping (\texttt{HTL.NL}). 
 
\begin{figure}[!htbp]
    \centering
    \begin{subfigure}[t]{0.48\textwidth}
        \centering
        \includegraphics[width=\textwidth]{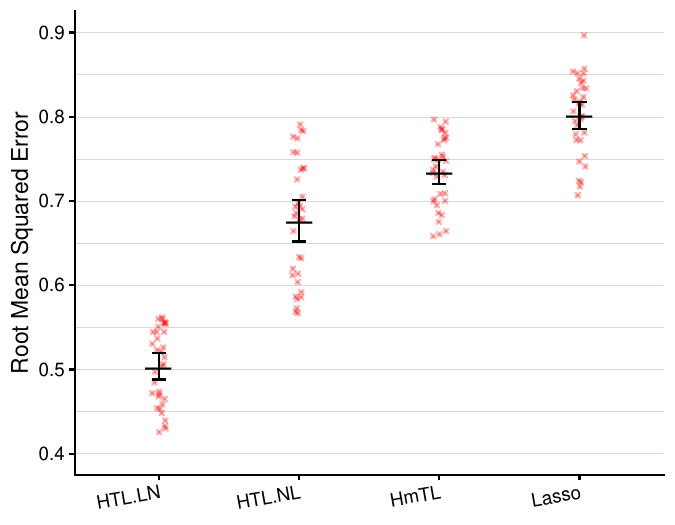}
        \caption{Root mean squared error (RMSE) for the approaches considered in Section~\ref{sec:app}. For each method, the dashed line denotes the mean RMSE, the error bar is an approximate 95\% interval for the mean RMSE 
        .}
        \label{fig:cv}
    \end{subfigure}
    \hfill
    \begin{subfigure}[t]{0.48\textwidth}
        \centering
        \includegraphics[width=\textwidth]{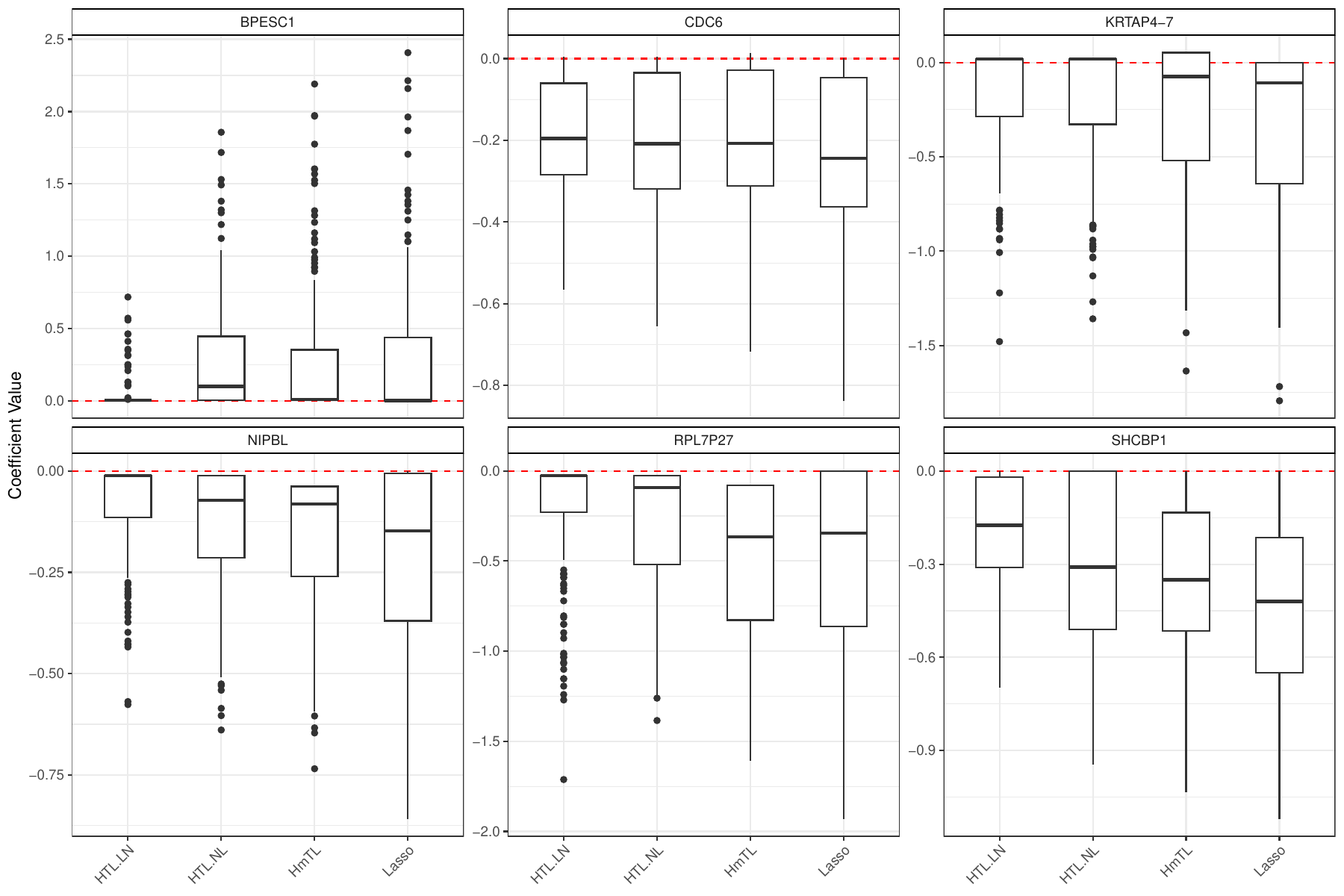}
        \caption{Bootstrap distributions of the six regression coefficients.}
        \label{fig:boot}
    \end{subfigure}
    \caption{Cross-validation RMSE and bootstrap results for the application in Section~\ref{sec:app}.}
    \label{fig:coc_results}
\end{figure}

Repeated cross-validation (CV) was used to approximate the out-of-sample prediction error by randomly dividing the target data into five folds, using four folds to train the model, and one fold to compute the root mean squared prediction error. This CV is repeated 10 times, yielding 50 CV errors in total. The prediction error estimates are biased because the response variable has been used in feature screening, but the relative comparison remains valid. The estimated CV errors are $0.802$ for \texttt{Lasso},  $0.734$ for \texttt{HmTL}, $0.677$ for \texttt{HTL.NL}, and $0.504$ for  \texttt{HTL.LN}.
Figure \ref{fig:cv} shows the results with trimmed CV errors, keeping only points between the 15th and 85th percentiles. All TL methods considered improve prediction over \texttt{Lasso}, and our approaches (\texttt{HTL.LN} and \texttt{HTL.NL}) yield the smallest CV errors.

Comparing the two HTL approaches, we find that \texttt{HTL.LN} has a significantly smaller CV error than \texttt{HTL.NL}. This can be attributed to the source sample size (113), which may be too small to recover the complex basis expansions, whereas the linear feature mapping estimation is more robust with column-wise lasso regression. An alternative explanation is that the true unknown feature mapping is nearly linear,  making the linear estimate more accurate. In this case, sieve estimation can introduce unnecessary approximation errors due to higher-order Fourier expansions.

To better understand the differences among the considered methods, we also compare the distribution of regression coefficients obtained via non-parametric bootstrap on the target data, with the source data held fixed. 
Figure \ref{fig:boot} shows the bootstrap distribution of six regression coefficients estimated using \texttt{HTL.NL}, \texttt{HTL.LN},
\texttt{HmTL}, and \texttt{Lasso}.
Although no ground truth is known, the medians of the bootstrap distributions often agreed across the four methods, suggesting that the approaches have comparable bias. For example, for the gene \textit{BPESC1}, the median is approximately zero across all samples. However, genes such as \textit{RPL7P27} showed greater shrinkage towards zero with HTL methods than with HmTL or Lasso.
Because of the more extensive use of the source data, the variability in the coefficient estimates of HTL methods is typically smaller than that of Lasso, suggesting a smaller standard error for the \texttt{HTL.NL} and \texttt{HTL.LN} estimators.

\section{Discussion}
In this article, we focus on heterogeneous transfer learning when the source domain has auxiliary features that are not observed in the target domain.
The fundamental premise that needs to be validated before applying our method is that missing features are true predictors of the target response variable, i.e., $\beta_2\st\ne(0)$.
Also, several critical avenues remain open for future investigation. First, there are other types of feature mismatches, such as cases where the target domain observes a larger feature space and cases where both the target and source domains observe unique features not shared between them. While we leveraged the rich data environment of the source domain to learn the feature mapping, these regimes may require fundamentally different imputation strategies.
Additionally, the current SAND algorithm hard-thresholds adversarial sources from the data pool. 
There is a criticism that excluding the entire dataset from knowledge transfer can lead to the loss of informative observations.
Developing an adaptive, continuous weighting scheme that can safely aggregate partial signals could further enhance the results. Finally, while our method focuses on parameter estimation and prediction, inferential results (e.g., confidence intervals) under feature mismatch remain open.

\paragraph{Acknowledgment}
The authors would like to acknowledge Kiljae Lee and Profs Jared Huling and Arnab Auddy for helpful discussions and feedback. This research is partially funded by an NSF Division of Mathematical Sciences grant (DMS 2529302).

\vspace{-10pt}

\bibliographystyle{apalike}
\bibliography{references.bib}

\newpage
\appendix
\addcontentsline{toc}{section}{Appendix}
\part*{Appendix}
\addtocontents{toc}{\protect\setcounter{tocdepth}{3}}
\tableofcontents

\section{Notations and Definitions}\label{sec:design}
Define a $p$-dimensional unit sphere as $S^{p-1}=\brcs{\bd\in\R^p;\nrm{\bd}{}=1}$.
Given a real-valued random variable $X$, denote its sub-Gaussian norm by $\nrm{X}{\psi_2}:=\inf\brcs{t>0;\E\exp\lrp{X^2/t^2}\le2}$.
We write $\lmd_{\min}(\bA)$ and $s_{\min}(\bA)$ to denote the minimum eigenvalue and the minimum singular value ($\lmd_{\max},s_{\max}$ are used analogously) of $\bA$.
For an index set $A\subseteq[p]$ of a vector $\bx\in\R^p$, define $\bx_A\in\R^{|A|}$ as $(\bx_A)_i:=x_i$ for $i\in A$.
For the convenience of asymptotics, we denote generic positive constants by $c,c'$.
We write $a_n=\cO(b_n)$ if $\lim\sup_{n\to\infty}|a_n/b_n|<\infty$ and $a_n \ll b_n$ if $\lim_{n\to\infty}|a_n/b_n|=0$.
Also, we write $a_n\asymp b_n$ if both $a_n=\cO(b_n)$ and $b_n=\cO(a_n)$ (hence $a_n\asymp1$ means that $|a_n|$ is bounded from below and above by positive constants). The notations 
$o_\pr(1)$ and $\cO_\pr(1)$ denote the conventional convergence and boundedness in probability, respectively.
For matrix norm notations, let $\nrm{\bA}{}:=\sup_{\bu\in\cS^{d-1}}\nrm{\bA\bu}{}$ denote the matrix $\ell_2$ norm (spectral norm), $\nrm{\bA}{F}:=\tr\lrp{\bA'\bA}^{1/2}$ denote the Frobenius norm, and $\nrm{\bA_{n\times d}}{\infty}:=\max_{i\le n}\nrm{\ba_i}{1}$ denote the maximum row sum norm.

To proceed, we need a few definitions. The following gives the formal definitions of sub-Gaussian random variables, random vectors, and random ensembles used in this paper.
\begin{definition}($\sgm$-sub-Gaussian random variable, vector, and ensemble).
    A zero-mean random variable $X$ is said to be a $\sgm$-sub-Gaussian random variable if there exists a constant $\sgm>0$ such that
    $\E\exp\lrp{uX}\le\exp(u^2\sgm^2/2)$
    for any $u\in\R$. Then, we write $X\sim SG(\sgm)$.
    If a zero-mean random vector $\bx\in\R^d$ satisfies $\bu'\bx\sim SG(\sgm)$ for all $\bu\in S^{d-1}$, then we say $\bx$ is a $\sgm$-sub-Gaussian random vector and write $\bx\sim SG(\sgm)$.
    If the rows of a centered random matrix $\bX\in\R^{n\times d}$, $\bx,...,\bx_n$, are i.i.d. $\sgm$-sub-Gaussian random vectors, then we say $\bX$ is a row-wise $\sgm$-sub-Gaussian ensemble and write $\bX\sim SG(\sgm)$.
\end{definition}

\section{Assumptions for nonparametric modeling of $\E(\bz|\bx)$}\label{nlfm_app}
The class of functions $\cH$ described in section \ref{sec:nlfm} is introduced in \cite{zhang2023regression}.
The sieve estimation method leverages the fact that $h_j$ can be expressed as an infinite linear combination. 
We write this sequence of basis functions as $\psi_m(\bx)$, which are unraveling of $\psi_\bq(\bx):=\prod_{k=1}^{p_1}\phi_{q_k}(x_k)$ for $\bq:=(q_1,\dots,q_{p_1})\in\bb N^{p_1}$ and 
\[\phi_{q}(x):={\bb I}_{\brcs{q=1}}+\sqrt{2}{\bb I}_{\brcs{q>1}}\cos\brcs{\frac{(q-1)\pi(x+a)}{2a}},\quad x\in[-a,a],\,q\in\bb N\]
which is an orthogonal univariate cosine basis such that $\int_{[-a,a]}\phi_q(x)\phi_r(x)dx$ is $0$
if $q\ne r$, and is $2a$ if $q=r$. 
We follow the c-unraveling of $\psi_\bq$ described in \cite{zhang2023regression} to obtain $\psi_m(x)$. 
Then, $h_j$ and $h_j\st$ have the expansion
\begin{align}\label{eq:tht-exp}
    h_j(\bx)=h\st_j(\bx_{S_1})=\sum_{m=1}^\infty\psi_m(\bx)\tht_{mj}\quad\text{for}\quad\brcs{\tht_{mj}}\in\R,
\end{align}
i.e., $\bh(\bx)=\sum_{m\ge1}\psi_m(\bx)\btht_m$ for $\btht_m:=(\tht_{m1},...,\tht_{m,{p_2}})'$.
We additionally assume that $\E[\bh(\bx)] = \b0_{p_2}$, i.e.,
$\int_{\cD_x}\sum_{m\ge1}\btht_m\psi_m(\bx)f(\bx)d\bx=\b0_{p_2},$
which implies $\E(\bz)=\b0_{p_2}$.

This naturally motivates the notation for $\bH(\bX)$ as $\bH(\bX)=\Psi_\infty(\bX)\Tht_\infty$,
where each entry of $\Psi_\infty$ is $\Psi_\infty(\bX)_{im}:=\psi_m(\bx_i)$ and $(\Tht_\infty)_{mj}:=\tht_{mj}$ for $m=1,2,...$ and $j\in[p_2]$ given $n$ covariate samples $\bx_1,...,\bx_n$. 
Write the $i$-th row of $\Psi_\infty(\bX)$ as $\b\psi_\infty(\bx_i)\in\R^\infty$ so that $\bh(\bx_i)=\Tht_\infty'\b\psi_\infty(\bx_i)$.
This formulation for both domains can be compactly expressed as $
\bH_t(\Xt)=\Psi_\infty(\Xt)\Tht_{t,\infty}, \bH_s(\Xp)=\Psi_\infty(\Xp)\Tht_{s,\infty}.$
We note that the basis expansion mechanism $\Psi_\infty$ is identical in both domains, whereas the coefficients $\Tht_{\infty}$ differ.

If the infinite-order basis expansion is truncated at $m=M$, then we write
$\Psi_M(\Xp):=[\b\psi_M(\bx_{1,s}),\dots,\b\psi_M(\bx_{\np,s})]'\in\R^{\np\times M}$,
for $\b\psi_M(\bx):=(\psi_1(\bx),\dots,\psi_M(\bx))'\in\R^M$, which contains unraveled basis functions evaluated at each data point $\bx_{i,s}$.

By construction, each element of $\Psi_M(\Xp)$ is bounded: according to the unraveling rule in \citet[Condition C.9]{zhang2023regression}, there exists some constant $c>0$ such that the basis function satisfies $\frac{\psi_m(\bx)}{c^{{p_1}'}}\in[-1,1]$ for $m\in[M]$ and some fixed ${p_1}'\ge\post$. Therefore, for some $c_1>0$,
\[\nrm{\bu'\b\psi_M(\bx)}{\psi_2}\lesssim c_1^{p_1'}=:\sgm_\psi\]
for any $\bu\in S^{M-1}$. Hence $\b\psi_M(\bx_1),...,\b\psi_M(\bx_n)$ are i.i.d. $\sgm_\psi$-sub-Gaussian random vectors.

By the orthogonality of the chosen sieve basis, the second moment matrix of the finite-dimensional feature mapping strictly reduces to the identity:
\begin{align*}
    \Sgm_{\b\psi_M} &:= \E[\b\psi_M(\bx)\b\psi_M(\bx)'] = \int_{\cD_x} \b\psi_M(\bx)\b\psi_M(\bx)' \frac{1}{\operatorname{Vol}(\cD_x)} d\bx = \bI_M.
\end{align*}
This orthogonality rigorously extends to the infinite-dimensional limit ($M \to \infty$). Consequently, under the assumed uniform measure $\bx \sim \operatorname{Uniform}([-a,a]^{p_1})$, the population variance of the true non-linear mapping $\bh(\bx)$ is fundamentally diagonalized, yielding:
\begin{align*}
\operatorname{Var}[\bh(\bx)] = \int_{\cD_x} \frac{1}{\operatorname{Vol}(\cD_x)} \sum_{m_1, m_2 \ge 1} \psi_{m_1}(\bx)\psi_{m_2}(\bx) \btht_{m_1}\btht_{m_2}' d\bx = \sum_{m \ge 1} \btht_m\btht_m' = \Tht_\infty' \Tht_\infty.
\end{align*}
Under Assumption \ref{asm:smooth-h}, each coordinate function of $\bh(\bx)$ resides within a Sobolev ellipsoid \citep{zhang2023regression}. This geometric restriction inherently enforces a strict decay rate on the basis coefficients. Specifically, under the $c$-unraveling sequence of $\psi_{\bq}$, the coefficient magnitudes satisfy:
\[\max_{j\le{p_2}}\sum_{m\ge1}w_m^2\tht_{mj}^2<\infty,\quad \text{where} \quad w_m:=\frac{m}{\log^{\post-1}m},\]
which immediately guarantees the uniform absolute convergence of the unweighted coefficient energy, i.e., $\max_{j\le{p_2}}\sum_{m\ge1}\tht_{mj}^2<\infty$.
For the following, we write $\pi:=\nrm{\Tht_\infty}{}$.
\begin{assumption}\label{asm:senergy}
    Assume that there exists some constant $C_H>0$ such that
    $$\sum_{m \ge 1}\lrp{\frac{m}{\log^{\post-1} m}}^2\nrm{\b\tht_m}{}^2 \le C_H\pi^2.$$
\end{assumption}
Assumption \ref{asm:senergy} imposes a collective spectral decay condition. Specifically, for any unit direction $\bu \in S^{p_2}$, the structured energy satisfies:$$\begin{aligned}
\bu' \left( \sum_{m\ge1} w_m^2 \btht_m \btht_m' \right) \bu = \sum_{m\ge1} w_m^2 (\btht_m' \bu)^2 &\le \sum_{m\ge1} w_m^2 \|\btht_m\|^2 \|\bu\|^2 \lesssim \pi^2.
\end{aligned}$$This uniform upper bound ensures that the operator norm of the population variance is strictly controlled by the aggregate spectral energy, yielding $\|\Var(\bh(\bx))\| = \mathcal{O}(\pi^2)$. 
We explicitly track the scaling of these spectral radii for both the source and target domains, denoted respectively by $\pp:=\nrm{\Tht_{s,\infty}}{}$ and $\pt:=\nrm{\Tht_{t,\infty}}{}$.

\section{Additional Theorems whose corollaries are in main}
This section presents the theorems whose corollaries are presented in the theory section of the main text.
\begin{theorem}\label{thm:lfm-sp-hm-err}
    Assume that $\max(\rp,\sop,\sxp,\sep,\sot) = \cO(1)$, $\log{p_1}\lesssim\nt$, and $\np\gtrsim{p_1}$. 
    \begin{enumerate}
        \item Further assume that $\nrm{\dlt_1\st}{1}\sqrt{\frac{\log{p_1}}{\nt}}=\cO\bxs{\set+(\sxt+\rt)\nrm{\beta_2\st}{}}.$
        Then, with probability at least $1-c\exp(-c'\log{p_1})$,
        \begin{align*}
            \nrm{\h\beta_{1\fm}-\beta_1\st}{}^2\lesssim\lrp{\nrm{\omg_2\st}{}^2+\frac{{p_1}}{\np}}\bxs{1\vee\frac{\nrm{\omg_2\st}{}^2+\frac{{p_1}}{\np}}{\brcs{\set+(\sxt+\rt)\nrm{\beta_2\st}{}}^2}} \\
            +\nrm{\dlt_1\st}{1}\bxs{\nrm{\dlt_1\st}{1}\wedge\brcs{\set+(\sxt+\rt)\nrm{\beta_2\st}{}}\sqrt{\frac{\log{p_1}}{\nt}}}.
        \end{align*}
        \item Further assume that $\dlt\st_{1\fm}:=\dlt_1\st+\Pt\dlt_2\st$ satisfies
        \[\nrm{\dst_{1\fm}}{1}\sqrt{\frac{\log{p_1}}{\nt}}=\cO\bxs{\sqrt{\frac{\nt}{\log p_1}}\nrm{\Sgot\Dlt_\bP\omg_2\st}{\infty}+\set+\sxt\nrm{\beta_2\st}{}+\nrm{\Dlt_\bP\omg_2\st}{}}.\]
        Then, with probability at least $1-c\exp(-c'\log{p_1})$,
        \begin{align*}
            &\nrm{\h\beta_{1\fm}-\beta_{1\fm}\st}{}^2  \lesssim\lrp{\nrm{\omg_2\st}{}\vee1}^2\frac{{p_1}}{\np}\bxs{1\vee\frac{\lrp{\nrm{\omg_2\st}{}\vee1}^2\frac{{p_1}}{\np}}{\nrm{\Sgot\Dlt_\bP\omg_2\st}{\infty}^2+\lrp{\set+\sxt\nrm{\beta_2\st}{}+\nrm{\Dlt_\bP\omg_2\st}{}}^2}} \\
            &~~+\nrm{\dlt\st_{1\fm}}{1}\bxs{\nrm{\dlt\st_{1\fm}}{1}\wedge\brcs{\nrm{\Sgot\Dlt_\bP\omg_2\st}{\infty}+\lrp{\set+\sxt\nrm{\beta_2\st}{}+\nrm{\Dlt_\bP\omg_2\st}{}}\sqrt{\frac{\log{p_1}}{\nt}}}}.
        \end{align*}
    \end{enumerate}
\end{theorem}

\section{Simulation setup in section \ref{sec:sim}}\label{sec:sim-set}
For each source domain $k=1,\dots,K$, we generate $\np\sk$ independent copies of the matched features $\xp\sk \sim \cN_{p_1}(\bo_{p_1}, \bI_{p_1})$ to form
$\Xp\sk \in \R^{\np\sk \times {p_1}}$. The rows of $\Xip\sk \in \R^{\np\sk \times {p_2}}$ are independently sampled from $\cN_{p_2}(\bo_{p_2}, \bI_{p_2})$.
To introduce explicit covariate shift, the target design matrix $\Xt \in \R^{\nt \times {p_1}}$ with i.i.d.\ entries from $\operatorname{Uniform}(0,\sqrt{12})$, and the rows of $\Xit \in \R^{\nt \times {p_2}}$ i.i.d.\ from $\cN_{p_2}(\bo_{p_2},\bI_{p_2})$. The test set with $n_t^\fn$ data points $(\Xt^\fn, \Xi_t^\fn)$ is generated using the same scheme. This setup induces heterogeneous distributions for the matched features $\bx$ across the source and target domains.

We assume in Section \ref{sec:sim-lfm} that all source domains are informative ($k\in A$) and can be transferred. We fix the entries of $\Pt$ after sampling independently from $10\cdot\operatorname{Beta}(10,10)$ distribution. To simulate structural mapping discrepancies, each source map $\Pp\sk$ is constructed by corrupting $\Pt$ entrywise with i.i.d. $\tfrac{1}{3}\left\{ \operatorname{Beta}(4,4) - \tfrac{1}{2} \right\}$ noise. Given these feature maps, the target and test mismatched features are generated as $\Zt = \Xt\Pt + \Xit$, while the source mismatched features are set to $\Zp\sk = \Xp\sk \Pp\sk + \Xip\sk$.

The response variables are generated using a bivariate normal error with unit variance and a cross-domain correlation of $0.5$.
The true target regression coefficient is a block-sparse vector $\beta=(1,0,\dots,0)'\in\R^{p/m}$ with $m = \lfloor 0.12 p \rfloor$.
The first $p_1$ entries correspond to the matched parameter ${\beta}_1\st$, and the remainder to ${\beta}_2\st$.
\begin{align*}
    \by_s\sk &= \Dp\sk\lrp{\beta\st - \delta\skt} + \eps_s\sk,~k=1,\dots,K\\ \by_t &= \Dt\beta\st + \eps_t, \quad ~\by_t^\fn = \bD_t^\fn\beta\st + \eps_t^\fn. 
\end{align*}
We evaluate the impact of domain discrepancies. Consider both sparse and dense regimes for $\delta_l\skt$, $l=1,2$. In the sparse case, we generate $\delta\skt_{lj}=C_j b_{lj}$ for $j=1,\dots,p_l$, where $C_1,\dots,C_{p_l}\overset{\mathrm{i.i.d.}}{\sim}\cN(0,0.05^2)$ and $(b_{l1},\dots,b_{lp_l})'$ is obtained by drawing without replacement from $\bdb_l=(\bon_{s_0}',\bo_{p_l-S_0}')'$ with $s_0=\lfloor \sqrt{p/2}\rfloor$. In the dense case, we take $\dst=(C_1,\dots,C_p)'$.

Under non-linear feature mapping in the section \ref{sec:sim-nlfm}, we generate the target matched features with i.i.d. coordinates $\Xt \in \mathbb{R}^{n_t \times p_1} \sim \operatorname{Uniform}(-2, 2)$. The non-linear mapping operates on an active subset of indices ${S_1} = \{1, 2, 3, 4, 5\}$, defining the coordinate-wise mapping:
\[h_j(\bx) = \sum_{i\in {S_1}}\brcs{(0.5 - |x_i-0.5|)\bb I_{\brcs{i\text{ is odd}}} + e^{-x_i}\bb I_{\brcs{i\text{ is even}}}}\]
for $j=1,\dots,p_2$. Cross-domain structural discrepancy is introduced via periodic oscillations: the source design is constructed as $\Zp\sk = \bH_s\sk(\Xp\sk) + \Xip\sk$ with $h_{s,j}\sk(\bx) := h_j(\bx) + \sin(h_j(\bx))$, while the target domain follows $\Zt = \bH(\Xt) + \Xit$. Error distributions and outcome coefficients mirror the previous DGP.

Section \ref{sec:sim-sand} evaluates the finite-sample selection accuracy of the proposed SAND algorithm and the downstream performance of SAND-HTL via 300 independent Monte Carlo replications. Informative source domain generation mirrors that in Section \ref{sec:sim-lfm}. To evaluate robustness against negative transfer, we introduce adversarial source domains ($k \notin A$) constructed as; for the sparse $\dlt\skt$ case, the contrast $C_{j}$ is drawn from $\mathcal{N}(-0.8, 0.1^2)$ and multiplied by a sparse binary mask $\bdb_l=(\bon_{\lfloor p/2\rfloor\wedge2s_0}',0,...,0)'_{p_l\times1}$, while for the dense case, it is drawn from $\mathcal{N}(-0.8, 0.5^2)$. $\Pp\sk$ is perturbed from $\Pt$ by adding a noise matrix with entries drawn from $\operatorname{Beta}(4,4)$.
The simulation parameters are varied along one axis at a time while holding the others at $(K,\np,\nt,{p_1},{p_2})=(4,400,30,30,30)$, unless noted otherwise on the horizontal axis.
When $K$ is grown over $\{4,8,16,32\}$, the number of adversarial sources is set to $K_{\rm adv}=\lfloor K^{2/3}\rfloor$ (so two adversarial sources when $K=4$).
For the $K$-sweep panels, $\nt,\np,{p_1},{p_2}$ remain at their baseline values; for the $n_t$-sweep panels, $K=4$ with two adversarial sources; and analogous fixed settings are used for the ${p_1}$ and ${p_2}$ sweeps.

\section{When multiple source domains are available}\label{sec:mp}
Now suppose we have $K$ source studies that are informative for a predictive task on the target domain. We study this setup separately from the case of $K=1$ because there are additional choices to make regarding the problem.

We pose the following DGPes for the $K$ source domains, $\brcs{\Xp\sk,\by_s\sk}$ for $k\in[K]$ and $K\ge1$:
\[Y_{s}\sk = {\xp\sk}\,'\omg\skt_1 + {\zp\sk}\,'\omg\skt_2 + \eps_s\sk,\]
while the target domain has the same data generating process as in the single-source case.
We observe $\np^{(1)}\ge\dots\ge \np^{(K)}$ sample points across $K$ source domains along with $\nt$ sample points in the target domain.
Let $\np=\sum_{k=1}^K\np\sk$ denote the gross source domain sample size.
With this augmented number of studies, we are now leveraging information from multiple source datasets.
Notations $\Xp\sk,\Zp\sk$ will stand for the matched and mismatched design matrices for each source domain $k$.
The (unknown) feature mapping between them is denoted by $\bH_s\sk$.
Note that the feature map $\bH_{s}\sk$ is allowed to vary across $k$,
reflecting heterogeneous relationships.
Then, we specify a similar process for mapping $\bX$ to $\bZ$ across multiple source domains. i.e.,

\[\Zp\sk=\bH_s\sk(\Xp\sk)+\Xi_s\sk,  \quad  k\in[K],\quad\bH_s\sk(\bX):=\E(\Zp\sk\vert\bX).
\]

The differences in covariate effects are assumed to be sparse in the following sense. Let $\dlt_l\skt = \beta\st_l - \omg\skt_l$.
We write $\bar\dlt\st:=\frac{1}{K}\sum_{k=1}^K\dlt\skt$ and $\bar\omg\st:=\frac{1}{K}\sum_{k=1}^K\omg\skt$ to aggregate the parameters across $K$ domains.
Under this notation, we have $\bar\dlt\st=\beta\st-\bar\omg\st$, where we recall that $\beta\st$ is the parameter vector in the target domain. 

\subsection{Assumptions under multiple source domains}
\paragraph{When $\bH(\bX)=\bX\bP$.} 
In the case where we match the source and target features by linear relationships, we assume that the zero-mean covariates $\xp\sk\in\R^{{p_1}}$, which are common for both domains, are $\sgm_x\sk$-sub-Gaussian random vectors for $\sgm_x\sk>0$. i.e., $\xp\sk\sim SG(\sgm_x\sk)$ identically over $k\in[K]$.
Accordingly, we assume that $\Xp\sk$ is a row-wise $\sgm_x\sk$-sub-Gaussian ensemble with $n$ independent rows and ${p_1}$ columns.
For the variance proxies of the design components, $\xp\sk$ for example, we write $\sop\so\ge...\ge\sop\sK$.
In the same ordering, we also write $\np\so\ge...\ge\np\sK$ to denote the source sample sizes. We assume Assumption \ref{asm:dsgn} analogously for each source domain $k$.
Finally, the feature mappings will be denoted by $\bH_s\sk(\Xp\sk)=\Xp\sk\Pp\sk$ for each source domain $k$.

The following assumption ensures that the feature map magnitudes are bounded uniformly across all source domains.
\begin{assumption}\label{asm:lfm-mp}
    There exists a global constant $C_\bP>0$ such that
    \[\max_{(i,j)\in[p_1]\times[p_2]}\max_{k\in[K]}|(\Pp\sk)_{ij}|\le C_\bP.\]
\end{assumption}
Finally, we guarantee the restricted strong convexity for the loss $\h\dlt-\bar\dlt\st$ rotated with respect to $\tDt=\Xt[\bI_{p_1},\h\Pp]$. For this, let us denote $S_\dlt:=\brcs{i\in[p];\bar\dlt\st_i\ne0}$ and assume:
\begin{assumption}\label{asm:re-mp}
    For some $K\in\bb N$, there exists a constant $c_\bP>0$ such that
    \[\nrm{\bv_1+\bar\bP_s\bv_2}{}\ge c_\bP\nrm{\bv}{}~~\forall\bv\in\brcs{\bv\in\R^p;\nrm{\bv_{S_\dlt^c}}{1}\le3\nrm{\bv_{S_\dlt}}{1}}.\]
\end{assumption}
Note that we work under the events $\nrm{\h\Pp-\bar\bP_s}{}=o_\pr(1)$ that is established in Lemma \ref{lem:lfm-err}.
\paragraph{When $\bH(\bX)\ne\bX\bP$.} If nonlinear feature relationships are specified, we assume that $\xt,\xp\so,\ldots,\xp\sK$ are distributed according to $\operatorname{Uniform}(\cD_x)$ distribution, since the case of heterogeneous supports can be accommodated by appropriate location and scale transformations. 
Then it is left to us to formalize a class of functions for non-linear feature relationships $\bh:\cD_x\to\R^{p_2}$, denoted by $\cH$, when multiple source populations emerge. For $K$ sources, we can now write $\bh_s\so,...,\bh_s\sK$ to denote unknown heterogeneous feature mappings, and $\bh_t$ to denote the feature mapping of the target domain. For simplicity, we assume that $\bh_s\so,...,\bh_s\sK$ all satisfy the assumptions stated in section \ref{sec:prf-thm-err}.

Sieve estimation will apply in the same manner, even under multiple source domains, as
\begin{align}\label{eq:tht-exp-mp}
    h_{s,j}\sk(\bx)=h_{s,j}\skt(\bx_{S_1\sk})=\sum_{m=1}^\infty\psi_m(\bx)\tht_{s,mj}\sk\quad\text{for}\quad\brcs{\tht_{s,mj}\sk}\in\R,\bx\in\R^{p_1},
\end{align}
i.e., $\bh_s\sk(\bx)=\sum_{m\ge1}\psi_m(\bx)\btht_{s,m}\sk$ for $\btht_{s,m}\sk:=(\tht_{s,m1}\sk,...,\tht_{s,m{p_2}}\sk)'$.
We can also write the expansion for $\bH_s\sk(\bX)$ as
\[\bH_s\sk(\bX)=\Psi_\infty(\bX)\Tht_\infty\sk\]
where $\Psi_\infty(\bX)_{im}:=\psi_m(\bx_i\sk)$ and $(\Tht_\infty\sk)_{mj}:=\tht_{s,mj}\sk$ for $m=1,2,...$ and $j\in[p_2]$ given $n\sk$ covariate samples $\bx_1\sk,...,\bx_{n\sk}\sk$. This formulation can be summarized as
\[
\bH_s\sk(\Xp\sk)=\Psi_\infty(\Xp\sk)\Tht_{s,\infty}\sk,~~k\in[K].
\]
As in the $K=1$ case, $\pp\sk,\pt$ denote $\nrm{\Tht_{s,\infty}\sk}{},\nrm{\Tht_{t,\infty}}{}$, respectively.

\begin{assumption}[Multiple active feature sets]\label{asm:active-dimension-mp}
    For each $k\le K$, there exists a $p_1\st$-variate vector function $\bh_s\skt:\R^{p_1\st}\to\R^{p_2}$ for some fixed $p_1\st\le{p_1}$ and a set of indices $S_1\sk\subset[{p_1}]$ with $\max_{k\le K}|S_1\sk|\le p_1\st$ such that for any $\bu\in\cD_x$, we have $\bh_s\sk(\bu)=\bh_s\skt(\bu_{S_1\sk})$. We assume the existence of such $\bh_t\st:\R^{\post}\to\R^{p_2}$ with $S_1\subset[p_1]$ in the target domain as well.
\end{assumption}
For the target domain, we can use $p_{1,t}\st, \bh_t\st$, and $A_t$ to denote the counterparts of the quantities in Assumption \ref{asm:active-dimension-mp}.
\begin{assumption}\label{asm:smooth-h-mp}
    $h_{t,j}\st,h_{s,j}\skt$ uniformly obey dominating mixed smoothness 1:
    \[\max_{j\le{p_2}}\sum_{\ba\in\cI\st}\nrm{{\rm D}^\ba h_{t,j}\st}{L_2(\cD_x\st)}^2\vee\max_{j\le{p_2},\,k\le K}\sum_{\ba\in\cI\st}\nrm{{\rm D}^\ba h_{s,j}\skt}{L_2(\cD_x\st)}^2<\infty.\]
\end{assumption}
This guarantees that a common sieve basis $\psi_m$ can approximate all feature relationship functions at comparable rates.

Naturally, there exists some constant variance proxy $\sgm_\psi>0$ such that the basis-expansions $\Psi_M(\Xt),\Psi_M(\Xp\so),...,\Psi_M(\Xp\sK)$ are $\sgm_\psi$-sub-Gaussian ensembles. Furthermore, we notice that $\operatorname{Var}[\bh_s\sk(\xp\sk)]={\Tht_{s,\infty}\sk}'\Tht_{s,\infty}\sk$.
Assumption \ref{asm:smooth-h-mp} again implies that the magnitude of $\Tht_{s,\infty}\sk$'s uniformly obey
\[\max_{j\le{p_2}}\sum_{m\ge1}{\tht_{t,mj}}^2+\max_{j\le{p_2},\,k\le K}\sum_{m\ge1}{\tht_{s,mj}\sk}^2<\infty.\]
\begin{assumption}\label{asm:orthgn-mp}
    Assume that there exists some constant $C_H>0$ such that
    $$\bu'\sum_{m \ge 1}\lrp{\frac{m}{\log^{\post-1} m}}^2\btht_{s,m}\sk{\btht_{s,m}\sk}'\bu \preceq C_H(\pp\sk)^2,~ \bu'\sum_{m \ge 1}\lrp{\frac{m}{\log^{\post-1} m}}^2\btht_{t,m}\btht_{t,m}'\bu \preceq C_H\pt^2$$
    for all unit vectors $\bu\in\R^{p_2}$ and $k\in[K]$.
\end{assumption}

\subsection{HTL with multiple source domains}
Now, we are ready to specify the following two-stage estimator:
\begin{enumerate}[{\bf Stage} \bf\Roman*.]
    \item Imputation:
    Let $\h\Pp:=\frac{1}{K}\sum_{k\le K}\h{\Pp\sk}$ and $\h\Tht_{s,M}:=\frac{1}{K}\sum_{k\le K}\h\Tht_{s,M}\sk$. Then,
    \begin{align*}
        \h\bZ_t=\begin{cases}
            \Xt\h\Pp,&\text{ when }\bh_t(\bx)=\Pt'\bx\\
            \Psi_M(\Xt)\h\Tht_{s,M},&\text{ when }\bh_t\in\cH
        \end{cases}
    \end{align*}
    \item Estimation:
    \begin{enumerate}[1)]
        \item $\h\omg=\frac{1}{K}\sum_{k=1}^K\h\omg\sk\text{ where }\h\omg\sk:={\arg\min}_\omg\,\frac{1}{\np\sk}\lrnrm{\by_{s}\sk-\Xp\sk\omg_1-\Zp\sk\omg_2}{}^2$
        \item $\h\beta = \arg\min_{\beta}\brcs{\frac{1}{\nt}\nrm{\by_{t}-\Xt\beta_1-\hZt{\beta}_{2}}{}^2 + \sum_{l=1}^2\lmd_l\nrm{\beta_l - \h\omg_l}{1}}$.
    \end{enumerate}
\end{enumerate}
This procedure uses $\Xt$ and the average of the $K$ source feature map estimators to approximate $\hZt$.
In contrast, a homogeneous transfer learning strategy with $K$-source domains will perform the following under our regime:
\begin{enumerate}[1)]
    \item $\h\omg_{1\fm} = \frac{1}{K}\sum_{k=1}^K\h\omg_{1\fm}\sk\text{ where }\h\omg_{1\fm}\sk:=\arg\min_{\omg_1}\frac{1}{\np\sk}\lrnrm{\by_{s}\sk-\Xp\sk\omg_1}{}^2$
    \item $\h\beta_1^{\rm hm} = \arg\min_{\beta_1}\brcs{\frac{1}{\nt}\nrm{\by_{t}-\Xt\beta_1}{}^2 + \lmd_1\nrm{\beta_1 - \h\omg_{1\fm}}{1}}$.
\end{enumerate}
We note that the first step in stage 2 of our HTL estimator (and the first step in our comparison homogeneous estimator) differs slightly from the data-pooling approach in \cite{li2022transfer,tian2022transfer}. Instead of pooling data from the source studies first and then obtaining a minimizer of $\omg$ in the pooled data, we separately obtain minimizers of $\omg$s in the source datasets and then average them to obtain the final estimate of $\omg$. In Remark \ref{rmk_multi} in the Appendix, we show both approaches are asymptotically equivalent. A similar observation has also been made in recent work \cite{he2024transfusion}, which makes a similar choice for its transfusion algorithm to make the method robust to covariate shift.

\subsection{Results under linear feature map}\label{sec:lfm-mp}
Recycling notations, we write $\nrm{\bar\Dlt_\bP}{}:=\frac{1}{K}\nrm{\sum_{k\le K}(\Pt-\Pp\sk)}{}$.
Write the maximum singular value of $\frac{1}{K}\sum_{k\le K}\Pp\sk\in\R^{{p_1}\times {p_2}}$ as $\bar\rp$. The following theorem provides high-probability bounds on the estimation and prediction errors of the HTL estimator in the multiple-source case. Let us introduce the following notations:
$$\lmd_{\bP,F}^2:=\sum_{k=1}^K\frac{s_\bP\sk\log{p_1}}{K\np\sk},~q_j\sk:=\nrm{(\Pp\sk)_{\cdot j}}{0},~\bar m_\bP:=\frac1K\sum_{k=1}^K\max_{j\le p_2}q_j\sk.$$
Then, we define $\t m_\bP:=\max_{k\le K}\max_{j\le p_2}q_j\sk$. $\lmd_{\bP,F}$ denotes the rate of convergence for the feature map estimator $\h\Pp$ in Frobenius norm. $q_j\sk$ denotes the sparsity of the $j$-th column of $\Pp\sk$, i.e., the number of relevant covariates in $\xp$ for predicting $\zp$. $\bar m_\bP$ denotes the average maximum-column-sparsity over all $K$ source domains. Finally, $\t m_\bP$ denotes the global maximum column sparsity of the source feature maps.

Note that $\bar m_\bP\le\t m_\bP$ and $\nrm{\bar\bP_s}{1}\le\max_{k\le K}\nrm{\Pp\sk}{1}$. Therefore, under Assumption \ref{asm:lfm-mp}, then if follows that $\nrm{\bar\bP_s}{1}=\cO(\t m_\bP)$.
\begin{theorem}[HTL with Linear Feature Map]\label{thm:lfm-err}
    Assume that $\max_{k\le K}\max(\rp\sk,\sop\sk,\sxp\sk,\sep\sk)+\sot = \cO(1)$, $\log p\lesssim\nt$, and $\np\sk\gtrsim (\rp\sk)^2p+s_\bP\sk\log{p_1}$ for all $k\le K$. Further assume that
    \begin{align}\label{eq:lfm-err}
        \t m_\bP^2\nrm{\bar\dlt\st}{1}\sqrt{\frac{\log p_1}{\nt}}=\cO\bxs{\set+\nrm{\beta_2\st}{}(\nrm{\bar\Dlt_\bP}{}+\sxt)}.
    \end{align}
    Then, under Assumptions \ref{asm:re-mp} and \ref{asm:lfm-mp}, with the same probability bounds as in Theorem \ref{thm:lfm-sp-err},
    \begin{align*}
        \nrm{\h\beta-\beta\st}{}^2&\lesssim\sum_{k=1}^K\frac{p}{K\np\sk}\bxs{1\vee\frac{\lrp{\bar\rp+\lmd_{\bP,F}}^{-2}\sum_{k=1}^K\frac{\t m_\bP^2p}{K\np\sk}}{\brcs{\set+\nrm{\beta_2\st}{}\lrp{\nrm{\bar\Dlt_\bP}{}+\sxt}}^2}} \\
        &\qquad+\nrm{\bar\dlt\st}{1}\bxs{\nrm{\bar\dlt\st}{1}\wedge\lrp{1+\lmd_{\bP,F}}\brcs{\set+\nrm{\beta_2\st}{}\lrp{\nrm{\bar\Dlt_\bP}{}+\sxt}}\sqrt{\frac{\log p_1}{\nt}}}
    \end{align*}
    and
    $$\frac{1}{n_t^\fn}\lrnrm{\E(\by_t^\fn|\Xt^\fn)-\tDt^\fn\h\beta}{}^2\lesssim \nrm{\bar\Dlt_\bP}{}^2\nrm{\beta_2\st}{}^2+\nrm{\h\beta-\beta\st}{}^2.$$
\end{theorem}
The proof is given in Section \ref{sec:prf-lfm-thm-err}. The next theorem is the result of our baseline HmTL method.
\begin{theorem}\label{thm:lfm-hm-err}
    Assume that $\max_{k\le K}\max(\rp\sk,\sop\sk,\sxp\sk,\sep\sk)+\sot = \cO(1)$, $\log{p_1}\lesssim\nt$, and $\np\sk\gtrsim{p_1}$ for all $k\le K$. 
    \begin{enumerate}
        \item Further assume that $\nrm{\bar\dlt_1\st}{1}\sqrt{\frac{\log p_1}{\nt}}=\cO\bxs{\set+(\sxt+\rt)\nrm{\beta_2\st}{}}.$
        Then, with the same probability bounds as in Theorem \ref{thm:lfm-sp-hm-err},
        \begin{align*}
            \nrm{\h\beta_{1\fm}-\beta_1\st}{}^2\lesssim\frac{1}{K}\sum_{k=1}^K\lrp{\nrm{\omg_2\skt}{}^2+p_1/\np\sk}\bxs{1\vee\frac{\frac{1}{K}\sum_{k=1}^K\lrp{\nrm{\omg_2\skt}{}^2+p_1/\np\sk}}{\brcs{\set+(\sxt+\rt)\nrm{\beta_2\st}{}}^2}} \\
            +\nrm{\bar\dlt_1\st}{1}\bxs{\nrm{\bar\dlt_1\st}{1}\wedge\brcs{\set+
            (\sxt+\rt)\nrm{\beta_2\st}{}}\sqrt{\frac{\log{p_1}}{\nt}}}.
        \end{align*}
    \item Let $\overline{\Dlt_\bP\omg_2\st}:=\frac{1}{K}\sum_{k=1}^K\Dlt_\bP\sk\omg_2\skt$. Further assume that $\bar\dlt_{1\fm}\st:=\bar\dlt_1\st+\Pt\bar\dlt_2\st$ satisfies $\nrm{\bar\dlt_{1\fm}\st}{1}\sqrt{\frac{\log{p_1}}{\nt}}=\cO\bxs{\sqrt{\frac{\nt}{\log{p_1}}}\nrm{\Sgot\overline{\Dlt_\bP\omg_2\st}}{\infty}+\set+\sxt\nrm{\beta_2\st}{}+\nrm{\overline{\Dlt_\bP\omg_2\st}}{}}.$ Then, with the same probability bounds as in Theorem \ref{thm:lfm-sp-hm-err},
    \begin{align*}
        &\nrm{\h\beta_{1\fm}-\beta_{1\fm}\st}{}^2\\
        &~ \lesssim\sum_{k=1}^K\frac{p_1(\nrm{\omg_2\skt}{}^2+1){}}{K\np\sk}\bxs{1\vee\frac{\sum_{k=1}^K\frac{p_1(\nrm{\omg_2\skt}{}^2+1){}}{K\np\sk}}{\nrm{\Sgot\overline{\Dlt_\bP\omg_2\st}}{\infty}^2+\lrp{\set+\sxt\nrm{\beta_2\st}{}+\nrm{\overline{\Dlt_\bP\omg_2\st}}{}}^2}} \\
        &~~+\nrm{\bar\dlt_{1\fm}\st}{1}\bxs{\nrm{\bar\dlt_{1\fm}\st}{1}\wedge\brcs{\nrm{\Sgot\overline{\Dlt_\bP\omg_2\st}}{\infty}+\lrp{\set+\sxt\nrm{\beta_2\st}{}+\nrm{\overline{\Dlt_\bP\omg_2\st}}{}}\sqrt{\frac{\log{p_1}}{\nt}}}}
    \end{align*}
    and $\frac{1}{n_t^\fn}\lrnrm{\E(\by_t^\fn|\Xt^\fn)-\Xt^\fn\h\beta_{1\fm}}{}^2\lesssim\nrm{\h\beta_{1\fm}-\beta_{1\fm}\st}{}^2+\rt^2\nrm{\beta_2\st}{}^2$.
    \end{enumerate}
\end{theorem}
\subsection{Results under $\bh\in\cH$}
Now, we write $\nrm{\bar\Dlt_\Tht}{}=\frac{1}{K}\lrnrm{\sum_{k\le K}(\Tht_{t,M}-\Tht_{s,M}\sk)}{}$ where $\Tht_{t,M},\Tht_{s,M}\sk$ denote the truncated Fourier coefficients in section \ref{sec:nlfm} at $m=M$, each from the target and source domains, respectively. Denote the source feature map estimation error rate by $\lmd_{R_\Tht}:=\frac{1}{K}\sum_{k=1}^K\lrp{{p_2}^{3/4}\log^{3/2}{p_2}\frac{\log^{\post-1}\np\sk}{{\np\sk}}}^{2/3}$.
\begin{theorem}\label{thm:err1}
    Assume that $\max_{k\le K}\max(\pp\sk,\sop\sk,\sxp\sk,\sep\sk)+\sot = \cO(1)$, $\np\sk\gtrsim (\pp\sk)^2p$ for all $k\le K$, $M\asymp\sqrt{{p_2}^3(\log p_2)^{2\post+1}}$, and $\log p\lesssim\nt$.
    Further assume that
    \[\nrm{\bar\dlt\st}{1}\sqrt{\frac{\log p}{n_t}}=\cO\bxs{\set+\nrm{\beta_2\st}{}(\nrm{\bar\Dlt_\Tht}{}+\sxt+\lmd_{R_\Tht})}.\]
    Then, with the same probability bounds as in Theorem \ref{thm:sp-err},
    \begin{align*}
        \nrm{\h\beta-\beta\st}{}^2\,\lesssim&\,
        \frac{1}{K}\sum_k\frac{p}{\np\sk}\cdot\cO\bxs{1\vee\frac{\frac{1}{K}\sum_kp/\np\sk}{\brcs{\set+(\sxt+\nrm{\bar\Dlt_\Tht}{}+\lmd_{R_\Tht})\nrm{\beta_2\st}{}}^2}} \\
        &\quad+\nrm{\bar\dlt\st}{1}\bxs{\nrm{\bar\dlt\st}{1}\wedge\brcs{\set+(\sxt+\nrm{\bar\Dlt_\Tht}{}+\lmd_{R_\Tht})\nrm{\beta_2\st}{}}\sqrt{\frac{\log p}{\nt}}}
    \end{align*}
    and
    \[\frac{1}{n_t^\fn}\lrnrm{\E(\by_t^\fn|\Xt^\fn)-\tDt^\fn\h\beta}{}^2\lesssim
    \nrm{\bar\Dlt_\Tht}{}^2\nrm{\beta_2\st}{}^2+\nrm{\h\beta-\beta\st}{}^2.\]
\end{theorem}
Notice that $\lambda_{R_\Tht}=o(1)$ under $n\sk\gg p_2^{3/4}\log^{\post+0.5}p_2$.

To compare with the homogeneous transfer learning strategy, we obtain the following bound on the estimator obtained from the homogeneous method.
\begin{theorem}\label{thm:hm-err}
    Assume that $\max_{k\le K}\max(\pp\sk,\sop\sk,\sxp\sk,\sep\sk)+\sot = \cO(1)$, $\log{p_1}\lesssim\nt$, and $\np\sk\gtrsim{p_1}$ for all $k\le K$. Further assume that
    \[\nrm{\bar\dlt_1\st}{1}\sqrt{\frac{\log{p_1}}{\nt}}=\cO\bxs{\set+(\sxt+\pt)\nrm{\beta_2\st}{}}.\]
    Then, with the same probability bounds as in Theorem \ref{thm:sp-hm-err},
    \begin{align*}
        \nrm{\h\beta_{1\fm}-\beta_1\st}{}^2\lesssim\frac1K\sum_{k=1}^K(\nrm{\omg_2\skt}{}^2+p_1/\np\sk)\bxs{1+\frac{\frac1K\sum_{k=1}^K(\nrm{\omg_2\skt}{}^2+p_1/\np\sk)}{\brcs{\set+(\sxt+\pt)\nrm{\beta_2\st}{}}^2}}\\
        +\nrm{\bar\dlt_1\st}{1}\bxs{\nrm{\bar\dlt_1\st}{1}\wedge\brcs{\set+(\sxt+\pt)\nrm{\beta_2\st}{}}\sqrt{\frac{\log{p_1}}{\nt}}}
    \end{align*}
    and
    \begin{align*}
        \pr\bxs{\frac{1}{n_t^\fn}\lrnrm{\E(\by_t^\fn|\Xt^\fn)-\Xt^\fn\h\beta_{1\fm}}{}^2
        \lesssim \pt^2\nrm{\beta_2\st}{}^2 + \nrm{\h\beta_{1\fm}-\beta_1\st}{}^2}\>1-c\exp(-c'n_t^\fn)-c\log^{-1/2}p_2.
    \end{align*}
\end{theorem}

\section{Preliminary Results}\label{sec:apdx}
\subsection{Restricted Strong Convexity}
We verify the non-trivial minimum eigenvalues of $\Var(\bd)$, $\Var(\t\bd)$ to apply the RSC lemma.
\begin{lemma}[\citealp{agarwal2012fast}]
    Suppose that an $n\times p$ matrix $\bX$ with $\E\bX=0$ is a $\sgm$-sub-Gaussian ensemble and let $\Sgm$ denote its covariance. Denote the minimum and maximum eigenvalues of $\Sgm$ by $\lmd_0$ and $\lmd_1$, respectively. If $\lmd_0,\lmd_1$ lie between positive absolute constants, then for any $\bv\in\R^p$ with probability at least $1-\exp\lrp{-cn}$,
    \begin{align*}
        \frac{1}{n}\nrm{\bX\bv}{}^2&\ge\frac{\lmd_0\nrm{\bv}{}^2}{2}-c'\frac{\sgm^2\log p}{n}\nrm{\bv}{1}^2, \\
        \frac{1}{n}\nrm{\bX\bv}{}^2&\le2\lmd_1\nrm{\bv}{}^2+c'\frac{\sgm^2\log p}{n}\nrm{\bv}{1}^2
    \end{align*}
    for some fixed $c,c'>0$.
\end{lemma}
\paragraph{Control of $\frac{1}{n}\nrm{\bD\bv}{}^2$.}
Let $\bd=(\bx',\bz')'\in\R^p$ and $\t\bd=(\bx',\h\bz')'$ to denote arbitrary row of $\bD$ and $\t\bD$, respectively.
Here, we mean by $\bx,\bz$ general matched and mismatched features, whose relationships are linear or non-linear, depending on the formulations discussed in the main text.
Let $\Gmm\in\brcs{\bP,\Tht_\infty}$ denote the choice of a feature mapper.
Let $\t\bx$ be $\bx$ if $\Gmm=\bP$ so that $\bz=\bP'\t\bx+\xi$. If $\Gmm=\Tht_\infty$, let $\t\bx=\b\psi_\infty(\bx)\in\R^{\infty}$ hence $\bz=\Tht_\infty'\t\bx+\xi$.

First,
\[
\Sgm:=\operatorname{Var}(\bd)=\begin{bmatrix}
\E(\bx\bx') & \E(\bx\t\bx')\Gmm \\[3pt]
\Gmm'\E(\t\bx\bx') & \Gmm'\E(\t\bx\t\bx')\Gmm + \Sgm_\xi
\end{bmatrix}=:\begin{bmatrix}
\Sgm_x & \Sgm_{x\t x}\Gmm \\[3pt]
\Gmm'\Sgm_{\t xx} & \Gmm'\Sgm_{\t x} \Gmm + \Sgm_\xi
\end{bmatrix},
\]
we can take the Schur complement as
\[\bS:=\Gmm'(\Sgm_{\t x}-\Sgm_{\t xx}\Sgm_x^{-1}\Sgm_{x\t x})\Gmm+\Sgm_\xi.\]
Therefore, by the block LU decomposition,
\[\Sgm=\begin{bmatrix}
    \bI & (0) \\ \Gmm'\Sgm_{\t xx}\Sgm_x^{-1} & \bI
\end{bmatrix}\begin{bmatrix}
    \Sgm_x & (0) \\ (0) & \bS
\end{bmatrix}\begin{bmatrix}
    \bI & \Sgm_x^{-1}\Sgm_{x\t x}\Gmm \\ (0) & \bI
\end{bmatrix}\]
hence ${\rm det}(\Sgm)={\rm det}(\Sgm_x){\rm det}(\bS)$. Since
$\b S\succeq\Sgm_\xi\succ(0)$ and we assume throughout the theory that 
\begin{itemize}
    \item All eigenvalues of $\Sgm_{x_s},\Sgot,\Sgxp,\Sgxt$ lie between absolute positive constants.
    \item $\max(\sgm_{\t x},\sop,\sot,\sxp)=\c\cO(1)$.
\end{itemize}
Therefore, we posit that $\lmd_{\min}(\Sgm)$ is bounded away from zero by some absolute positive constant.
The sub-gaussian variance proxy of $\bd_s$ is $\sgm_d=\sgm_x+\sgm_z=\sgm_x+\nrm{\Gmm_s}{}\sgm_{\t x}+\sgm_\xi=\cO(\nrm{\Gmm_s}{}+1)$ and similarly, $\tau_d=\cO(\nrm{\Gmm_t}{}+\sxt)$.

Hence, the lower bound event of the RSC lemma becomes:
\[\frac{1}{\np}\nrm{\Dp\bv}{}^2\gtrsim\nrm{\bv}{}^2-(\nrm{\Gmm_s}{}^2+1)\frac{\log p}{\np}\nrm{\bv}{1}^2,~~
\frac{1}{\nt}\nrm{\Dt\bv}{}^2\gtrsim\nrm{\bv}{}^2-(\nrm{\Gmm_t}{}^2+\sxt^2)\frac{\log p}{\nt}\nrm{\bv}{1}^2.\]
Using elementary results, we can also easily derive that
$\lmd_{\max}(\Sgm)\lesssim\nrm{\Gmm}{}(1\vee\nrm{\Gmm}{})\vee1$
and the upper bound event reflects this:
\[
    \frac{1}{\np}\nrm{\Dp\bv}{}^2\lesssim(\nrm{\Gmm_s}{}^2+1)\lrp{\nrm{\bv}{}^2+\frac{\log p}{\np}\nrm{\bv}{1}^2},
    \frac{1}{\nt}\nrm{\Dt\bv}{}^2\lesssim(\nrm{\Gmm_t}{}^2+\sxt^2)\lrp{\nrm{\bv}{}^2+\frac{\log p}{\nt}\nrm{\bv}{1}^2}.
\]
\paragraph{Control of $\frac{1}{n}\nrm{\t\bD\bv}{}^2$.}
Next, if $\t\bd_t:=(\xt',\t\bx_t'\h\Gmm_s)'$ is given where $\h\Gmm_s$ is an estimated $\Gmm_s$ and independent with $\bx_t,\t\bx_t$, then conditionally given $\h\Gmm_s$ which is independent with $\xt$ and $\t\bx_t$,
\[
\operatorname{Var}(\t \bd_t|\h\Gmm_s)=:\Sgm_{\t d}=
\begin{bmatrix}
\Sgot & \Sgm_{x_t\t x_t}\h\Gmm_s\\
 & \h\Gmm_s'\Sgm_{\t x_t}\h\Gmm_s
\end{bmatrix}.
\]
The variance proxy is conditionally $\tau_{\t d}=\tau_x+\nrm{\h\Gmm_s}{}\tau_{\t x}=\cO(\nrm{\h\Gmm_s}{}+1)$.

Unlike the population design $\bd$, the imputed design is \emph{rank-deficient} if $\Gmm=\bP$: the imputed mismatched block $\t\bx_t'\h\Gmm_s=\xt'\h\Pp$ produces a population covariance $\Sgm_{\t d}$ with a zero Schur complement and $\lmd_{\min}(\Sgm_{\t d})=0$. 
Writing $\h\Pp^+:=[\bI_{p_1},\h\Pp]$, we have
\begin{equation}\label{eq:pred-dir}
    \frac1\nt\nrm{\tDt(\h\dlt-\bar\dlt\st)}{}^2=\frac1\nt\nrm{\Xt\h\Pp^+(\h\dlt-\bar\dlt\st)}{}^2,
\end{equation}
so the loss is flat along the null directions $\h\dlt-\bar\dlt\st\in\cN:=\brcs{\bv\in\R^p:\h\Pp^+\bv=\bo}$. 

To guarantee the identifiability of the discrepancy parameter $\dlt\st$ in the HTL, we must establish that the Restricted Eigenvalue (RE) condition holds with high probability.
Let $\cC$ denote the $\ell_1$-cone $\brcs{\bv\in\R^p:\nrm{\bv_{S_\dlt^c}}{1}\le3\nrm{\bv_{S_\dlt}}{1}}$. By RSC lemma, with probability at least $1-c\exp(-c'n_t)$,
\begin{equation}
    \frac{1}{n_t}\|\Xt\h\Pp^+\bv\|^2 \gtrsim \|\h\Pp^+\bv\|^2 - \frac{\log p_1}{n_t}\|\h\Pp^+\bv\|_1^2
\end{equation}
for $\bv\in\cC$. 
With $\Pp^+:=[\bI_{p_1},\bar\bP_s]$, since $\Pp^+(\Pp^+)'=\bI_{p_1}+\Pp\Pp'\succeq\bI_{p_1}$, every nonzero singular value of $\Pp^+$ is at lease one. 
Also, we have $\nrm{\h\Pp^+\bv}{1}^2\le(1+\nrm{\h\Pp}{1}^2)\nrm{\bv}{1}^2$.
Under  Assumption \ref{asm:re-mp}, it remains to invoke the event $\brcs{\nrm{\bar\bR_{\Pp}}{F}^2\le\lmd_{\bP,F}^2}$ for $\bar\bR_{\Pp}:=\frac{1}{K}\sum_{k\le K}(\Pp\sk-\h{\Pp\sk})$. On this event,
\[\nrm{\h\Pp^+\bv}{}\ge\nrm{\Pp^+\bv}{}-\lmd_{\bP,F}\nrm{\bv_2}{}\ge\lrp{c_\bP-\lmd_{\bP,F}}\nrm{\bv}{}.\]
Hence for $\bv\in\cC$, we have $\nrm{\h\Pp^+\bv}{}^2/\nrm{\bv}{}^2\ge c_\bP-\lmd_{\bP,F}^2$ with probability at least $1-cKp_2\exp(-c\log p_1)$ (Lemma \ref{lem:lfm-err}). This leads to
\begin{align*}
    \frac{1}{n_t}\nrm{\tDt\bv}{}^2\gtrsim(c_\bP-\lmd_{\bP,F}^2)\nrm{\bv}{}^2-(\nrm{\h\Pp}{1}^2+1)\frac{\log p_1}{n_t}\nrm{\bv}{1}^2.
\end{align*}
Similarly, for $\lmd_{\bP,1}^2:=\sum_{k=1}^K\frac{(m_\bP\sk)^2\log{p_1}}{K\np\sk}$, on the event $\brcs{\nrm{\bar\bR_{\Pp}}{1}^2\le\lmd_{\bP,1}^2}$ we have
\begin{align*}
    \frac{1}{n_t}\nrm{\tDt\bv}{}^2\gtrsim(c_\bP-\lmd_{\bP,F}^2)\nrm{\bv}{}^2-(\nrm{\bar\bP_s}{1}^2+\lmd_{\bP,1}^2+1)\frac{\log p_1}{n_t}\nrm{\bv}{1}^2.
\end{align*}
By definition, we have $m_\bP\sk\le s_\bP\sk$ for all $k$. Therefore, for $\bar m_\bP:=\sum_{k=1}^K\frac{m_\bP\sk}{K}$, we have the following: under the Assumption \ref{asm:re-mp} and $n_s\sk$ sufficiently larger than $s_\bP\sk\log p_1$ for all of $k\in[K]$, we have
\begin{align}\label{eq:rsc-lfm}
    \pr\bxs{\frac{1}{n_t}\nrm{\tDt\bv}{}^2\gtrsim\nrm{\bv}{}^2-\lrp{\nrm{\bar\bP_s}{1}^2+\bar m_\bP}\frac{\log p_1}{n_t}\nrm{\bv}{1}^2}\gtrsim1-Kp_2\exp(-c\log p_1).
\end{align}

\paragraph{Control of $\bZ$.}
On the other hand, the mismatched feature $\bz$ satisfies a sub-Gaussian tail bound.
In the linear case $\zp = \bP'\xp + \xi_s$, we have
\[
\zp \sim SG\bigl(\sgm_x\|\bP\| + \sgm_\xi \bigr),
\]
and the covariance satisfies
\[
\Var(\zp)
=
{\Pp}'\Sgm_{x_s}\Pp + \Sgm_{\xi_s},
\qquad
\lmd_{\min}(\Var(\zp))
\ge
\lmd_{\min}(\Sgm_{x_s})\,s_{\min}(\Pp)^2
+
\lmd_{\min}(\Sgm_{\xi_s})>0.
\tag{A}
\]
We now show an analogous property for the nonlinear feature map
$\bz = \bh(\bx)+\xi$ with $\bh \in \mathcal{H}$.
\begin{lemma} 
    Let $\zp = \bh_s(\xp) + \xi_s$ where $\bh \in \mathcal{H}$ is the function class defined in Section \ref{sec:nlfm}.
    Then,
    \[
    \zp \sim SG(\sgm_z), \qquad \sgm_z := \sgm_h \|\Tht_\infty\| + \sgm_{\xi s},
    \]
    where $\sgm_h$ depends only on the Sobolev-ellipsoid constant of $\mathcal{H}$ and the basis bounds.
\end{lemma}
\begin{proof}
By the cosine basis expansion \eqref{eq:tht-exp}, for any $\bu \in S^{p_2-1}$ we may write
\[
\bu' \bh_s(\xp) = \sum_{m \ge 1} \psi_m(\xp) \alpha_m, \qquad \alpha_m := \bu' \btht_m.
\]
By introducing the Sobolev smoothing weights $w_m = m / \log^{\post-1} m$, Cauchy---Schwarz yields
\[
|\bu' \bh_s(\xp)| \le \Biggl( \sum_{m \ge 1} \frac{\psi_m(\xp)^2}{w_m^2} \Biggr)^{1/2} \Biggl( \sum_{m \ge 1} w_m^2 \alpha_m^2 \Biggr)^{1/2}.
\]
Since $\sup_{\bx} |\psi_m(\bx)| \le c_\psi$ and $\sum_{m \ge 1} w_m^{-2} < \infty$, the first term is uniformly bounded by an absolute constant $C_\psi > 0$ almost surely. 

For the second term, we can express it as a quadratic form of the weighted coefficient operator:
\[
\sum_{m \ge 1} w_m^2 \alpha_m^2 = \bu' \Biggl( \sum_{m \ge 1} w_m^2 \btht_m \btht_m' \Biggr) \bu =: \bu' \bT \bu.
\]
By assumption,
\[
\bT = \sum_{m \ge 1} w_m^2 \btht_m \btht_m' \preceq C_H \sum_{m \ge 1} \btht_m \btht_m' = C_H \Tht_\infty \Tht_\infty'
\]
hence
\[
\lmd_{\max}(\bT) = \sup_{\bu \in S^{p_2-1}} \bu' \bT \bu \le C_H \|\Tht_\infty\|^2.
\]
Consequently, we have almost surely
\[
|\bu' \bh_s(\xp)| \le C_\psi \sqrt{\lmd_{\max}(\bT)} \le \sgm_h \|\Tht_\infty\|,
\]
where $\sgm_h := C_\psi \sqrt{C_H}$. Hoeffding's lemma then directly yields
\[
\mathbb{E} \exp \bigl(t\bu' \bh_s(\xp) \bigr) \le \exp \left( \frac{t^2 \sgm_h^2 \|\Tht_\infty\|^2}{2} \right)~~\forall t\in\R,
\]
establishing that $\bh_s(\xp)$ is $\sgm_h \|\Tht_\infty\|$-sub-Gaussian.
\end{proof}

\subsection{Estimation on source regression model}
\begin{lemma}\label{lem:ols-spectral}
    Let $\bX\in\R^{n\times p}$ be a row-wise mean-zero $\sgm_x$-sub-Gaussian ensemble with $\E(\bx\bx')=\Sgm$.
    Assume $\lmd_{\fm}\le\lmd_{\min}(\Sgm)$ for some $\lmd_{\fm}>0$ and $n>p$.
    Consider $\by=\bX\beta\st+\bcE$ with $\bcE=(\eps_1,\ldots,\eps_n)'$, where the $\eps_i$ are independent, mean-zero, $\sgm_\varepsilon$-sub-Gaussian, and independent of $\bX$.
    For the OLS estimator $\h\beta:=(\bX'\bX)^{-1}\bX'\by$, if $n\gtrsim\sgm_x^4 p/\lmd_{\fm}^2$, then
    \[
        \pr\bxs{\nrm{\h\beta-\beta\st}{}^2\gtrsim\frac{\sgm_x^2\sgm_\varepsilon^2 p}{\lmd_{\fm}^2n}}\le c\exp\lrp{p\log5-c'n}.
    \]
\end{lemma}
\begin{proof}
    Write $\h\Sgm:=\bX'\bX/n$.
    By \cite{wainwright2019high}, there exist absolute constants $c_1,c_2,c_3>0$ such that, for every $t\ge0$,
    \[
        \pr\lrp{\frac{\nrm{\h\Sgm-\Sgm}{}}{\sgm^2}\ge c_1\sqrt{\frac{p}{n}}+t}
        \le c_2\exp\bxs{-c_3 n\min\lrp{t,t^2}}.
    \]
    Set $t:=\lmd_{\fm}/(2\sgm^2)-c_1\sqrt{p/n}\ge0$, which holds under $n\gtrsim\sgm^4 p/\lmd_{\fm}^2$.
    Then $\lmd_{\fm}/(2\sgm^2)=c_1\sqrt{p/n}+t$, so
    \[
        \pr\lrp{\nrm{\h\Sgm-\Sgm}{}\ge\frac{\lmd_{\fm}}{2}}
        \le c_2\exp\bxs{-c_3 n\min\lrp{t,t^2}}.
    \]
    On the event $\brcs{\nrm{\h\Sgm-\Sgm}{}\le\lmd_{\fm}/2}$, Weyl's inequality and $\lmd_{\fm}\le\lmd_{\min}(\Sgm)$ give $\lmd_{\min}(\h\Sgm)\ge\lmd_{\fm}/2$ and
    \[
        \nrm{\h\beta-\beta\st}{}^2\le\frac{4}{\lmd_{\fm}^2}\lrnrm{\frac{\bX'\bcE}{n}}{}^2.
    \]
    Fix $\bu\in S^{p-1}$ and condition on $\bX$.
    Since $\bcE\perp\bX$ and the $\eps_i$ are mean-zero, $\sgm_\varepsilon$-sub-Gaussian, the variables $\eps_i\bu'\bx_i$ are conditionally independent given $\bX$, each with conditionally $(\sgm_\varepsilon\vrt{\bu'\bx_i})$-sub-Gaussian.
    Therefore, their average satisfies, for every $t>0$,
    \[
        \pr\lrp{\vrt{\frac{1}{n}\sum_{i=1}^n\eps_i\bu'\bx_i}\ge t\given \bX}
        \le 2\exp\lrp{-\frac{cnt^2}{\sgm_\varepsilon^2\nrm{\bX\bu}{}^2/n}}.
    \]
    Taking $t\asymp\sgm_\varepsilon\nrm{\bX\bu}{}/\sqrt{n}$ and integrating over $\bX$ yields, for every fixed $\bu$,
    \[
        \pr\lrp{\vrt{\frac{1}{n}\sum_{i=1}^n\eps_i\bu'\bx_i}\gtrsim\frac{\sgm_\varepsilon\nrm{\bX\bu}{}}{\sqrt{n}}}\le c_4\exp\lrp{-c_5 n}.
    \]
    Let $ N$ be a $1/2$-net of $S^{p-1}$ with $| N|\le5^p$.
    For any $\bu\in S^{p-1}$, pick $\bv\in N$ with $\nrm{\bu-\bv}{}\le1/2$; then
    \[
        \vrt{\langle\bX\bu,\bcE\rangle}
        \le \vrt{\langle\bX\bv,\bcE\rangle}+\vrt{\langle\bX(\bu-\bv),\bcE\rangle}{}
        \le \vrt{\langle\bX\bv,\bcE\rangle}+\frac{1}{2}\nrm{\bX'\bcE}{},
    \]
    so $\nrm{\bX'\bcE/n}{}\le 2\max_{\bv\in N}\vrt{\langle\bX\bv,\bcE\rangle/n}$.
    A union bound over $\bu\in N$ gives
    \[
        \max_{\bu\in N}\vrt{\frac{\langle\bX\bu,\bcE\rangle}{n}}
        \lesssim\frac{\sgm_\varepsilon}{\sqrt{n}}\max_{\bu\in N}\nrm{\bX\bu}{}
    \]
    with probability at least $1-c_4| N|\exp\lrp{-c_5 n}$.
    Finally, since the rows of $\bX$ are i.i.d.\ $\sgm_x$-sub-Gaussian, another union bound over $ N$ yields
    $\max_{\bu\in N}\nrm{\bX\bu}{}^2\lesssim\sgm_x^2 n p$ with probability at least $1-c_6\exp\lrp{-c_7 n}$.
    Combining the last two displays gives
    $\nrm{\bX'\bcE/n}{}^2\lesssim\sgm_x^2\sgm_\varepsilon^2 p/n$ with probability at least $1-c_4| N|\exp\lrp{-c_5 n}-c_6\exp\lrp{-c_7 n}$.
    Since $| N|\le5^p$, we have the conclusion.
\end{proof}
We note that the source regression model under a linear feature map $\Zp\sk=\Xp\sk\Pp\sk+\Xip\sk$ can be written as
\[\by_s\sk = \Xp\sk\omg_{1\fm}\skt + \Epm\sk\]
where $\omg_{1\fm}\skt := \omg_1\skt + \Pp\sk\omg_2\skt$ is transferred by the homogeneous transfer learning approach and $\Epm\sk := \Xip\sk\omg_2\skt + \Ep\sk$ is the new error term that includes the ignored signal from the mismatched features. $\Epm\sk$ is another sub-Gaussian vector with variance proxy $\sgm_{\eps_\fm}\sk:=\sxp\sk\nrm{\omg_2\skt}{}+\sep\sk$.

Define $\nu\sk:=\h\omg\sk-\bar\omg\st$.
In the following lemma, we list convergence rates for $\nu:=\h\omg-\bar\omg\st=\frac{1}{K}\sum_{k=1}^K\nu\sk$, $\nu_1:=\h\omg_{1\fm}-\bar\omg_1\st$, and $\nu_{1\fm}:=\h\omg_{1\fm}-\bar\omg_{1\fm}\st$. We first establish an elemental OLS convergence rate on the data-rich source environment.
\begin{lemma}[source regression estimation]\label{lem:nu-bound}
    Denote $\sp\sk{\sep\sk}$ by $\sgm_{d\eps}\sk$, $\sop\sk\sep\sk$ by $\sgm_{x\eps}\sk$, and $\sop\sk\stp\sk$ by $\sgm_{xz}\sk$ for $k\le K$.
    Assume that $\np\sk\gtrsim({\sgm_d\sk})^2p$ and $\sop\sk$ are fixed for all $k\le K$. Then,
    \begin{align*}
        \pr&\bxs{\nrm{\nu}{}^2\gtrsim\sum_{k=1}^K\frac{(\sgm_{d\eps}\sk)^2p}{K\np\sk}}\lesssim K\exp(-c p), \\
        \pr&\bxs{\nrm{\nu_{1\fm}}{}^2\gtrsim\sum_{k=1}^K\lrp{\sxp\sk\nrm{\omg_2\skt}{}+\sep\sk}^2\frac{{p_1}}{K\np\sk}}\lesssim K\exp(-c{p_1}), \\
        \pr&\bxs{\nrm{\nu_1}{}^2\gtrsim \sum_{k=1}^K\frac{(\sgm_{xz}\sk)^2}{K}\nrm{\omg_2\skt}{}^2+\sum_{k=1}^K\frac{{p_1}(\sgm_{x\eps}\sk)^2}{K\np\sk}}\lesssim K\exp(-c{p_1}).
    \end{align*}
\end{lemma}
Since $\h\omg\sk$ and $\h\omg_{1\fm}\sk$ are dense OLS estimators on the source domain, the stochastic parts of these bounds match the minimax source floors $p/\np$ and ${p_1}/\np$ (Theorems \ref{thm:htl-lb}---\ref{thm:hmtl-lb}); the $\log p$ factor in generic Lasso/RSC bounds is not needed here.
\begin{proof}
    For each $k$, $\h\omg\sk$ is OLS in $\by_{s}\sk=\Dp\sk\omg\st+\Ep\sk$ with design $\Dp\sk\in\R^{\np\sk\times p}$.
    We have $\lmd_{\min}(\Sgm\sk)\gtrsim1$ uniformly in $k$, so Lemma \ref{lem:ols-spectral} with $d=p$ and $\sgm_\varepsilon=\sep\sk$ yields
    \[\nrm{\nu\sk}{}^2\lesssim\frac{(\sop\sk\sep\sk)^2 p}{\np\sk}=\frac{(\sgm_{d\eps}\sk)^2 p}{\np\sk}\]
    with probability at least $1-c\exp(-c'\np\sk)$, provided $\np\sk\gtrsim(\sop\sk)^2 p$.
    Averaging over $k$ and using $\nrm{\nu}{}^2\le\frac{1}{K}\sum_k\nrm{\nu\sk}{}^2$ gives the bound for $\nu$.

    For $\nu_{1\fm}\sk:=\h\omg_{1\fm}\sk-\bar\omg_{1\fm}\st$, OLS on $\Xp\sk$ with response $\by_{s}\sk=\Xp\sk\omg_{1\fm}\skt+\Epm\sk$ and effective noise scale $\sgm_{\eps_\fm}\sk:=\sxp\sk\nrm{\omg_2\skt}{}+\sep\sk$ gives
    \[\nrm{\nu_{1\fm}\sk}{}^2\lesssim\frac{(\sop\sk\sgm_{\eps_\fm}\sk)^2{p_1}}{\np\sk}\]
    by Lemma \ref{lem:ols-spectral} with $d={p_1}$, using $\lmd_{\min}(\Sgm_{x_s}\sk)\gtrsim1$ and $\np\sk\gtrsim(\sop\sk)^2{p_1}$.
    Averaging over $k$ yields the bound for $\nu_{1\fm}$.

    For $\nu_1$, the same OLS normal equations on $\Xp\sk$ give
    \[\lrp{\Xp\sk}'\Xp\sk\,\nu_1=\lrp{\Xp\sk}'\lrp{\Zp\sk\omg_2\skt+\Ep\sk}.\]
    On $\brcs{\lmd_{\min}\lrp{\lrp{\Xp\sk}'\Xp\sk/\np\sk}\ge \lmd_{\fm}/2}$,
    \[\nrm{\nu_1}{}\lesssim\frac{1}{\np\sk}\lrnrm{\lrp{\Xp\sk}'\Zp\sk\omg_2\skt}{}+\frac{1}{\np\sk}\lrnrm{\lrp{\Xp\sk}'\Ep\sk}{}.\]
    The first term is $\sgm_{xz}\sk\nrm{\omg_2\skt}{}$ (cross-covariance bias); the second is $\sgm_{x\eps}\sk\sqrt{{p_1}/\np\sk}$ by Lemma \ref{lem:ols-spectral}.
    Squaring and averaging over $k$ gives the bound for $\nu_1$.
\end{proof}

\begin{remark}
\label{rmk_multi}
    Consider a simple orthogonal design case $\tr(\Xp\sk\,'\Xp\sk)=\np\sk{p_1}(\sop\sk)^2$.
    Under our problem setup, with high probability
    \[
    \nrm{\nu_{1\fm}}{}^2=\cO\bxs{\sum_{k=1}^K\lrp{\sxp\sk\nrm{\omg_2\skt}{}+\sep\sk}^2\frac{{p_1}}{K\np\sk}}
    \] 
    while the pooling estimator with pooled source data, $\lrp{\sum_k\Xp\sk\,'\Xp\sk}^{-1}\sum_k\Xp\sk\,'\by_s\sk$, would yield the error rate of
    \[
    \cO\bxs{\sum_{k=1}^K\lrp{\sxp\sk\nrm{\omg_2\skt}{}+\sep\sk}^2\frac{{p_1}}{\sum_k\np\sk}}.
    \]
    Under the assumption that $K\np\sk \asymp \sum_k\np\sk$ for all $k$, both estimators achieve comparable error rates, and the pooling estimator does not yield a faster rate than the non-pooling estimator.
\end{remark}

\subsection{Feature map estimation}\label{sec:mapest}

First, recall that our true feature map is expressed as $\bH_l(\bX_s)=\E(\bZ_l|\bX_s)$ for $l,s\in\brcs{t,s}$.
In our transfer learning problem, the basic idea of imputing mismatched features $\bZ_t$ is to leverage the relationships:
\[\bH_t(\Xt)=\E(\Zt|\Xt)\approx\E(\Zp|\Xt)=\bH_s(\Xt).\]
So, the success of the imputation strategy depends on the estimation of the conditional mean function $\h\bh_s(\bx)=\h\E(\zp|\bx)$ and
\[\bR_\bZ:=\Zt-\hZt=\Xit+\bH_t(\Xt)-\h\bH_s(\Xt).\]
To control $\frac{1}{\nt}\nrm{\bR_\bZ\bu}{}^2$ for an arbitrary vector $\bu\in\R^{p_2}$, we can apply the RSC lemma again. 
Note that the rows of $\bR_\bZ$ are independent.  
Both $\bH_t(\Xt)$ and $\h\bZ_t=\h\bH_s(\Xt)$ are obtained through 
row-wise linear transformations of $\Xt$, so the imputation residual inherits 
the independence structure of the rows of $\Xt$, regardless of whether the 
feature map is linear or nonlinear.

\subsubsection{Linear feature map estimation in section \ref{sec:lfm}}\label{sec:prf-lfm}
Here, we study the case when we have $\bh$ as a linear map on both populations, i.e.,
\[\bh(\bx)={\bP}'\bx, \bP\in\R^{{p_1}\times{p_2}}.\]
Now we consider the multiple-source study case.
Let $\bR_{\Pp}\sk:=\Pp\sk-\h{\Pp\sk}$ and $\bar\bR_{\Pp}:=\frac{1}{K}\sum_{k\le K}\bR_{\Pp}\sk$.
Then, for $\bar\Dlt_\bP=\Pt-\bar\bP_s$ for $\bar\bP_s:=\frac{1}{K}\sum_{k=1}^K\Pp\sk$, the control of an empirical imputation error
\[\bR_\bZ=\Zt-\hZt=\Xt(\bar\Dlt_\bP+\bar\bR_{\Pp})+\Xit\]
is conditionally a $\tau_{r_z}$-sub-Gaussian ensemble given $\bar\bR_{\Pp}$ (since $\bar\bR_{\Pp}$ is a function of $\Xp\sk,\Xip\sk$, it is independent with $\Xt,\Xit$) where $\tau_{r_z}:=\sot(\nrm{\bar\bR_{\Pp}}{}+\nrm{\bar\Dlt_\bP}{})+\sxt$.

For an arbitrary row $\bd_s$ of $\Dp$, we know that $\bd_s$ has the sub-Gaussian norm of order $\rp=\nrm{\Pp}{}$.
For an arbitrary row $\t\bd_t$ of $\tDt$, for $\bu=(\bu_1',\bu_2')'\in S^{{p_1}+{p_2}-1}$ we have
\[\t\bd_t'\bu=\xt'\bu_1+\xt'(\bar\bP_s-\bar\bR_{\Pp})\bu_2, \E(\t\bd_t)=\bo_p.\]
Therefore,
\begin{align*}
    \nrm{\t\bd_t'\bu}{\psi_2}&\lesssim1+\bar\rp+\nrm{\bar\bR_{\Pp}}{F}=:\tau_{\t d}.
\end{align*}
i.e., $\tDt$ is conditionally a $\tau_{\t d}$-sub-Gaussian ensemble given $\bar\bR_{\Pp}$. 

To justify this, first note that on the event $\{\|\bar\bR_{\Pp}\|_F\le r\}$ we have
\[
    \Pr\lrp{|\t\bd_t'\bu|>(1+\bar\rp+r)t|\bar\bR_{\Pp}}\le
    \Pr\lrp{|\t\bd_t'\bu|>\tau_{\t d}t|\bar\bR_{\Pp}}
   < 2\exp(-c t^{2})
\]
almost surely.  
Since the indicator $\mathbb I_{\{\|\bar\bR_{\Pp}\|_F\le r\}}$ is measurable with respect to $\bar\bR_{\Pp}$, taking expectations yields
\[
    \Pr\left(\|\bar\bR_{\Pp}\|_F\le r,|\t\bd_t'\bu|>(1+\bar\rp+r)t\right)
   \le 
   2\exp(-c t^{2})\Pr\lrp{\|\bar\bR_{\Pp}\|_F\le r}.
\]
Therefore,
\[
\Pr\bxs{|\t\bd_t'\bu|>(1+\bar\rp+r)t}
   \le 
   2\exp(-c t^{2})
   + \Pr\left(\|\bar\bR_{\Pp}\|_F>r\right),
\]
which establishes the claim.  
The second term on the right-hand side becomes negligible when the source sample
size is large enough to guarantee a high-probability concentration of
$\bar\bR_{\Pp}$ around zero.  
The same argument will be used repeatedly when specifying the variance proxies for the imputed design matrices throughout the analysis.

\paragraph{Control of $\bar\bR_{\Pp}$.}
\begin{lemma}[Linear map estimation via column-wise lasso]
    \label{lem:lfm-err}
    Assume each column $\bp_{s,j}\sk$ of the true feature map matrix $\Pp\sk$ has at most $q_j\sk:=\nrm{\bp\sk_{s,j}}{0}$ nonzero entries, and let $s_\bP\sk=\sum_{j=1}^{{p_2}} q_j\sk$ and $m_\bP\sk:=\max_{j\le p_2}q_j\sk$. Set penalty parameters $\gamma_j\sk = C_k\sop\sk\sqrt{\frac{\log{p_1}}{\np\sk}}$ for each $j\le{p_2}$ and $k\le K$. If
    \[\np\sk\gtrsim\lrp{\sop\sk\sxp\sk}^2 m_\bP\sk\log{p_1}~ \forall k\le K,\]
    then with probability at least $1-K{p_2}\exp(-c\log{p_1})$,
    \begin{align*}
        \nrm{\bar\bR_{\Pp}}{F}^2&\lesssim\frac{1}{K}\sum_{k=1}^K(\sop\sk\sxp\sk)^2\frac{s_\bP\sk\log{p_1}}{\np\sk},\\
        \nrm{\bar\bR_{\Pp}}{1}^2&\lesssim\frac{1}{K}\sum_{k=1}^K(\sop\sk\sxp\sk)^2\frac{\lrp{m_\bP\sk}^2\log{p_1}}{\np\sk}
    \end{align*}
\end{lemma}
The bound improves with the number of common features ${p_1}$, and the rate is faster when the feature map is sparser (i.e., smaller $s_\bP$) and the source sample size $\np\sk$ is larger. Also, for the high-probability guarantee to hold, ${p_1}$ needs to grow faster than ${p_2}$ and $K$.
\begin{proof}
    For each column $j$, consider the lasso problem
    \[\h\bp_{s,j}\sk := {\arg\min}_\bp\lrp{\frac{1}{\np\sk}\nrm{(\Zp\sk)_{.j} - \Xp\sk\bp}{}^2 + \gamma_j\sk\nrm{\bp}{1}}.\]
    
    By standard lasso theory (e.g., \citealp{buhlmann2011statistics}), under the condition $\np\sk\gtrsim{\sxp\sk}^2 q_j\sk\log{p_1}$ and the choice $\gamma_j\sk = C_k\sop\sk\sqrt{\frac{\log{p_1}}{\np\sk}}$, with probability at least $1-c\exp(-c'\log p_1)$,
    \[\nrm{\br_{\Pp,j}\sk}{}^2 \lesssim (\sop\sk\sxp\sk)^2\frac{q_j\sk\log{p_1}}{\np\sk},\quad \nrm{\br_{\Pp,j}\sk}{1}^2\lesssim(\sop\sk\sxp\sk)^2\frac{(q_j\sk)^2\log{p_1}}{\np\sk}.\]
    
    Applying the union bound over all $j \le {p_2}$ columns, the complementary event's probability is at most ${p_2}\exp(-c\log{p_1})$. Thus, with the same confidence,
    \[\nrm{\bR_{\Pp}\sk}{F}^2 = \sum_{j=1}^{{p_2}}\nrm{\br_{\Pp,j}\sk}{}^2 \lesssim (\sop\sk\sxp\sk)^2\frac{s_\bP\sk\log{p_1}}{\np\sk},\quad\nrm{\bR_{\Pp}\sk}{1}^2\lesssim(\sop\sk\sxp\sk)^2\frac{(m_\bP\sk)^2\log{p_1}}{\np\sk}\]
    where $s_\bP\sk:=\sum_{j=1}^{{p_2}} q_j\sk$ and $m_\bP\sk:=\max_{j\le p_2}q_j\sk$. This completes the proof.
\end{proof}
Therefore, this concludes that
\[\tau_{\t d}\lesssim1+\bar\rp+\sqrt{\frac{1}{K}\sum_{k=1}^K\frac{s_\bP\sk\log{p_1}}{\np\sk}}\]
almost surely.

\subsubsection{Nonparametric feature map with Sieve estimation}

We work under the setup in sections \ref{sec:design} and \ref{sec:nlfm}.
First, recall that our true feature map is expressed as $\bH(\bX)=\Psi_\infty(\bX)\Tht_\infty$
for the oracle parameter sequence $\Tht_\infty$ according to section \ref{sec:nlfm}.
In our transfer learning problem, the basic idea of imputing mismatched features $\bZ_t$ is to leverage the relationships:
\begin{align*}
    \Psi_\infty(\Xt)\Tht_{t,\infty}\approx\Psi_M(\Xt)\Tht_{t,M}\approx\Psi_M(\Xt)\Tht_{s,M}\\
    \text{ where }\Tht_{s,M}:=\frac{1}{K}\sum_{k\le K}\Tht_{s,M}\sk\approx\h\Tht_{s,M}:=\frac{1}{K}\sum_{k\le K}\h\Tht_{s,M}\sk
\end{align*}
for matrices $\Tht_{t,M},\Tht_{s,M}\sk\in\R^{M\times{p_2}}$ containing Fourier coefficients truncated at $M$.
Now, let $\bR_{\Tht_{s,M}}\sk:=\Tht_{s,M}\sk-\h\Tht_{s,M}\sk$ and $\bar\bR_{\Tht_{s,M}}:=\frac{1}{K}\sum_{k\le K}\bR_{\Tht_{s,M}}\sk$.
Then, for $\bR_\bZ=\Zt-\hZt$ and $\bE_\bH:=\Psi_\infty(\Xt)\Tht_{t,\infty}-\Psi_M(\Xt)\Tht_{t,M}=\bH_t(\Xt)-\Psi_M(\Xt)\Tht_{t,M}$ we have
\begin{align*}
    \bR_\bZ-\Xit&=\bE_\bH+\Psi_M(\Xt)\Tht_{t,M}-\Psi_M(\Xt)\h\Tht_{s,M}=\bE_\bH+\Psi_M(\Xt)(\bar\Dlt_\Tht+\bar\bR_{\Tht_{s,M}})
\end{align*}
for $\bar\Dlt_\Tht:=\Tht_{t,M}-\bar\Tht_{s,M}$ and $\bar\Tht_{s,M}:=\frac{1}{K}\sum_{k\le K}\Tht_{s,M}\sk$.
That is, the imputation error depends on the series approximation error $\bE_\bH$ in the target domain and the truncated Fourier coefficient estimation errors in the source domain $\bar\bR_{\Tht_{s,M}}$.
We will obtain the control of $\Psi_M(\Xt)(\bar\Dlt_\Tht+\bar\bR_{\Tht_{s,M}})$ using the RSC Lemma.

As $\bh\in\cH$, recall that our imputation becomes $\tDt=[\Xt,\,\hZt]=[\Xt\vert\Psi_M(\Xt)\h\Tht_{s,M}]$.
Note that for an arbitrary row $\t\bd_t$ of $\tDt$, for $\bu=(\bu_1',\bu_2')'\in S^{{p_1}+{p_2}-1}$ we have
\[\t\bd_t'\bu=\xt'\bu_1+\b\psi_M(\xt)'(\bar\Tht_{s,M}-\bar\bR_{\Tht_{s,M}})\bu_2, \E(\t\bd_t)=\bo_p.\]
Therefore,
\begin{align*}
    \nrm{\t\bd_t'\bu}{\psi_2}&\le\sot+\nrm{\lrp{\bar\Tht_{s,M}-\bar\bR_{\Tht_{s,M}}}\bu_2}{}\sgm_\psi
    \lesssim\bar\pi_s+\nrm{\bar\bR_{\Tht_{s,M}}}{}=:\tau_{\t d}
\end{align*}
for $\bar\pi_s:=\nrm{\bar\Tht_{s,\infty}}{}$. i.e., $\t\bd_t$ is conditionally $\tau_{\t d}$-sub-Gaussian given $\bar\bR_{\Tht_{s,M}}$. For any $\bu\in S^{p-1}$ and $R(r):=\brcs{\nrm{\bar\bR_{\Tht_{s,M}}}{}<r}$, we have for $\tau_{\t d}=\bar\pi_s+r$:
\[\pr\lrp{\vrt{\bu'\t\bd_t}>t\vert R(r)}<2\exp(-ct^2/\tau_{\t d}^2).\]
\paragraph{Control of $\bE_\bH$.} According to \citet[Lemma C.7 and equation 68]{zhang2023regression}, an oracle parameter $\btht_j(M)=(\tht_{1j},...,\tht_{Mj})\in\R^M$, which contains Fourier coefficients truncated at $M$, satisfies
\begin{align*}
    \sum_{j\le{p_2}}\E[e_j(\bx)^2]=\E\nrm{\be_h(\bx)}{}^2=\cO\bxs{{p_2}\lrp{\frac{\log^{\post-1}M}{M}}^{2/3}}
\end{align*}
for series approximation errors $e_j(\bx):=h_j(\bx)-{\btht_j(M)}'\b\psi_M(\bx)$ and their concatenation $\be_h(\bx):=\bh(\bx)-\Tht'\b\psi_M(\bx)$ under the function class for $\bh$ we consider.
This implies that $\nrm{\be_h(\bx)}{}=o_\pr(1)$ provided that ${p_2}\ll\lrp{\frac{M}{\log^{\post-1}M}}^{2/3}$ as $p,M$ grow properly.

\paragraph{Control of $\bar\bR_{\Tht_{s,M}}$ and the choice of $M$.} \cite{zhang2023regression} also provides a near-minimax optimal rates of $\E_{\xt}\nrm{\Psi_M(\Xt)\h\Tht_{t,M}-\bH_t(\Xt)}{F}^2$ where $\h\Tht_{t,M}$ is obtained through solving the same $\ell_1$-penalized multivariate regression on the target domain.
However, $\h\Tht_{t,M}$ is not obtainable as we do not observe $\Zt$ in our heterogeneous transfer learning.
Instead, we use $\h\Tht_{s,M}$ to address this issue, and to guarantee the convergence of $\bar\bR_{\Tht_{s,M}}$, we need to specify the order of series approximation $M=M_k(p,\np\sk)$ for each $k\in[K]$ where $M_k$ is decreasing in $k$ (as $\np\sk$ would decrease in $k$).
The following Proposition is adapted from \citet[Corollary E.6]{zhang2023regression}.
\begin{proposition}\label{prp-zhngsmn1}
    Let ${p_1}'\ge\post$ be an initial guess of $\post$ and $\gamma_{j,k}\sim C_{{p_1}'}\sqrt{\frac{\log M_k}{\np\sk}}$ for all $j,k$. 
    Define
    \[\h\Tht_{s,M}\sk:={\arg\min}_{\Tht\in\R^{M\times{p_2}}}\brcs{\frac{1}{\np\sk}\nrm{\Zp\sk-\Psi_{M_k}(\Xp\sk)\Tht}{F}^2+\sum_{j\le{p_2}}\gamma_{j,k}\nrm{\btht^j}{1}}.\]
    Then,
    \begin{align*}
        \nrm{\bR_{\Tht_{s,M}}\sk}{1}\ge\log M_k\cdot\lrp{\frac{\log^{\post-1}\np\sk}{{\np\sk}}}^{2/3}
    \end{align*}
    with probability at most
    \[\frac{{p_2}}{2M_k}+\frac{{p_2}\exp(-c_{{p_1}'}\np\sk)}{M_k^2}+{p_2}\exp\lrp{-c_{\post}\sqrt[3]{\np\sk\log^{2(\post-1)}\np\sk}}.\]
\end{proposition}
The fixed estimate ${p_1}'$ is used to facilitate an efficient unraveling of basis functions and computations.
Using the matrix norm equivalence $\frac{1}{{p_2}}\nrm{\bR_{\Tht_{s,M}}\sk}{F}^2\le\nrm{\bR_{\Tht_{s,M}}\sk}{1}^2$ and taking union bound over $K$ source domains, we have
\begin{align*}
    \nrm{\bar\bR_{\Tht_{s,M}}}{F}^2\ge\sum_{k=1}^K\frac{{p_2}\log^2M_k}{K}\cdot\lrp{\frac{\log^{\post-1}\np\sk}{{\np\sk}}}^{4/3}
\end{align*}
with probability at most
\[{p_2}\sum_{k=1}^K\brcs{M_k^{-1}+\exp\lrp{-c_{\post}\sqrt[3]{\np\sk\log^{2(\post-1)}\np\sk}}}.\]
So, if the following holds:
\[{p_2}=\cO\bxs{\min\brcs{1/\sum_kM_k^{-1},\sum_k\exp\lrp{\sqrt[3]{\np\sk\log^{2(\post-1)}\np\sk}}}},\]
then this probability becomes negligible, and we can guarantee a high-probability upper bound for $\nrm{\bar\bR_{\Tht_{s,M}}}{}$.

To aggregate these requirements, let $M_k=M$ for all $k$. The order of $M$ must satisfy two conditions: (i) from the probability bound, $M \gg K{p_2}$ to make $K{p_2}/M=o(1)$; and (ii) from the series approximation error control, $M$ must be as large as $p_2^{3/2}(\log p_2)^{\post+0.5}$ to ensure ${p_2}^{3/2}\ll\frac{M}{\log^{\post-1}M}$ thereby $\nrm{\bE_\bH}{F}=o_\pr(\sqrt{\nt})$. Therefore, we specify
\[M\asymp\max\lrp{K{p_2},p_2^{3/2}(\log p_2)^{\post+0.5}} = {p_2}\max\lrp{K,\sqrt{p_2(\log p_2)^{2\post+1}}}.\]
Consequently, for $t=1/\sqrt[4]{\log p_2}$ we have
\[\pr\lrp{\nrm{\be_h(\bx)}{}>t}<\frac{\E\nrm{\be_h(\bx)}{}^2}{t^2}\lesssim\frac{1}{\sqrt{\log p_2}}\]
by Markov.
\begin{corollary}\label{cor:RTht}
    Assume that $K$ is fixed, $M\asymp\sqrt{p_2^3(\log p_2)^{2\post+1}}$ and $\np\sk\log^{2(\post-1)}\np\sk\gg(\log{p_2})^3$ for all $k\le K$. Then, if $K$ is fixed,
    \begin{align*}
        \pr\bxs{\nrm{\bar\bR_{\Tht_{s,M}}}{F}^2\gtrsim\frac{1}{K}\sum_{k=1}^K\lrp{{p_2}^{3/4}\log^{3/2}{p_2}\frac{\log^{\post-1}\np\sk}{{\np\sk}}}^{4/3}}\lesssim\frac{1}{\sqrt{p_2}(\log p_2)^{\post+0.5}}.
    \end{align*}
\end{corollary}
To summarize, for an arbitrary row $\br_\bZ:=\zt-\h\bz_t$ of $\br_\bZ$ and $\bu\in S^{p-1}$,
\[\br_\bZ=\xi_t+\be_h(\xt)+(\bar\Dlt_\Tht+\bar\bR_{\Tht_{s,M}})'\psi(\xt),\quad \E(\br_\bZ)=\bo_{p_2}\]
and conditionally given $\be_h(\xt)$ and $\bar\bR_{\Tht_{s,M}}$,
\[\nrm{\bu'\br_\bZ}{\psi_2}\le\sxt+\nrm{\be_h(\xt)}{}+(\nrm{\bar\Dlt_\Tht}{}+\nrm{\bar\bR_{\Tht_{s,M}}}{F})\sgm_\psi\lesssim \sxt+\nrm{\be_h(\xt)}{}+\nrm{\bar\Dlt_\Tht}{}+\nrm{\bar\bR_{\Tht_{s,M}}}{F}=:\tau_{r_z}.\]
Note that $\nrm{\be_h(\xt)}{}+\nrm{\bar\bR_{\Tht_{s,M}}}{F}=o_\pr(1)$ under the previous discussions.

\section{Proofs of Main Results}
\subsection{Minimax Optimality}
Now, recall that $\Gmm\in\brcs{\Tht_\infty,\bP}$ denotes a choice of feature mapping process.
For each law $\cP\in\Pi\sht$, write the parameters of interest as $(\beta\st,\omg\st,\Gmm_s,\Gmm_t)$ for the true regression and feature map parameters generating $\cP$.
Let $D_{\rm KL}(F\,\Vert\,G)$ denote the KL divergence between two probability laws $F,G$. We use a similar approach to that in \cite{li2022transfer,tian2022transfer} to prove the minimax optimality of $\h\beta$ and $\h\beta_{1\fm}$ by Fano reduction \citep{tsybakov2009introduction} and hypercube packing arguments \citep{raskutti2011minimax}.
\begin{lemma}[Fano reduction, \citealp{tsybakov2009introduction}]\label{lem:fano}
    Let $\brcs{\cP_1,\dots,\cP_M}$ be probability measures on a common space, indexed by parameters $\tht_1,\dots,\tht_M$ such that $\rho(\tht_j,\tht_m)\ge2\dlt>0$ for $j\ne m$ in a semimetric $\rho$. 
    Then, for any estimator $\h\tht$,
    \[\max_{1\le m\le M}\cP_m\lrp{\rho(\h\tht,\tht_m)\ge\dlt}\ge1-\frac{\max_{j\ne m}D_{\mathrm{KL}}(\cP_j\Vert\cP_m)+\log2}{\log M}.\]
\end{lemma}
\begin{proof}
    Define the minimum-distance test $\h\psi:=\arg\min_m\rho(\h\tht,\tht_m)$. If $\rho(\h\tht,\tht_j)<\dlt$ then by the triangle inequality and $2\dlt$-separation $\h\psi=j$, so $\cP_j(\h\psi\ne j)\le \cP_j(\rho(\h\tht,\tht_j)\ge\dlt)$. Fano's inequality gives $\max_m \cP_m(\h\psi\ne m)\ge1-(\bar I+\log2)/\log M$ with $\bar I\le\max_{j\ne m}D_{\mathrm{KL}}(\cP_j\|\cP_m)$, which yields the claim.
\end{proof}
\begin{lemma}\label{lem:gv}
    For any integers $1\le s\le d/2$ there is a subset $$\cL(s)\subseteq\brcs{z\in\brcs{-1,0,1}^d:\nrm{z}{0}=s}$$ with pairwise Hamming distance at least $s/2$ and $\log|\cL(s)|\ge\frac{s}{2}\log\frac{d-s}{s/2}$. In particular $\nrm{z-z'}{}^2\ge s/2$ for distinct $z,z'\in\cL(s)$, and for $s\le d/2$ one has $\log|\cL(s)|\ge c\,s\log(d/s)$ for an absolute constant $c>0$.
\end{lemma}
Write a joint law as $\cP=\cP^{s}\otimes\cP^{t}$ on $\cX=\cX^{s}\times\cX^{t}$. For two hypotheses $j,m$, the independence between source and target gives
    \begin{align*}
        D_{\mathrm{KL}}(\cP_j\,\|\,\cP_m)
        &=\E_{\cP_j}\log\frac{\partial\cP_j}{\partial\cP_m}(\bd_s,\bd_t)
        =\E_{\cP_j}\log\frac{\partial\cP^{s}_j}{\partial\cP^{s}_m}(\bd_s)+\E_{\cP_j}\log\frac{\partial\cP^{t}_j}{\partial\cP^{t}_m}(\bd_t) \\
        &=D_{\mathrm{KL}}(\cP^{s}_j\,\|\,\cP^{s}_m)+D_{\mathrm{KL}}(\cP^{t}_j\,\|\,\cP^{t}_m).
    \end{align*}
    With $\np$ i.i.d.\ source observations and $\nt$ i.i.d.\ target observations,
    \begin{equation}\label{eq:kl-additive}
        D_{\mathrm{KL}}(\cP_j\,\|\,\cP_m)=\np\,D_{\mathrm{KL}}(\cP^{s}_{1,j}\,\|\,\cP^{s}_{1,m})+\nt\,D_{\mathrm{KL}}(\cP^{t}_{1,j}\,\|\,\cP^{t}_{1,m}),
    \end{equation}
    where $\cP^{s}_{1,\cdot},\cP^{t}_{1,\cdot}$ denote the single-observation laws.
\begin{proof}[Proof of Theorem \ref{thm:htl-lb}]
    Suppose $\eps\sim\cN(0,\sgm^2)$ independent of $\bd$, and $\E(\bd\bd')=\Sgm$. For two coefficient vectors $\beta^{\scriptscriptstyle(j)},\beta^{(h)}$ with $\bs(j,h):=\beta^{\scriptscriptstyle(j)}-\beta^{(h)}$, the conditional laws of $Y|\bd$ differ only by a shift $\langle\bd,\bs(j,h)\rangle$, so
    \begin{align*}
        D_{\mathrm{KL}}(\cP_{1,j}\|\cP_{1,m})
        =\E_j\bxs{\log\frac{p_j(Y\mid\bd)}{p_m(Y\mid\bd)}}
        =\E_j\bxs{\frac{\langle\bd,\bs(j,h)\rangle^2}{2\sgm^2}}
        =\frac{1}{2\sgm^2}\bs(j,h)'\,\Sgm\,\bs(j,h) \\
        \le\frac{\lmd_{\max}(\Sgm)}{2\sgm^2}\nrm{\bs(j,h)}{}^2
    \end{align*}
    where $\Sgm=\begin{bmatrix}
        \Sgm_x & \Sgm_{x\t x}\Gmm \\ \Gmm'\Sgm_{\t xx} & \Gmm'\Sigma_{\t x}\Gmm+\Sgm_\xi
    \end{bmatrix}$ using the notations from Section \ref{sec:apdx}.
    Throughout the following constructions, we let $\sop=\sxp=\sep=\sxt=\set=1$ and fix $\Sgm_{x_s}=\Sgot=\bI_{p_1}$, $\Gmm_s'\Gmm_s=\Gmm_s'\Gmm_t=\Gmm_t'\Gmm_s=\Gmm_t'\Gmm_t=\Sgxp=\bI_{p_2}$.
    Consequently, every construction below lies in $\Pi\sht$.
    \paragraph{Construction A.}
    Fix $\dst=\b0_p$ and $\beta_1\st=\b0_{{p_1}}$. Then $\beta\st=\omg\st$ and $\nrm{\dst}{1}=0$. Suppose that $\nt\le\np$ and $p_1=p_2$. Let the mismatched block vary over
    \begin{equation}\label{eq:pack-A}
        \cT_A:=\brcs{\beta_2\st=r_A \bz:\ \bz\in\cL(s_A)},\qquad s_A:=\lceil{p_2}/4\rceil,\qquad r_A:=c_0\,\sqrt{\frac{2}{\np}},
    \end{equation}
    with $\cL(s_A)\subseteq\brcs{-1,0,1}^{p_2}$ from Lemma \ref{lem:gv}. For distinct $\bz^{\scriptscriptstyle(j)},\bz^{(m)}\in\cL(s_A)$,
    \begin{equation}\label{eq:sep-A}
        \nrm{\beta_2^{\scriptscriptstyle(j)}-\beta_2^{(m)}}{}^2=r_A^2\nrm{\bz^{\scriptscriptstyle(j)}-\bz^{(m)}}{}^2\ge\frac{s_A}{2}\,r_A^2\ge c'\,\frac{p_2}{\np},
    \end{equation}
    and $\log|\cT_A|\ge c''\,p_2$.
    For $\bs_2(j,m):=\beta_2^{\scriptscriptstyle(j)}-\beta_2^{(m)}$, the source block contributes
    \[\np\,D_{\mathrm{KL}}(\cP^{s}_{1,j}\|\cP^{s}_{1,m})=\frac{\np}{2}\,\bs_2(j,m)'\Sgm_{z_s}\bs_2(j,m)\le\frac{\np}{2}\,\lmd_{\max}(\Sgm_{z_s})\nrm{\bs_2(j,m)}{}^2,\]
    with $\Sgm_{z_s}=\Pp'\Sgm_{x_s}\Pp+\Sgxp=2\bI_{p_2}$ hence $\lmd_{\max}(\Sgm_{z_s})=2$. 
    The target block contributes
    \[\nt\,D_{\mathrm{KL}}(\cP^{t}_{1,j}\,\|\,\cP^{t}_{1,m})=\frac{\nt}{2}\nrm{\Pt\bs_2(j,m)}{}^2\le\frac{\nt}{2}\nrm{\bs_2(j,m)}{}^2.\]
    Using \eqref{eq:kl-additive} and $\nrm{\bs_2(j,m)}{}^2\le2s_Ar_A^2\le2c_0^2p_2/\np$,
    \begin{equation}\label{eq:kl-A}
        D_{\mathrm{KL}}
        (\cP_j\,\|\,\cP_m)\le2c_0^2p_2+\frac{\nt c_0^2p_2}{\np}\le3c_0^2p_2.
    \end{equation}
    Set $\dlt^2:=\tfrac{s_A}{4}r_A^2\asymp p_2/\np$. For small enough $c_0$, \eqref{eq:kl-A} gives $D_{\mathrm{KL}}\le\tfrac14\log|\cT_A|$ hence Lemma \ref{lem:fano} with $M=|\cT_A|$ yields
    \begin{equation}\label{eq:fano-A}
        \max_{\beta_2\st\in\cT_A}\cP_{\beta_2\st}\lrp{\nrm{\h\beta-\beta\st}{}\ge\dlt}\ge\frac12.
    \end{equation}
    Since $\nrm{\dst}{1}=0$ on $\cF_A:=\brcs{\cP_{\beta_2\st};\beta_2\st\in\cT_A}$, we have
    \[\max_{\cP\in\cF_A}\cP\lrp{\nrm{\h\beta-\beta\st}{}^2\gtrsim\frac{p_1}{\np}}\ge\frac12,\]
    matching the first term of the floor.
    
    The condition $p_1=p_2$ is consistent with the high-dimensional regime $p_1\asymp p_2\asymp p$. For the high probability upper bounds in section \ref{sec:theory} to hold, we require $p_2\lesssim p_1$; hence, this is also the sub-regime of the main construction in section \ref{sec:HTL}.
    \paragraph{Construction B.}
    Fix $\omg_1\st=0$, $\omg_2\st=w\bu$ for some $w>0$ and a unit norm vector $\bu\in\R^{p_2}$, $\beta_2\st=\omg_2\st$ and let only $\dlt_1\st$ varies with $\beta_1\st=\dlt_1\st$ and $\dst=(\dlt_1\st,\b0_{p_2})$, so $\nrm{\dst}{1}=\nrm{\dlt_1\st}{1}$. Therefore, we can write
    \[Y_t=\bd_t'\beta\st+\eps_t=\xt'\dlt_1\st+w\zt'\bu+\eps_t=\xt'\dlt_1\st+w\t\bx_t'\Gmm_t\bu+w\xi_t'\bu+\eps_t.\]
    The source law is fixed, hence $D_{\mathrm{KL}}(\cP_j\,\|\,\cP_m)=D_{\mathrm{KL}}(\cP_j^{t}\,\|\,\cP_m^{t})$.
    For $\tau_{\eps_\fm}^2:=1+w^2$, define $\zeta(h):=h\lrp{h\wedge\tau_{\eps_\fm}\sqrt{\log{p_1}/\nt}}$.
    \begin{enumerate}
        \item When $h\le\tau_{\eps_\fm}\sqrt{\log p_1/\nt}$.
        
        Let $\dlt_1^{\scriptscriptstyle(j)},\dlt_1^{(m)}$ be drawn from $\cT_B^{(1)}(h):=\brcs{c_0'h\be_j:\ j\in[p_1]}$ with $M={p_1}$. For $j\ne m$, $\nrm{\dlt_1^{\scriptscriptstyle(j)}-\dlt_1^{(m)}}{}^2=2(c_0'h)^2$ and $\nrm{\dlt_1^{\scriptscriptstyle(j)}}{1}=c_0'h$. We have
        \[
            D_{\mathrm{KL}}(\cP_j^{t}\,\|\,\cP_m^{t})
            =\frac{\nt}{2\tau_{\eps_\fm}^2}\,\nrm{\dlt_1^{\scriptscriptstyle(j)}-\dlt_1^{(m)}}{}^2
            =\frac{\nt}{\tau_{\eps_\fm}^2}(c_0'h)^2
            \le c_0'^2\log{p_1}=c_0'^2\log M
        \]
        For $c_0'$ small, $D_{\rm KL}\le\tfrac14\log M$ and Lemma \ref{lem:fano} with $\dlt=c_0'h/2$ gives
        \[
            \max_{j\le{p_1}}\pr\lrp{\nrm{\h\beta_1-\beta_1\st}{}\ge c_0'h/2}\ge\frac12.
        \]
        Since $\dst=(\dlt_1\st,\b0)$ and $\zeta(h)=h^2$ in this regime, and $\nrm{\h\beta-\beta\st}{}\ge\nrm{\h\beta_1-\beta_1\st}{}$, the event $\nrm{\h\beta-\beta\st}{}^2\ge c_1 \zeta(\nrm{\dst}{1})$ holds on $\cF_B^{(1)}(h):=\brcs{\cP_{\dlt_1\st}:\dlt_1\st\in\cT_B^{(1)}(h)}$ with $\nrm{\dst}{1}=c_0'h$.
        \item When $h>\tau_{\eps_\fm}\sqrt{\log p_1/\nt}$.
        
        Fix a constant $a\in(0,1)$. Let $\dlt_0:=c_0'\tau_{\eps_\fm}\sqrt{\log{p_1}/\nt}$ and $s:=\lfloor h/\dlt_0\rfloor\wedge\lfloor{p_1}^{1-a}\rfloor\ge1$, which keeps the construction within the sparse-contrast regime $s\le p_1^{1-a}$ (so that $\log(p_1/s)\ge a\log p_1$). Set $\cT_B^{(2)}(h):=\brcs{\dlt_0 \bz:\ \bz\in\cL(s)}\subset\R^{p_1}$ with $|\cT_B^{(2)}(h)|=|\cL(s)|$.
        For $\bz^{\scriptscriptstyle(j)}\ne \bz^{(m)}$ in $\cL(s)$, Lemma \ref{lem:gv} gives $\nrm{\bz^{\scriptscriptstyle(j)}-\bz^{(m)}}{}^2\ge s/2$, hence for $\dlt_1^{\scriptscriptstyle(j)},\dlt_1^{(m)}\in\cT_B^{(2)}(h)$,
        \[
            \nrm{\dlt_1^{\scriptscriptstyle(j)}-\dlt_1^{(m)}}{}^2
            =
            \dlt_0^2\nrm{\bz^{\scriptscriptstyle(j)}-\bz^{(m)}}{}^2
            \ge
            \tfrac{s}{2}\dlt_0^2,
            \qquad
            \nrm{\dlt_1\st}{1}=s\dlt_0\le h.
        \]
    \end{enumerate}
    Using $\nrm{\dlt_1^{\scriptscriptstyle(j)}-\dlt_1^{(m)}}{}^2\le2s\dlt_0^2$ and $\Var(\eps_{t\fm})\asymp\tau_{\eps_\fm}^2$, the KL divergence satisfies
    \[
        D_{\mathrm{KL}}(\cP_j^{t}\,\|\,\cP_m^{t})=
        \frac{\nt}{2\Var(\eps_{t\fm})}\,\nrm{\dlt_1^{\scriptscriptstyle(j)}-\dlt_1^{(m)}}{}^2\lesssim
        \frac{\nt s\dlt_0^2}{\tau_{\eps_\fm}^2}=
        c_0'^2\,s\log{p_1}\le\frac{c_0'^2}{a}\,s\log\frac{p_1}{s}\le
        \tfrac14\log|\cL(s)|
    \]
    by Lemma \ref{lem:gv} and $\log(p_1/s)\ge a\log p_1$, for $c_0'$ small; hence Lemma \ref{lem:fano} with $\dlt^2=\tfrac{s}{4}\dlt_0^2$ gives
    \[
        \max_{\dlt_1\st\in\cT_B^{(2)}(h)}\pr\lrp{\nrm{\h\beta_1-\beta_1\st}{}^2\ge\tfrac{s}{4}\dlt_0^2}\ge\frac12.
    \]
    Since $h>\tau_{\eps_\fm}\sqrt{\log{p_1}/\nt}$, we have $\zeta(h)=h\,\tau_{\eps_\fm}\sqrt{\log{p_1}/\nt}\asymp h\dlt_0$, and the realized contrast $\nrm{\dlt_1\st}{1}=s\dlt_0\le h$ satisfies $\dlt^2=\tfrac{s}{4}\dlt_0^2\asymp \zeta(\nrm{\dlt_1\st}{1})\le \zeta(h)$. Thus the transfer rate holds on $\cF_B^{(2)}(h)=\brcs{\cP_{\dlt_1\st}:\dlt_1\st\in\cT_B^{(2)}(h)}$.
    \paragraph{Concluding bound.}
    Fix any contrast level $\nrm{\dst}{1}$ satisfying the condition $$\nrm{\dst}{1}\tau_{\eps_\fm}\sqrt{\log{p_1}/\nt}\le c_0(\nrm{\dst}{1}^2\vee1)$$ of Theorem \ref{thm:htl-lb}, and let $\cF_B$ be the matching packing ($\cF_B^{(1)}$ or $\cF_B^{(2)}$ according to the regime). Constructions A and B give, for any fixed $\h\beta$,
    \[
        \max_{\cP\in\cF_A}{\cP}\lrp{\nrm{\h\beta-\beta\st}{}^2\gtrsim\frac{p_1}{\np}}\ge\frac12
        \quad\text{and}\quad
        \max_{\cP\in\cF_B}{\cP}\bxs{\nrm{\h\beta-\beta\st}{}^2\gtrsim \zeta(\nrm{\dst}{1})}\ge\frac12.
    \]
    The laws in $\cF_A$ have $\nrm{\dst}{1}=0$, so their rate is $\max\lrp{p_1/\np,\zeta(0)}=p_1/\np$, while those in $\cF_B$ gives $\nrm{\dst}{1}$. Hence whichever of $p_1/\np$ and $\zeta(\nrm{\dst}{1})$ is larger is witnessed by some $\cP\in\cF_A\cup\cF_B\subseteq\Pi\sht$, so
    \[
        \sup_{\cP\in\Pi\sht}{\cP}\bxs{\nrm{\h\beta-\beta\st}{}^2\gtrsim\max\brcs{\frac{p_1}{\np},\zeta(\nrm{\dst}{1})}}\ge\frac12.
    \]
    Taking the infimum over $\h\beta$ proves \eqref{eq:htl-lb}.
\end{proof}
\begin{remark}
    To validate the claim that our upper bound in Theorem \ref{thm:lfm-sp-err} is minimax optimal, consider the least favorable submodel we used to prove Theorem \ref{thm:htl-lb}: $\Sgm_{x_s} = \Sgot = \bI_{p_1}$, $\sop = \sgm_\xi = \sep = 1$, $\Pp'\Pp=\Pp'\Pt=\Pt'\Pp=\Pt'\Pt=\bI_{p_2}$, and $n_t \le n_s$. 
    In addition, the probability distributions of errors and design components were assumed to be Gaussian.
    Under this submodel, the upper bound in Theorem \ref{thm:lfm-sp-err} collapses as
    \begin{align*}
        \nrm{\h\beta-\beta\st}{}^2&\lesssim\frac{p}{\np}+\nrm{\dst}{1}\lrp{\nrm{\dst}{1}\wedge\tau_{\eps_\fm}\sqrt{\frac{\log p_1}{\nt}}},
    \end{align*}
    which agrees with the prescribed lower bound in Theorem \ref{thm:htl-lb}.
\end{remark}
\begin{proof}[Proof of Theorem \ref{thm:hmtl-lb}]
    Under HmTL with $\Gmm=\bP$, the marginalized target parameter is $\beta_{1\fm}\st=\beta_1\st+\Pt\beta_2\st$ and the contrast is $\dlt_{1\fm}\st:=\beta_{1\fm}\st-\t\omg_{1\fm}\st$. Fix $\t\omg_{1\fm}\st=0$ (equivalently $\Pt\omg_2\st=-\omg_1\st$), $\omg_2\st=w\bu$, $\beta_2\st=\omg_2\st$, and vary only $\beta_{1\fm}\st$ (equivalently $\dlt_{1\fm}\st$) as in Construction B. The source law is identical across hypotheses, so $D_{\mathrm{KL}}(\cP_j\,\|\,\cP_m)=D_{\mathrm{KL}}(\cP_j^{t}\,\|\,\cP_m^{t})$. 
    The target regression is $Y_t=\xt'\beta_{1\fm}\st+\eps_{t\fm}$. For each $\nrm{\dlt_{1\fm}\st}{1}=h$ satisfying the condition, apply (B1) if $h\le\tau_{\eps_\fm}\sqrt{\log{p_1}/\nt}$ with packing $\brcs{c_0'h\be_j}$ and (B2) otherwise with $\cT_B^{(2)}(h)$. The KL and Fano calculations are identical to Construction B, replacing $\dlt_1\st$ by $\dlt_{1\fm}\st$ and $\beta_1\st$ by $\beta_{1\fm}\st$, yielding
    \[
        \inf_{\h\beta_{1\fm}}\,\sup_{\cP\in\Pi\shm}\cP\bxs{\nrm{\h\beta_{1\fm}-\beta_{1\fm}\st}{}^2\ge c_1\nrm{\dlt_{1\fm}\st}{1}\lrp{\nrm{\dlt_{1\fm}\st}{1}\wedge\tau_{\eps_\fm}\sqrt{\frac{\log{p_1}}{\nt}}}}\ge\frac12.
    \]
    Now, verify the source convergence rate term.
    Fix $\dlt_{1\fm}\st=\bo$, so $\beta_{1\fm}\st=\t\omg_{1\fm}\st$, and set $\beta_2\st=\omg_2\st=w\bu$ for a unit vector $\bu$. Recall $\beta_{1\fm}\st=\omg_{1\fm}\st+\Pt\omg_2\st$ with $\Pt\omg_2\st=w\Pt\bu$ fixed, so packing $\beta_{1\fm}\st$ is equivalent to packing the source marginal $\omg_{1\fm}\st$ up to a fixed offset. Varying $\omg_{1\fm}\st$ now perturbs \emph{both} the source model $Y_s=\xp'\omg_{1\fm}\st+\eps_{s\fm}$, with $\eps_{s\fm}=\eps_s+\xi_s'\omg_2\st$ being Gaussian of variance $\Var(\eps_{s\fm})=\sep^2+\sxp^2\nrm{\omg_2\st}{}^2=1+w^2$, and the target model $Y_t=\xt'\beta_{1\fm}\st+\eps_{t\fm}$ with $\Var(\eps_{t\fm})=1+w^2$. 
    Vary $\omg_{1\fm}\st=r\bz$ with $\bz\in\cL(s)$, $s=\lceil{p_1}/4\rceil$, and $r:=c_0''\sqrt{\nrm{\omg_2\st}{}^2/\np}.$
    For $\bz^{\scriptscriptstyle(j)}\ne \bz^{(m)}$, Lemma \ref{lem:gv} gives
    \[
        \nrm{\beta_{1\fm}^{\scriptscriptstyle(j)}-\beta_{1\fm}^{(m)}}{}^2=\nrm{\omg_{1\fm}^{\scriptscriptstyle(j)}-\omg_{1\fm}^{(m)}}{}^2=
        r^2\nrm{\bz^{\scriptscriptstyle(j)}-\bz^{(m)}}{}^2\ge
        \tfrac12sr^2\asymp\nrm{\omg_2\st}{}^2\frac{p_1}{\np}.
    \]
    Both means shift by the same vector with common noise variance $1+w^2$; hence, using $\nrm{\bz^{\scriptscriptstyle(j)}-\bz^{(m)}}{}^2\le2s$ and $\nt\le\np$,
    \[
        D_{\mathrm{KL}}(\cP_j\,\|\,\cP_m)=
        \frac{\np+\nt}{2(1+w^2)}\,\nrm{\omg_{1\fm}^{\scriptscriptstyle(j)}-\omg_{1\fm}^{(m)}}{}^2\le
        \frac{2\np\,sr^2}{1+w^2}=\frac{2c_0''^2\,s\,w^2}{1+w^2}\le2c_0''^2\,s.
    \]
    Since $\log|\cL(s)|\ge c\,s\log({p_1}/s)\asymp{p_1}\asymp s$, for $c_0''$ small we obtain $D_{\mathrm{KL}}\le\tfrac14\log|\cL(s)|$.
    Lemma \ref{lem:fano} with $\dlt^2=\tfrac14sr^2$ gives
    \[
        \max_{\cP\in\cF\shm}\cP\lrp{\nrm{\h\beta_{1\fm}-\beta_{1\fm}\st}{}^2\ge c_1'\,\nrm{\omg_2\st}{}^2\frac{{p_1}}{\np}}\ge\frac12,
    \]
    where $\cF\shm\subseteq\Pi\shm$ indexes the packing over $\beta_{1\fm}\st$ with $\dlt_{1\fm}\st=\bo$. As in the concluding bound for Theorem \ref{thm:htl-lb}, whichever of $\nrm{\omg_2\st}{}^2p_1/\np$ and $\zeta(\nrm{\dlt_{1\fm}\st}{1})$ is larger is witnessed within $\Pi\shm$; taking $\inf_{\h\beta_{1\fm}}\,\sup_{\cP\in\Pi\shm}$ proves the $\beta_{1\fm}\st$ lower bound in Theorem \ref{thm:hmtl-lb} when $\Gmm=\bP$.

    \emph{Proof of \ref{thm:hmtl-lb}-2.}
    When $\Gmm=\Tht_\infty$, the linear marginal identity $\beta_{1\fm}\st=\beta_1\st+\Gmm_t\beta_2\st$ no longer holds. HmTL uses $\xt$ only, so the misspecified target model is
    \[
        Y_t={\xt}'\beta_1\st+{\bh_t(\xt)}'\beta_2\st+\eps_{t\fm},
    \]
    with confounding ${\bh_t(\xt)}'\beta_2\st$ absorbed into the effective noise. The rest of the proof follows the same steps as before and is therefore omitted.
\end{proof}

\paragraph{Upper bounds for estimation and prediction errors.}
The estimation and prediction of $\h\beta$ and $\h\beta_{1\fm}$ from a single source domain can be considered special cases of transfer learning with multiple source studies. In this regard, we include the proof only for the case where $K\ge1$.

Let us write $\Et=\lrp{\eps_{1,t},...,\eps_{\nt,t}}'$ to denote the vector of target model errors.
The source parameter estimation errors across sources are defined as $\nu_l\sk:=\h\omg_l - \omg_l\skt$ for $l=1,2$ and source $k$ (and $\nu$). 
To streamline the analysis, we define the aggregated source error $\nu:=\frac{1}{K}\sum_{k=1}^K\nu\sk$, which captures the average deviation of our source estimates from the true source parameters.
Through standard Lasso analysis arguments, we obtain the following oracle inequality that bounds the target domain debiasing error:
\begin{align}\label{eq:ht-orclineq}
    -\frac{2}{\nt}\langle\Et+\bR_\bZ(\nu_2+\beta_2\st)-\Xt\nu,\tDt(\hdlt-\bar\dlt\st)\rangle+\frac{1}{\nt}\nrm{\tDt(\hdlt-\bar\dlt\st)}{}^2\le\lmd\sht\lrp{\nrm{\bar\dlt\st}{1}-\nrm{\hdlt}{1}}.
\end{align}
A key distinction from the standard homogeneous transfer learning framework is the presence of the heterogeneity error term $\bR_\bZ$ in the target domain inequality \eqref{eq:ht-orclineq}. This term captures the impact of the feature distribution shift and the feature map estimation error on target-domain estimation.

For the convergence of $\nu$, we need $\np\sk\gtrsim (\sgm_d\sk)^2p$ and $p\gg K$. If $\bh$ is an unknown map and estimated using sieve, recall that $\sgm_{d\eps}$ scales as $\pp\sk$ and we require $\np\sk\gtrsim (\pp\sk)^2p$ for all $k\le K$. Otherwise, we would require $\np\sk\gtrsim(\rp\sk)^2p$. Before we dive into the proof, we exploit the following fact throughout this section:
\begin{proposition}
    Let $\bX_{n\times p}$ be a $\sgm$-Sub-Gaussian ensemble with mean zero and $\bv\in\R^p$ be any random vector that is independent with $\bX$. Then, there exists some constant $c>0$ such that
    \[\Pr\bxs{\frac{1}{n}\nrm{\bX\bv}{}^2\gtrsim\sgm^2\nrm{\bv}{}^2}\le\exp(-cn).\]
\end{proposition}

\subsection{Under non-linear feature map}\label{sec:prf-thm-err}

Now, define the following events:
\begin{align*}
    W\sht & := \Biggl\{\frac{2}{\nt}(\nrm{\tDt'\Et}{\infty}+\nrm{\tDt'\bR_\bZ\beta_2\st}{\infty})\le\frac{\lmd_t}{2},\nrm{\bar\bR_{\Tht_{s,M}}}{F}\le\lmd_{R_\Tht},\nrm{\be_h(\xt)}{}<\frac{1}{\sqrt[4]{\log p_2}}\Biggr\},\\
    V\sht &:= \brcs{\nrm{\nu}{} \le \lmd_s}
\end{align*}
where
\begin{align*}
    \lmd_t=\bar\pi_s\brcs{\set+(\sxt+\nrm{\bar\Dlt_\Tht}{}+\lmd_{R_\Tht})\nrm{\beta_2\st}{}}\sqrt{\frac{\log p}{\nt}},\quad\lmd_s^2=\frac{1}{K}\sum_{k=1}^K\frac{(\pp\sk)^2p}{\np\sk},\\
    \lmd_{R_\Tht}=\frac{1}{K}\sum_{k=1}^K\lrp{{p_2}^{3/4}\log^{3/2}{p_2}\frac{\log^{\post-1}\np\sk}{{\np\sk}}}^{2/3}
\end{align*}
are specified according to the previous discussions on controlling the imputation error and the source parameter estimation error in Lemma \ref{lem:nu-bound}.
\begin{lemma}\label{lem:instrmnt}
    Assume that $\np\sk\gtrsim(\pp\sk)^2p$ for all $k\le K$, $M^2\asymp{p_2}^3(\log p_2)^{2\post-1}$, and $K$ is fixed. Then, if $\log p\le\nt$,
    \[\pr\lrp{W\sht\cap V\sht}\gtrsim1-\frac{1}{\sqrt{p_2}(\log{p_2})^{\post+0.5}}-\exp(-c\log p)-\frac{1}{\sqrt{\log p_2}}.\]
\end{lemma}
\begin{proof}
    Define $\tau_{\t d\eps_t}+\tau_{\t dr_z}:=\tau_{\t d}\lrp{\set+\tau_{r_z}}$.
    Then, since $\lmd_{R_\Tht}=o(1)$ and $\nrm{\be_h(\xt)}{}=o_\pr(1)$, on the corresponding event we have
    \begin{align*}
        \tau_{\t d\eps_t}+\tau_{\t dr_z}\nrm{\beta_2\st}{}\lesssim\brcs{\bar\pi_s+1+o(1)}\brcs{\set+\nrm{\beta_2\st}{}\lrp{\sxt+o(1)+\nrm{\bar\Dlt_\Tht}{}+\lmd_{R_\Tht}}}\\
        \propto\bar\pi_s\brcs{\set+\nrm{\beta_2\st}{}(\sxt+\nrm{\bar\Dlt_\Tht}{}+\lmd_{R_\Tht})}.
    \end{align*}
    First, the condition on $\np\sk$ satisfies the conditions in Corollary \ref{cor:RTht}.
    Using standard arguments verifies that
    \[\pr\bxs{\nrm{\bar\bR_{\Tht_{s,M}}}{}\le\lmd_{R_\Tht}, \frac{2}{\nt}\nrm{\tDt'\Et}{\infty}\lesssim\tau_{\t d\eps_t}\sqrt{\frac{\log p}{\nt}}}\gtrsim1-\exp\lrp{-c'\log p}-\frac{1}{\sqrt{p_2}(\log p_2)^{\post+0.5}}\]
    for some large $c'>0$.
    Similarly, with the same high probability, we have
    \[\nrm{\bar\bR_{\Tht_{s,M}}}{}\le\lmd_{R_\Tht}, \frac{2}{\nt}\nrm{\tDt'\bR_\bZ\beta_2\st}{\infty}\lesssim\tau_{\t dr_z}\nrm{\beta_2\st}{}\sqrt{\frac{\log p}{\nt}}.\]
    Recall that the order of $\tau_{\t dr_z}$ is $\bar\pi_s(\sxt+\nrm{\bar\Dlt_\Tht}{}+o(1))$.

    For the order of $\lmd_s$, the condition on $\np\sk$ for all $k\le K$ satisfies the condition in Lemma \ref{lem:nu-bound}:
    \[\nrm{\nu}{}^2\lesssim\frac{1}{K}\sum_{k=1}^K\bigl[(\sop\sk)^2(\pp\sk)^2+ (\sxp\sk)^2\bigr]\,(\sep\sk)^2\,\frac{p}{n_s\sk}\]
    by the order of $\sgm_{d\eps s}\sk=(\sop\sk(1+\pp\sk)+\sxp\sk)\sep\sk$.
    Hence $\pr(V\sht)\ge 1-K\exp(-c\log p)$ for some $c>0$ by the definition of $\lmd_s$.
    Combining this with the bound for $W\sht$ above and applying the union bound yields the conclusion.
\end{proof}
From now on, we assume that $\sxp\sk,\sep\sk$ are fixed for all $k\le K$.
\begin{proof}[\textbf{Proof of Theorem \ref{thm:err1}}]
    We adapt the arguments in \cite{li2022transfer}.
    Write $\br_\dlt:=\hdlt-\bar\dlt\st$.
    On $W\sht\cap V\sht$, by \eqref{eq:ht-orclineq} and $\bR_\bZ(\nu_2+\beta_2\st)-\Xt\nu=-\tDt\nu+\bR_\bZ\beta_2\st$ we have
    \begin{align*}
        \frac{1}{\nt}\nrm{\tDt\br_\dlt}{}^2\le&\lmd_t\lrp{\nrm{\bar\dlt\st}{1}-\nrm{\hdlt}{1}}+\frac{2}{\nt}\vrt{\langle\br_\dlt,\tDt'(\Et+\bR_\bZ\beta_2\st-\tDt\nu)\rangle} \\
        &\quad\le\lmd_t\lrp{\nrm{\bar\dlt\st}{1}-\nrm{\hdlt}{1}}+\frac{2}{\nt}\nrm{\br_\dlt}{1}\nrm{\tDt'(\Et+\bR_\bZ\beta_2\st)}{\infty}+\frac{2}{n_t}\vrt{\langle\tDt\br_\dlt,\tDt\nu\rangle}
    \end{align*}
    and the last two terms are bounded by
    \[\frac{\lmd_t}{2}\nrm{\br_\dlt}{1}+\frac{2}{\nt}\nrm{\tDt\nu}{}^2+\frac{1}{2\nt}\nrm{\tDt\br_\dlt}{}^2\]
    using the fact that $2|\langle\ba,\bdb\rangle|\le2\nrm{\ba}{}^2+\nrm{\bdb}{}^2/2$. Hence,
    \begin{align*}
        \frac{1}{2\nt}\nrm{\tDt\br_\dlt}{}^2\le\lmd_t\lrp{\nrm{\bar\dlt\st}{1}-\nrm{\hdlt}{1}}+\frac{\lmd_t}{2}\nrm{\br_\dlt}{1}+\frac{2}{\nt}\nrm{\tDt\nu}{}^2 \\
        \le2\lmd_t\nrm{\bar\dlt\st}{1}-\frac{\lmd_t}{2}\nrm{\br_\dlt}{1}+\frac{2}{\nt}\nrm{\tDt\nu}{}^2.
    \end{align*}
    \begin{enumerate}
        \item If $2\lmd_t\nrm{\bar\dlt\st}{1}\ge\frac{2}{\nt}\nrm{\tDt\nu}{}^2$, then $\nrm{\br_\dlt}{1}\le8\nrm{\bar\dlt\st}{1}$ and
        \[\frac{1}{2\nt}\nrm{\tDt\br_\dlt}{}^2\le4\lmd_t\nrm{\bar\dlt\st}{1}.\]
        By the RSC Lemma, since $\tau_{\t d}\asymp\bar\pi_s$, this implies
        \[\nrm{\br_\dlt}{}^2\lesssim\nrm{\br_\dlt}{1}^2\frac{\tau_{\t d}^2\log p}{\nt}+\lmd_t\nrm{\bar\dlt\st}{1}\lesssim\lrp{\frac{\bar\pi_s^2\log p}{\nt}\nrm{\bar\dlt\st}{1}+\lmd_t}\nrm{\bar\dlt\st}{1}\]
        hence since $\frac{\bar\pi_s^2\log p}{\nt}\nrm{\bar\dlt\st}{1}=\cO(\lmd_t)$,
        \[\nrm{\br_\dlt}{}^2\lesssim\nrm{\bar\dlt\st}{1}(\nrm{\bar\dlt\st}{1}\wedge\lmd_t).\]
        \item If $2\lmd_t\nrm{\bar\dlt\st}{1}<\frac{2}{\nt}\nrm{\tDt\nu}{}^2$, then with probability at least $1-\exp(-c\nt)$, we have
        \[\frac{1}{2\nt}\nrm{\tDt\br_\dlt}{}^2\le\frac{4}{\nt}\nrm{\tDt\nu}{}^2\lesssim\tau_{\t d}^2\nrm{\nu}{}^2,~ \lmd_t\nrm{\br_\dlt}{1}\lesssim\frac{1}{\nt}\nrm{\tDt\nu}{}^2\lesssim\tau_{\t d}^2\nrm{\nu}{}^2\]
        and by RSC Lemma again and using $\tau_{\t d}\asymp\bar\pi_s$, we have
        \[\nrm{\br_\dlt}{}^2\lesssim(\bar\pi_s^2+1)\lrp{\nrm{\nu}{}^2+\nrm{\br_\dlt}{1}^2\frac{\log p}{\nt}}\]
        and
        \[\lmd_t\nrm{\br_\dlt}{1}\lesssim(\bar\pi_s^2+1)\lmd_s^2=(\bar\pi_s^2+1)\sum_{k=1}^K\frac{(\pp\sk)^2 p}{K\np\sk}.\]
        So provided that $\max_{k\le K}\pp\sk=\cO(1)$,
        \begin{align*}
            \nrm{\br_\dlt}{}^2&\lesssim(\bar\pi_s^2+1)\lmd_s^2\brcs{1+(\bar\pi_s^4+1)\lmd_s^2\frac{\log p}{\lmd_t^2\nt}}.
        \end{align*}
    \end{enumerate}
    This proves the error bound in Theorem \ref{thm:err1}.
    For the prediction error bound, note that
    \[\E(\by_t^\fn|\Xt^\fn)-\tDt^\fn\h\beta=[\E(\Dt^\fn|\Xt^\fn)-\tDt^\fn]\beta\st-\tDt^\fn(\h\beta-\beta\st)\]
    where $[\E(\Dt^\fn|\Xt^\fn)-\tDt^\fn]\beta\st=(\bR_\bZ^\fn-\Xit^\fn)\beta_2\st=[\bE_{H_t}^\fn+\Psi_M(\Xt^\fn)(\bar\Dlt_\Tht+\bar\bR_{\Tht_{s,M}})]\beta_2\st$.
    Hence
    \[\lrnrm{\E(\by_t^\fn|\Xt^\fn)-\tDt^\fn\h\beta}{}^2\lesssim o_\pr(n_t^\fn)+(\nrm{\bar\Dlt_\Tht}{}^2+o_\pr(1))\nrm{\beta_2\st}{}^2\]
    on $W\sht\cap V\sht$ under the given asymptotic conditions.
    Also,
    \[\pr\lrp{\frac{1}{n_t^\fn}\nrm{\tDt^\fn(\h\beta-\beta\st)}{}^2\ge2\tau_{\t d}^2\nrm{\h\beta-\beta\st}{}^2}\le\exp\lrp{-cn_t^\fn}\]
    as $\h\beta$ is independent with $\tDt^\fn$.
    Noting that $\tau_{r_z}\asymp\sxt+\nrm{\bar\Dlt_\Tht}{}$ and $\tau_{\t d}^2\asymp\bar\pi_s^2$ on $W\sht\cap V\sht$,
    \begin{align*}
        \frac{1}{n_t^\fn}\lrnrm{\E(\by_t^\fn|\Xt^\fn)-\tDt^\fn\h\beta}{}^2
        \lesssim o_\pr(1)+\nrm{\bar\Dlt_\Tht}{}^2\nrm{\beta_2\st}{}^2+\bar\pi_s^2\nrm{\h\beta-\beta\st}{}^2.
    \end{align*}
\end{proof}

\subsubsection{Theorem \ref{thm:hm-err}: bounds for $\h\beta_{1\fm}-\beta_1\st$} 
\begin{theorem}\label{thm:sp-hm-err}
    Assume that $\pp+\sxp + \sep = \cO(1)$, $\log{p_1}\le\nt$ and $\np\gg{p_1}$. Further assume that
    \[\nrm{\dlt_1\st}{1}=\cO\bxs{\brcs{\set+(\sxt+\pt)\nrm{\beta_2\st}{}}\sqrt{\frac{\nt}{\log{p_1}}}}.\]
    Then, with probability at least $1-c\exp(-c'\log{p_1})$,
    \begin{align*}
        \nrm{\h\beta_{1\fm}-\beta_1\st}{}^2\lesssim\lrp{\nrm{\omg_2\st}{}^2+p_1/\np}\bxs{1+\frac{\nrm{\omg_2\st}{}^2+p_1/\np}{\brcs{\set+(\sxt+\pt)\nrm{\beta_2\st}{}}^2}}\\
        +\nrm{\dlt_1\st}{1}\bxs{\nrm{\dlt_1\st}{1}\wedge\brcs{\set+(\sxt+\pt)\nrm{\beta_2\st}{}}\sqrt{\frac{\log{p_1}}{\nt}}}.
    \end{align*}
    Further assume that we observe $n_t^\fn<\infty$ new observations. Then, with probability at least $1-c\exp(-c'n_t^\fn)$,
    \begin{align*}
        \frac{1}{n_t^\fn}\lrnrm{\E(\by_t^\fn|\Xt^\fn)-\Xt^\fn\h\beta_{1\fm}}{}^2
        \lesssim\pt^2\nrm{\beta_2\st}{}^2 + \nrm{\h\beta_{1\fm}-\beta_1\st}{}^2.
    \end{align*}
\end{theorem}
Recall that $\Zt\beta_2\st=\brcs{\bH_t(\Xt)+\Xit}\beta_2\st$. So the square loss for the homogeneous transfer learning estimator $\h\beta_{1\fm}$ can be written as
\begin{align*}
    \frac{1}{\nt}\lrnrm{\by_t-\Xt\beta_1}{}^2&=\frac{1}{\nt}\lrnrm{\Etm+\Xt(\beta_1\st-\beta_1)}{}^2
\end{align*}
where $\Etm:=\Et+\Zt\beta_2\st$.
So, the correction term $\h\dlt_1\shm$ is obtained as the solution to:
\begin{align*}
    \h\dlt_1\shm = \underset{\dlt_1\in\R^{p_1}}{\arg\min}\left\{\frac{1}{\nt}\lrnrm{\Etm+\Xt(\bar\dlt\st_1-\dlt_1-\nu_1)}{}^2 + \lmd_{1\fm}\nrm{\dlt_1}{1}\right\}.
\end{align*}
This leads to the following oracle inequality for $\h\dlt_1\shm$:
\begin{align}\label{eq:hm-orclineq-nlfm}
    -\frac{2}{\nt}\langle\Etm+\Xt\nu_1,\Xt\br_{\dlt_1}\rangle+\frac{1}{\nt}\nrm{\Xt\br_{\dlt_1}}{}^2\le\lmd_{1\fm}\lrp{\nrm{\bar\dlt\st_{1}}{1}-\nrm{\h\dlt_1\shm}{1}}.
\end{align}

Define the following events:
\begin{align*}
    W\shm := \brcs{\frac{2}{\nt}\nrm{\Xt'\Etm}{\infty}\le\frac{\lmd_t}{2}},~~V\shm := \brcs{\nrm{\nu_1}{} \le \lmd_s}
\end{align*}
where
\begin{align*}
    \lmd_t=\brcs{\set+\nrm{\beta_2\st}{}(\pt+\sxt)}\sqrt{\frac{\log{p_1}}{\nt}},~~ \lmd_s^2=\sum_{k=1}^K\frac{(\pp\sk+\sxp\sk)^2\nrm{\omg_2\skt}{}^2}{K}+\sum_{k=1}^K\frac{{p_1}}{K\np\sk}
\end{align*}
by Lemma \ref{lem:nu-bound}.
\begin{lemma}
    Assume that $\np\sk\gtrsim (\pp\sk)^2 p$ for all $k\le K$. Then, if $\log{p_1}\le\nt$,
    \[\pr\lrp{W\shm}\ge1-c\exp\lrp{-c'\log{p_1}}.\]
\end{lemma}
\begin{proof}
    Define $\tau_{x\eps_\fm}:=\sot\lrp{\set+\stt\nrm{\beta_2\st}{}}\asymp\set+(\sxt+\pt)\nrm{\beta_2\st}{}$.
    Using standard arguments again verifies that
    \[\pr\bxs{\frac{2}{\nt}\nrm{\Xt'\Etm}{\infty}\lesssim\tau_{x\eps_\fm}\sqrt{\frac{\log{p_1}}{\nt}}}\ge1-c\exp\lrp{-c'\log{p_1}}\]
    for some large $c'>0$. Using the same argument, we obtain the conclusion.
\end{proof}
\begin{proof}
    [\textbf{Proof of Theorem \ref{thm:hm-err}}]
    On $W\shm$, by \eqref{eq:hm-orclineq-nlfm}
    \begin{align*}
        \frac{1}{\nt}\nrm{\Xt\br_{\dlt_1}}{}^2\le\lmd_t\lrp{\nrm{\bar\dlt_1\st}{1}-\nrm{\hdhm}{1}}+\frac{2}{\nt}\vrt{\langle\br_{\dlt_1},\Xt'(\Et+\Zt\beta_2\st-\Xt\nu_1)\rangle}
    \end{align*}
    hence by similar arguments as in the previous section,
    \begin{align*}
        \frac{1}{\nt}\nrm{\Xt\br_{\dlt_1}}{}^2\le2\lmd_t\nrm{\bar\dlt_1\st}{1}-\frac{\lmd_t}{2}\nrm{\br_{\dlt_1}}{1}+\frac{2}{\nt}\nrm{\Xt\nu_1}{}^2.
    \end{align*}
    using the same arguments as in proving $\h\beta$, we obtain the estimation error bound:
    \begin{enumerate}
        \item If $2\lmd_t\nrm{\bar\dlt_1\st}{1}\ge\frac{2}{\nt}\nrm{\Xt\nu_1}{}^2$, then $\nrm{\br_{\dlt_1}}{1}\le8\nrm{\bar\dlt_1\st}{1}$ and
        \[\frac{1}{\nt}\nrm{\Xt\br_{\dlt_1}}{}^2\le4\lmd_t\nrm{\bar\dlt_1\st}{1}.\]
        By the RSC Lemma,
        \begin{align*}
            \nrm{\br_{\dlt_1}}{}^2\lesssim\nrm{\br_{\dlt_1}}{1}^2\sot^2\frac{\log{p_1}}{\nt}+\lmd_t\nrm{\bar\dlt_1\st}{1}\lesssim\lrp{\frac{\log{p_1}}{\nt}\nrm{\bar\dlt_1\st}{1}+\lmd_t}\nrm{\bar\dlt_1\st}{1} \\
            =\cO(\lmd_t\nrm{\bar\dlt_1\st}{1})
        \end{align*}
        provided that $\frac{\log{p_1}}{\nt}\nrm{\bar\dlt_1\st}{1}=\cO(\lmd_t)$.
        Therefore,
        \[\nrm{\br_{\dlt_1}}{}^2\lesssim\nrm{\bar\dlt_1\st}{1}(\nrm{\bar\dlt_1\st}{1}\wedge\lmd_t).\]
        \item If $2\lmd_t\nrm{\bar\dlt_1\st}{1}<\frac{2}{\nt}\nrm{\Xt\nu_1}{}^2$, then 
        \[\frac{1}{\nt}\nrm{\Xt\br_{\dlt_1}}{}^2\le\frac{4}{\nt}\nrm{\Xt\nu_1}{}^2\lesssim\nrm{\nu_1}{}^2\]
        and by RSC again
        \[\nrm{\br_{\dlt_1}}{}^2\lesssim\nrm{\nu_1}{}^2+\nrm{\br_{\dlt_1}}{1}^2\sot^2\frac{\log{p_1}}{\nt}\]
        and
        \[\lmd_t\nrm{\br_{\dlt_1}}{1}\lesssim\frac{1}{\nt}\nrm{\Xt\nu_1}{}^2\lesssim\nrm{\nu_1}{}^2.\]
        So,
        \[\nrm{\br_{\dlt_1}}{}^2\lesssim\lmd_s^2+\frac{\lmd_s^4}{\lmd_t^2}\frac{\log{p_1}}{\nt}\lesssim\lmd_s^2\bxs{1+\frac{\lmd_s^2}{\brcs{\set+(\sxt+\pt)\nrm{\beta_2\st}{}}^2}}.\]
    \end{enumerate}
    Summarizing the rates, we have
    \begin{align*}
        \nrm{\h\beta_{1\fm}-\beta_1\st}{}^2&\lesssim\lmd_s^2\bxs{1+\frac{\lmd_s^2}{\brcs{\set+(\sxt+\pt)\nrm{\beta_2\st}{}}^2}}+\nrm{\bar\dlt_1\st}{1}(\nrm{\bar\dlt_1\st}{1}\wedge\lmd_t)
    \end{align*}
    for \[\lmd_s^2=\frac{1}{K}\sum_{k=1}^K\brcs{(\pp\sk+\sxp\sk)^2\nrm{\omg_2\skt}{}^2+\frac{{p_1}}{\np\sk}}.\]
    Next, note that
    \[\lrnrm{\E(\by_t^\fn|\Xt^\fn)-\Xt^\fn\h\beta_{1\fm}}{}^2\lesssim \lrnrm{\bH_t(\Xt^\fn)\beta_2\st}{}^2+\nrm{\Xt^\fn(\h\beta_{1\fm}-\beta_1\st)}{}^2\]
    and as $\beta_2\st$ is a constant vector and $\beta_1\shm$ is independent of $\Xt^\fn$, with probability at least $1-\exp(-cn_t^\fn)$, we have
    \begin{align*}
        \frac{1}{n_t^\fn}\lrnrm{\bH_t(\Xt^\fn)\beta_2\st}{}^2=\frac{1}{n_t^\fn}\lrnrm{\Psi_\infty(\Xt^\fn)\Tht_{t,\infty}\beta_2\st}{}^2&\lesssim \pt^2\nrm{\beta_2\st}{}^2,\\
        \frac{1}{n_t^\fn}\nrm{\Xt^\fn(\h\beta_{1\fm}-\beta_1\st)}{}^2&\lesssim\nrm{\h\beta_{1\fm}-\beta_1\st}{}^2
    \end{align*}
    so
    \begin{align*}
        \frac{1}{n_t^\fn}\lrnrm{\E(\by_t^\fn|\Xt^\fn)-\Xt^\fn\h\beta_{1\fm}}{}^2
        \lesssim \pt^2\nrm{\beta_2\st}{}^2 + \nrm{\h\beta_{1\fm}-\beta_1\st}{}^2.
    \end{align*}
\end{proof}

\subsection{Under linear feature map}\label{sec:prf-lfm-thm-err}

When $\E(\bz|\bx)$ is assumed to be linear (i.e., $\bz=\bP'\bx+\xi$), then we define
\begin{align*}
    W_\cL\sht & := \brcs{\frac{2}{\nt}(\nrm{\tDt'\Et}{\infty}+\nrm{\tDt'\bR_\bZ\beta_2\st}{\infty})\le\frac{\lmd_{t,\cL}}{2},\nrm{\bar\bR_{\Pp}}{}\le\lmd_{\bP,F}},\\
    V_\cL\sht & :=\brcs{\nrm{\nu}{}<\lmd_{s,\cL}}
\end{align*}
where
\begin{align*}
    \lmd_{t,\cL}&=\lrp{\bar\rp+\lmd_{\bP,F}}\brcs{\set+\nrm{\beta_2\st}{}(\nrm{\bar\Dlt_\bP}{}+\sxt)}\sqrt{\frac{\log p_1}{\nt}},\quad\lmd_{\bP,F}^2=\frac{1}{K}\sum_{k=1}^K\frac{s_\bP\sk\log{p_1}}{\np\sk},\\
    \lmd_{s,\cL}^2&=\sum_{k=1}^K\frac{p}{K\np\sk}
\end{align*}
by the results in Lemma \ref{lem:nu-bound} and the control of the imputation error.
The following result is for the case when $\bh$ is assumed to be a linear map of $\bx$.
\begin{lemma}
    Assume that $\np\sk\gtrsim(\rp\sk)^2p+s_\bP\sk\log{p_1}$ for all $k\le K$.
    Then,
    \[\pr\lrp{W_\cL\sht\cap V_\cL\sht}\ge1-c\exp\lrp{-c'\log p}.\]
\end{lemma}
\begin{proof}
    Note that $\tDt=\Xt[\bI_{p_1},\h\Pp]$ whose rows are i.i.d. $\sot(\bar\rp+\nrm{\bR_{\Pp}}{F})$-sub-gaussian ensemble. The rest of the proof follows the same steps as the previous section.
\end{proof}
\begin{proof}[\textbf{Proof of Theorem \ref{thm:lfm-err}}]
    Now, on $W_\cL\sht$, we start from
    \begin{align*}
        \frac{1}{\nt}\nrm{\tDt\br_\dlt}{}^2
        \le2\lmd_{t,\cL}\nrm{\bar\dlt\st}{1}-\frac{\lmd_{t,\cL}}{2}\nrm{\br_\dlt}{1}+\frac{2}{\nt}\nrm{\tDt\nu}{}^2.
    \end{align*}
    \begin{enumerate}
        \item If $2\lmd_{t,\cL}\nrm{\bar\dlt\st}{1}\ge\frac{2}{\nt}\nrm{\tDt\nu}{}^2$, then $\nrm{\br_\dlt}{1}\le8\nrm{\bar\dlt\st}{1}$ and
        \[\frac{1}{\nt}\nrm{\tDt\br_\dlt}{}^2\le4\lmd_{t,\cL}\nrm{\bar\dlt\st}{1}.\]
        By the RSC Lemma \eqref{eq:rsc-lfm}, this implies
        \[\nrm{\br_\dlt}{}^2\lesssim\nrm{\br_\dlt}{1}^2(\bar m_\bP+\nrm{\bar\bP_s}{1}^2)\frac{\log p_1}{\nt}+\lmd_{t,\cL}\nrm{\bar\dlt\st}{1}\lesssim\brcs{(\bar m_\bP+\nrm{\bar\bP_s}{1}^2)\frac{\log p_1}{\nt}\nrm{\bar\dlt\st}{1}+\lmd_{t,\cL}}\nrm{\bar\dlt\st}{1}\]
        hence provided that $(\bar m_\bP+\nrm{\bar\bP_s}{1}^2)\frac{\log p_1}{\nt}\nrm{\bar\dlt\st}{1}=\cO(\lmd_{t,\cL})$,
        \[\nrm{\br_\dlt}{}^2\lesssim\nrm{\bar\dlt\st}{1}(\nrm{\bar\dlt\st}{1}\wedge\lmd_{t,\cL}).\]
        \item If $2\lmd_{t,\cL}\nrm{\bar\dlt\st}{1}<\frac{2}{\nt}\nrm{\tDt\nu}{}^2$, then 
        \[\frac{1}{\nt}\nrm{\tDt\br_\dlt}{}^2\le\frac{4}{\nt}\nrm{\tDt\nu}{}^2,~ \lmd_{t,\cL}\nrm{\br_\dlt}{1}\lesssim\frac{1}{\nt}\nrm{\tDt\nu}{}^2\]
        and by RSC and $\tau_{\t d}\asymp\bar\rp$,
        \[\nrm{\br_\dlt}{}^2\lesssim\bar\rp^2\lmd_{s,\cL}^2+\nrm{\br_\dlt}{1}^2(\bar m_\bP+\nrm{\bar\bP_s}{1}^2)\frac{\log p_1}{\nt}\]
        and
        \[\lmd_{t,\cL}\nrm{\br_\dlt}{1}\lesssim\bar\rp^2\nrm{\nu}{}^2\le\bar\rp^2\lmd_{s,\cL}^2.\]
        So,
        \begin{align*}
            \nrm{\br_\dlt}{}^2&\lesssim\bar\rp^2\lmd_{s,\cL}^2\brcs{1+\frac{\bar\rp^2\lmd_{s,\cL}^2}{\lmd_{t,\cL}^2}(\bar m_\bP+\nrm{\bar\bP_s}{1}^2)\frac{\log p_1}{\nt}}.
        \end{align*}
    \end{enumerate}
    Plugging in $\lmd_{t,\cL}$ and $\lmd_{s,\cL}$, we have
    \begin{align*}
        \nrm{\h\beta-\beta\st}{}^2&\lesssim\bar\rp^2\lmd_{s,\cL}^2\brcs{1+\frac{\bar\rp^2\lmd_{s,\cL}^2}{\lmd_{t,\cL}^2}(\bar m_\bP+\nrm{\bar\bP_s}{1}^2)\frac{\log p_1}{\nt}}+\nrm{\bar\dlt\st}{1}(\nrm{\bar\dlt\st}{1}\wedge\lmd_{t,\cL}) \\
        &\lesssim\sum_{k=1}^K\frac{p}{K\np\sk}\bxs{1\vee\frac{(\bar m_\bP+\nrm{\bar\bP_s}{1}^2)(\bar\rp+\lmd_{\bP,F})^{-2}}{\brcs{\set+\nrm{\beta_2\st}{}\lrp{\nrm{\bar\Dlt_\bP}{}+\sxt}}^2}\sum_{k=1}^K\frac{p}{K\np\sk}} \\
        &\qquad+\nrm{\bar\dlt\st}{1}\bxs{\nrm{\bar\dlt\st}{1}\wedge\lrp{\bar\rp+\lmd_{\bP,F}}\brcs{\set+\nrm{\beta_2\st}{}\lrp{\nrm{\bar\Dlt_\bP}{}+\sxt}}\sqrt{\frac{\log p_1}{\nt}}}.
    \end{align*}
    Next, note that
    \[\lrnrm{\E(\by_t^\fn|\Xt^\fn)-\tDt^\fn\h\beta}{}^2\lesssim \lrnrm{\bR_\bZ^\fn\beta_2\st}{}^2+\nrm{\tDt^\fn(\h\beta-\beta\st)}{}^2\]
    so as $\tau_{r_z}\asymp\sxt+{\nrm{\bar\Dlt_\bP}{}}$,
    \begin{align*}
        \frac{1}{n_t^\fn}\lrnrm{\E(\by_t^\fn|\Xt^\fn)-\tDt^\fn\h\beta}{}^2
        \lesssim({{\nrm{\bar\Dlt_\bP}{}}}^2+\sxt^2)\nrm{\beta_2\st}{}^2+\bar\rp^2\nrm{\h\beta-\beta\st}{}^2.
    \end{align*}
\end{proof}

\subsubsection{Theorem \ref{thm:lfm-hm-err}-1: bounds for $\h\beta_{1\fm}-\beta_1\st$} Now, define the following events:
\begin{align*}
    W_\cL\shm := \brcs{\frac{2}{\nt}\nrm{\Xt'\Etm}{\infty}\le\frac{\lmd_{t,\cL}}{2}},~~V_\cL\shm := \brcs{\nrm{\nu_1}{} \le \lmd_{s,\cL}}
\end{align*}
where
\begin{align*}
    \lmd_{t,\cL}=\brcs{\set+(\sxt+\rt)\nrm{\beta_2\st}{}}\sqrt{\frac{\log{p_1}}{\nt}}, \lmd_{s,\cL}^2=\sum_{k=1}^K\lrp{\frac{\nrm{\omg_2\skt}{}^2}{K}+\frac{{p_1}}{K\np\sk}}
\end{align*}
from Lemma \ref{lem:nu-bound}.
\begin{lemma}
    Assume that $\np\sk\gtrsim{p_1}$ for all $k\le K$. Then,
    \[\pr\lrp{W_\cL\shm}\ge1-c\exp\lrp{-c'\log p}.\]
\end{lemma}
\begin{proof}
    [\textbf{Proof of Theorem \ref{thm:lfm-hm-err}-1.}]
    The proof follows the same steps as in \ref{thm:hm-err}.
\end{proof}

\subsubsection{Theorem \ref{thm:lfm-hm-err}-2: bounds for $\h\beta_{1\fm}-\beta_{1\fm}\st$} 
$\hdhm$ satisfies an oracle inequality for $\br_{\dlt_{1\fm}}:=\hdhm-\bar\dlt_{1\fm}\st$:
\begin{align}\label{eq:hm-orclineq}
    -\frac{2}{\nt}\langle\Etm+\Xt(\overline{\Dlt_\bP\omg_2\st}-\nu_{1\fm}),\Xt\br_{\dlt_{1\fm}}\rangle+\frac{1}{\nt}\nrm{\Xt\br_{\dlt_{1\fm}}}{}^2\le\lmd_{1\fm}\lrp{\nrm{\bar\dlt_{1\fm}\st}{1}-\nrm{\hdhm}{1}}
\end{align}
where $\overline{\Dlt_\bP\omg_2\st}:=\frac{1}{K}\sum_{k=1}^K\Dlt_\bP\sk\omg_2\skt$.
Define the following events:
\begin{align*}
    W_\cL\shm := \brcs{\frac{2}{\nt}(\nrm{\Xt'\Etm}{\infty}+\nrm{\Xt'\Xt\overline{\Dlt_\bP\omg_2\st}}{\infty})\le\frac{\lmd_{t,\cL}}{2}},~~V_\cL\shm := \brcs{\nrm{\nu_{1\fm}}{} \le \lmd_{s,\cL}}
\end{align*}
where
\begin{align*}
    \lmd_{t,\cL}&:=\nrm{\Sgot\overline{\Dlt_\bP\omg_2\st}}{\infty}+\lrp{\set+\sxt\nrm{\beta_2\st}{}+\nrm{\overline{\Dlt_\bP\omg_2\st}}{}}\sqrt{\frac{\log{p_1}}{\nt}},\\
    \lmd_{s,\cL}^2&:=\sum_{k=1}^K\lrp{\nrm{\omg_2\skt}{}+1}^2\frac{{p_1}}{K\np\sk}
\end{align*}
selected according to Lemma \ref{lem:nu-bound}.
\begin{lemma}
    Assume that $\np\sk\gtrsim{p_1}$ for all $k\le K$. Then,
    \[\pr\lrp{W_\cL\shm}\ge1-c\exp\lrp{-c'\log{p_1}}.\]
\end{lemma}
\begin{proof}[\textbf{Proof of Theorem \ref{thm:lfm-hm-err}-2.}]
    This can be shown similarly to Theorem \ref{thm:lfm-hm-err}-1.
\end{proof}

\subsection{Negative transfer defense (Section \ref{sec:ntd})}
The following regularity assumption is made for the theory of the SAND algorithm.
\begin{assumption}\label{asm:ntd-desgn}
    $\max_{k\in[K]}\sep\sk$, $\max_{k\in[K]}\sxp\sk$, $\set,\sxt=\cO(1)$, $\max_{k\in[K]}\E\nrm{\bh_s\sk(\xp\sk)}{}$, and $\sot,\max_{k\in[K]}\sop\sk=\cO(1)$. Assume that $\lmd_{\min}(\Sgot),\lmd_{\max}(\Sgot)$ are positive absolute constants.
    Moreover, $\log{p_1}\lesssim\nt$, and $\np\sk\gtrsim{p_1}+s_\bP\sk\log p_1$ for each $k$.
\end{assumption}

Above assumption ensures that the variance proxy of the composite target noise scales as $\tau_{\eps_\fm} = \tau_\eps + \tau_\xi \|\beta_2\st\| = \mathcal{O}(\|\beta_2\st\|)$, which implies $\mathcal{R}(\beta_{1\fm}\st) \le \tau_{\eps_\fm}^2 = \mathcal{O}(\|\beta_2\st\|^2)$ when $\beta_2\st$ is a dense parameter vector.

\begin{lemma}[concentration of $\check\omg_{1\fm}\sk$]\label{lem:htl-theta-conc}
    Fix $k\in[K]$.
    Suppose $\np\sk\gtrsim p+s_\bP\sk\log p_1$, the source designs satisfy Assumption \ref{asm:ntd-desgn}.
    Define $\check\omg_{1\fm}\sk:=\h\omg_1\sk+\h{\Pp\sk}\h\omg_2\sk$ and $\omg_{1\fm}\skt:=\omg_1\skt+\Pp\sk\omg_2\skt$.
    Then, with probability at least $1-cp_2\exp(-c'p_1)$,
    \[
        \nrm{\check\omg_{1\fm}\sk-\omg_{1\fm}\skt}{}
        \lesssim
        \check\lmd\sk
        :=\sqrt{\frac{p}{\np\sk}}+\nrm{\omg_2\skt}{}\sqrt{\frac{s_\bP\sk\log p_1}{\np\sk}}.
    \]
\end{lemma}
\begin{proof}[Proof of Lemma \ref{lem:htl-theta-conc}]
    Decompose
    \begin{align*}
        \check\omg_{1\fm}\sk-\omg_{1\fm}\skt
        =[\bI_{p_1}\,\Pp\sk]\nu
        +\lrp{\h{\Pp\sk}-\Pp\sk}(\nu_2\sk+\omg_2\skt)
    \end{align*}
    where $\nu_2\sk:=\h\omg_2\sk-\omg_2\skt$ (this is defined differently from $\nu_1\sk=\h\omg_{1\fm}\sk-\omg_1\skt$).
    By Lemma \ref{lem:nu-bound},
    \[
        \pr\bxs{\nrm{\nu\sk}{}
        \lesssim\sqrt{\frac{p}{\np\sk}}}\gtrsim1-\exp(-cp).
    \]
    The 2---norm of the first term above is therefore $\cO\lrp{\sqrt{p/\np\sk}}$ with high probability.
    The column-wise lasso rate gives
    $\lrnrm{\lrp{\h{\Pp\sk}-\Pp\sk}\omg_2\skt}{}\lesssim\lmd_{\bP}\sk\nrm{\omg_2\skt}{}$ when $\np\sk\gtrsim s_\bP\sk\log{p_1}$.
    The remaining term converges faster under the same condition.
\end{proof}

\begin{assumption}\label{asm:ntd-cv}
    A candidate penalty grid $\Lambda$ satisfies $\log|\Lambda|=\cO(\log{p_1})$ and contains the null penalty $\lmd_{\max}$ (at which $\t\beta_{1\fm}^{\scriptscriptstyle(j)}(\lmd_{\max})=0$), and $\kappa$ is fixed.
    There is a deterministic $\lmd_0\asymp\sqrt{\log{p_1}/\nt}$ with $\lmd_{\rm1SE}\ge\lmd_0$ with probability at least $1-c\exp(-c'\nt\t\eta^2)$.
\end{assumption}

The $1$-SE rule selects a penalty within one standard error above $\lmd_{\min}$, which in the ${p_1}>\nt$ regime is bounded below at the order $\sqrt{\log{p_1}/\nt}$.
The following lemma proves that cross-validated score concentrates around the achievable baseline risk $\bar\cR(\lmd_{\rm1SE})$ in \eqref{eq:Rft}, \emph{uniformly} over the data-dependent choice of $\lmd_{\rm1SE}$.

\begin{lemma}[Baseline cross-validation concentration]\label{lem:cv-base}
    Under Assumptions \ref{asm:ntd-desgn} and \ref{asm:ntd-cv}, there exist constants $c,c_1,c_2>0$ such that
    \[
        \pr\lrp{\vrt{S_t - \bar\cR(\lmd_{\rm1SE})} > c\brcs{\tau_{\eps_\fm}^2+\bar\cR(\lmd_{\rm1SE})}\sqrt{\frac{\log{p_1}}{\nt}}}
        \le
        c_1\exp\lrp{-c_2\log{p_1}}.
    \]
\end{lemma}
\begin{proof}[Proof of Lemma \ref{lem:cv-base}]
    Fix a fold $j$ and condition on its training set $\cT_j:=\cD_t\setminus\cV_j$, so that $\t\beta_{1\fm}^{\scriptscriptstyle(j)}(\lmd)$ is deterministic and the validation points $\brcs{(\bx_{t,i},Y_{t,i})}_{i\in\cV_j}$ form an independent sample.
    The validation residuals
    \[
        u_i(\lmd) := Y_{t,i} - {\bx_{t,i}}'\t\beta_{1\fm}^{\scriptscriptstyle(j)}(\lmd)
        = \eps_{t m,i} + {\bx_{t,i}}'\lrp{\beta_{1\fm}\st - \t\beta_{1\fm}^{\scriptscriptstyle(j)}(\lmd)}
    \]
    are i.i.d.\ given $\cT_j$, with conditional second moment
    $\cR_j(\lmd):=\E\brcs{u_i(\lmd)^2|\cT_j}=\cR\lrp{\t\beta_{1\fm}^{\scriptscriptstyle(j)}(\lmd)}$.
    Since $u_i(\lmd)$ is conditionally $\tau_j(\lmd)$-sub-Gaussian with
    $\tau_j(\lmd):=\tau_{\eps_\fm}+\sot\nrm{\t\beta_{1\fm}^{\scriptscriptstyle(j)}(\lmd)-\beta_{1\fm}\st}{}$,
    the product inequality for sub-Gaussian variables gives $u_i(\lmd)^2-\cR_j(\lmd)$ sub-exponential tail with scale $\cO\lrp{\tau_j(\lmd)^2}$.
    
    Since $\xt\perp\eps_{t\fm}$ and $\E(\xt)=0$,
    \begin{equation}\label{eq:Rj-decomp-app}
        \cR_j(\lmd)
        =\Var(\eps_{t\fm})+\lrnrm{\Sgot^{1/2}\lrp{\beta_{1\fm}\st-\t\beta_{1\fm}^{\scriptscriptstyle(j)}(\lmd)}}{}^2.
    \end{equation}
    Write $a_j(\lmd):=\lrnrm{\Sgot^{1/2}\lrp{\beta_{1\fm}\st-\t\beta_{1\fm}^{\scriptscriptstyle(j)}(\lmd)}}{}$.
    For the sub-Gaussian scale, $\eps_{t m,i}$ is $\tau_{\eps_\fm}$-sub-Gaussian and ${\bx_{t,i}}'\lrp{\beta_{1\fm}\st-\t\beta_{1\fm}^{\scriptscriptstyle(j)}(\lmd)}$ is $\sot\nrm{\t\beta_{1\fm}^{\scriptscriptstyle(j)}(\lmd)-\beta_{1\fm}\st}{}$-sub-Gaussian, so the sum-of-sub-Gaussians inequality gives
    \begin{equation}\label{eq:tauj-add-app}
        \tau_j(\lmd):=\tau_{\eps_\fm}+\sot\nrm{\t\beta_{1\fm}^{\scriptscriptstyle(j)}(\lmd)-\beta_{1\fm}\st}{}.
    \end{equation}
    Using $\nrm{\t\beta_{1\fm}^{\scriptscriptstyle(j)}(\lmd)-\beta_{1\fm}\st}{}^2\le\lmd_{\min}(\Sgot)^{-1}a_j(\lmd)^2$,
    \begin{equation}\label{eq:tauj-sq-app}
        \tau_j(\lmd)^2
        \le
        2\tau_{\eps_\fm}^2+\frac{2\sot^2}{\lmd_{\min}(\Sgot)}a_j(\lmd)^2
        \lesssim
        \tau_{\eps_\fm}^2+\cR_j(\lmd),
    \end{equation}
    where the last step uses $a_j(\lmd)^2\le\cR_j(\lmd)$ from \eqref{eq:Rj-decomp-app}.
    Recalling that $R_j(\lmd)=\frac{1}{|\cV_j|}\sum_{i\in\cV_j}u_i(\lmd)^2$, Bernstein's inequality conditional on $\cT_j$ gives, for all $t>0$,
    \begin{equation}\label{eq:fold-bern-app}
        \pr\lrp{\vrt{R_j(\lmd)-\cR_j(\lmd)} > t\tau_j(\lmd)^2 \Big| \cT_j}
        \le
        2\exp\lrp{-c|\cV_j|\min\brcs{t^2,t}}.
    \end{equation}
    Taking a union bound over the $\kappa$ folds and the penalty grid $\Lambda$ in Assumption \ref{asm:ntd-cv}, and setting $t=c'\sqrt{\frac{\log{p_1}}{\nt}}$,
    \[
        \pr\lrp{\max_{j\le\kappa}\sup_{\lmd\in\Lambda}\frac{\vrt{R_j(\lmd)-\cR_j(\lmd)}}{\tau_j(\lmd)^2} > c'\sqrt{\frac{\log{p_1}}{\nt}}}
        \le
        2\kappa|\Lambda|\exp\lrp{-c\log{p_1}}.
    \]
    On the complementary event, \eqref{eq:tauj-sq-app} yields, for every $j$ and $\lmd\in\Lambda$,
    \[
        \vrt{R_j(\lmd)-\cR_j(\lmd)}
        \lesssim
        \sqrt{\frac{\log{p_1}}{\nt}}
        \brcs{\tau_{\eps_\fm}^2+\cR_j(\lmd)}.
    \]
    Evaluating at the data-dependent penalty $\lmd_{\rm1SE}$ and using $\cR_j(\lmd_{\rm1SE})\le\bar\cR(\lmd_{\rm1SE})$,
    \[
        \vrt{R_j(\lmd_{\rm1SE})-\cR_j(\lmd_{\rm1SE})}
        \lesssim
        \sqrt{\frac{\log{p_1}}{\nt}}
        \brcs{\tau_{\eps_\fm}^2+\bar\cR(\lmd_{\rm1SE})}~ \text{for every }j.
    \]
    Averaging over $j$ gives
    \[
        \vrt{S_t-\bar\cR(\lmd_{\rm1SE})}
        \lesssim
        \sqrt{\frac{\log{p_1}}{\nt}}
        \brcs{\tau_{\eps_\fm}^2+\bar\cR(\lmd_{\rm1SE})},
    \]
    which is exactly the lemma statement after absorbing the implicit constants into~$c$.
\end{proof}
Next, we denote the localized estimation error boundary by
$$r\sk:=2\sqrt{\lmd_{\max}(\Sgot)\lrp{\cR(\omg_{1\fm}\skt)-\cR(\beta_{1\fm}\st)}}\check\lmd\sk+\lmd_{\max}(\Sgot)\lrp{\check\lmd\sk}^2.$$
\begin{assumption}[Margin scaling]\label{asm:ntd-margin}
    There exists some $c_\eta>8c_{\t\eta}$ such that
    \[
        \eta_\star-\eta_0\ge c_\eta\brcs{\tau_{\eps_\fm}^2+\bar\cR(\lmd_{\rm1SE})}\sqrt{\frac{\log{p_1}}{\nt}}.
    \]
    Furthermore, adversarial sources satisfy $\max_{k\in A_{\eta_0}^c}r\sk\le\frac{\eta_\star-\eta_0}{4}.$
\end{assumption}

\begin{proof}[Proof of Theorem \ref{thm:ntd}]\label{app:proof-thm-ntd}
    Fix a source index $k$ on the event $E_k:=\{k\in A_{\eta_0}^c\}=\{\cR(\omg_{1\fm}\skt)\ge\bar\cR(\lmd_{\rm1SE})+\eta_\star\}$.
    By the selection rule, $k\in\h A$ only if $S_t-\t S\sk\ge\t\eta$, so it suffices to show that $\t S\sk > S_t-\t\eta$ with high probability on $E_k$.

    The HTL transfer coefficient $\check\omg_{1\fm}\sk$ is independent of the target sample, so
    \[
        \E(\t S\sk | \check\omg_{1\fm}\sk)=\cR(\check\omg_{1\fm}\sk)=
        \cR(\beta_{1\fm}\st)+\lrnrm{\Sgot^{1/2}\lrp{\beta_{1\fm}\st-\check\omg_{1\fm}\sk}}{}^2.
    \]
    By the triangle inequality and Lemma \ref{lem:htl-theta-conc},
    \[
        \lrnrm{\Sgot^{1/2}\lrp{\beta_{1\fm}\st-\check\omg_{1\fm}\sk}}{}\ge
        \lrnrm{\Sgot^{1/2}\lrp{\beta_{1\fm}\st-\omg_{1\fm}\skt}}{}
        -\sqrt{\lmd_{\max}(\Sgot)}\check\lmd\sk=:a_k-b_k.
    \]
    Hence
    \begin{align*}
        \cR(\check\omg_{1\fm}\sk)
        &=\cR(\beta_{1\fm}\st)+\lrnrm{\Sgot^{1/2}\lrp{\beta_{1\fm}\st-\check\omg_{1\fm}\sk}}{}^2 \\
        &\ge\cR(\beta_{1\fm}\st)+(a_k-b_k)^2 =\cR\lrp{\omg_{1\fm}\skt}-2a_kb_k+b_k^2.
    \end{align*}
    By the definition of $r\sk$ in Assumption \ref{asm:ntd-margin}, $r\sk=2a_kb_k+b_k^2$, so
    \[
        \cR(\check\omg_{1\fm}\sk)\ge
        \cR\lrp{\omg_{1\fm}\skt}-r\sk.
    \]
    On $E_k$, the separation condition \eqref{eq:sep} gives $\cR(\omg_{1\fm}\skt)\ge\bar\cR(\lmd_{\rm1SE})+\eta_\star$, so
    \begin{equation}\label{eq:oracle-lb-app}
        \cR(\check\omg_{1\fm}\sk)\ge
        \bar\cR(\lmd_{\rm1SE}) + \eta_\star - r\sk,~~r\sk \le \frac{\eta_\star-\eta_0}{4},
    \end{equation}
    where the last inequality is part of Assumption \ref{asm:ntd-margin}.
    Conditional on $\check\omg_{1\fm}\sk$, define the source residuals
    \[
        v_i:=
        Y_{t,i}-{\bx_{t,i}}'\check\omg_{1\fm}\sk=
        \eps_{t m,i}+{\bx_{t,i}}'\lrp{\beta_{1\fm}\st-\check\omg_{1\fm}\sk},\qquad w_i:=v_i^2-\cR(\check\omg_{1\fm}\sk).
    \]
    Conditional on $\check\omg_{1\fm}\sk$, the pairs $\brcs{(\bx_{t,i},Y_{t,i})}_{i=1}^{\nt}$ are i.i.d., so $\brcs{w_i}_{i=1}^{\nt}$ are i.i.d.\ with
    $\E[w_i|\check\omg_{1\fm}\sk]=0$ and $\E[v_i^2|\check\omg_{1\fm}\sk]=\cR(\check\omg_{1\fm}\sk)$ by \eqref{eq:risk-fn}.
    Write
    \[
        \tau_k:=\tau_{\eps_\fm}+\sot\nrm{\check\omg_{1\fm}\sk-\beta_{1\fm}\st}{},
    \]
    using the same sum-of-sub-Gaussians argument as in \eqref{eq:tauj-add-app}, and set
    $a_k':=\lrnrm{\Sgot^{1/2}\lrp{\beta_{1\fm}\st-\check\omg_{1\fm}\sk}}{}$.
    Then $\nrm{\check\omg_{1\fm}\sk-\beta_{1\fm}\st}{}^2\le\lmd_{\min}(\Sgot)^{-1}{a_k'}^2$ gives
    \[
        \tau_k^2
        \lesssim\tau_{\eps_\fm}^2+{a_k'}^2
        =\tau_{\eps_\fm}^2+\cR(\check\omg_{1\fm}\sk)-\cR(\beta_{1\fm}\st)
        \lesssim\tau_{\eps_\fm}^2+\cR(\check\omg_{1\fm}\sk),
    \]
    where the last step uses $\cR(\beta_{1\fm}\st)\ge0$.
    Since $v_i$ is $\tau_k$-sub-Gaussian, $w_i$ is sub-exponential with scale $\cO\lrp{\tau_k^2}$; hence, for all $t>0$,
    \[
        \pr\lrp{\vrt{\t S\sk-\cR(\check\omg_{1\fm}\sk)}>t\tau_k^2 \Big| \check\omg_{1\fm}\sk}
        \le2\exp\brcs{-c\nt\min\lrp{t^2,t}}.
    \]
    Taking $t=c'\sqrt{\log{p_1}/\nt}$ and using $\tau_k^2\lesssim\tau_{\eps_\fm}^2+\cR(\check\omg_{1\fm}\sk)$,
    \[
        \pr\lrp{\t S\sk \le \cR(\check\omg_{1\fm}\sk) - c'\brcs{\tau_{\eps_\fm}^2+\cR(\check\omg_{1\fm}\sk)}\sqrt{\frac{\log{p_1}}{\nt}} \Big| \check\omg_{1\fm}\sk}
        \le2\exp\lrp{-c_2\log{p_1}}.
    \]
    Equivalently, with probability at least $1-2\exp(-c_2\log{p_1})$,
    \begin{equation}\label{eq:bern-tSk-app}
        \t S\sk \ge \cR(\check\omg_{1\fm}\sk) - c'\brcs{\tau_{\eps_\fm}^2+\cR(\check\omg_{1\fm}\sk)}\sqrt{\frac{\log{p_1}}{\nt}}.
    \end{equation}
    For the baseline, Lemma \ref{lem:cv-base} gives
    $S_t\le\bar\cR(\lmd_{\rm1SE})+c\brcs{\tau_{\eps_\fm}^2+\bar\cR(\lmd_{\rm1SE})}\sqrt{\frac{\log{p_1}}{\nt}}$
    with the same probability tail.

    Work on the intersection of $E_k$, the event in \eqref{eq:bern-tSk-app}, and Lemma \ref{lem:cv-base}; by a union bound, this intersection has probability at least $1-c_1\exp(-c_2\log{p_1})$.
    Abbreviate $s:=\sqrt{\log{p_1}/\nt}$, which satisfies $s\le1/(4c')$ once $\nt$ is large (Assumption \ref{asm:ntd-desgn} gives $\log{p_1}\lesssim\nt$).
    Rewriting \eqref{eq:bern-tSk-app} multiplicatively and using $1-c's>0$,
    \[
        \t S\sk\ge\cR(\check\omg_{1\fm}\sk)\lrp{1-c's}-c'\tau_{\eps_\fm}^2 s.
    \]
    The right-hand side is increasing in $\cR(\check\omg_{1\fm}\sk)$, so the oracle lower bound \eqref{eq:oracle-lb-app} can be substituted directly:
    \[
        \t S\sk
        \ge
        \lrp{\bar\cR(\lmd_{\rm1SE})+\eta_\star-r\sk}\lrp{1-c's}-c'\tau_{\eps_\fm}^2 s.
    \]
    Subtracting the baseline bound and collecting terms,
    \[
        S_t-\t S\sk\le-\eta_\star+r\sk+\bxs{(c+c')\brcs{\tau_{\eps_\fm}^2+\bar\cR(\lmd_{\rm1SE})}+c'\lrp{\eta_\star-r\sk}}s,
    \]
    which holds \emph{uniformly} over $k\in A_{\eta_0}^c$.
    By Assumption \ref{asm:ntd-margin}, $r\sk\le(\eta_\star-\eta_0)/4\le\eta_\star/4$; the margin condition $\eta_\star-\eta_0\ge c_\eta\brcs{\tau_{\eps_\fm}^2+\bar\cR(\lmd_{\rm1SE})}s$ with $c_\eta>8(c+c')$ gives $(c+c')\brcs{\tau_{\eps_\fm}^2+\bar\cR(\lmd_{\rm1SE})}s\le\eta_\star/8$; and $s\le1/(4c')$ gives $c'\eta_\star s\le\eta_\star/4$.
    Hence
    \[
        S_t-\t S\sk
        \le-\eta_\star+\frac{\eta_\star}{4}+\frac{\eta_\star}{8}+\frac{\eta_\star}{4}
        =-\frac{3\eta_\star}{8}<0\le\t\eta,
    \]
    so $k\notin\h A$, proving \eqref{eq:fp-bound}.
\end{proof}

\section{Properties of sub-Gaussians}
The following results will be fundamental in our discussion on the asymptotics of the proposed methodology.
\begin{proposition}[\citealp{wainwright2019high}]\label{prp:sg}
    Let $Z$ be a zero-mean $\sgm$-sub-Gaussian random variable, i.e., $Z\sim SG(\sgm)$ for some $\sgm>0$. Then, for any $u \ge 0$, $\pr(|Z| \ge u) \le 2\exp\brcs{-u^2/(2\sgm^2)}$.
\end{proposition}
Note that $Z$ need not be zero-mean, as a sub-Gaussian random variable shifted by a location parameter is also sub-Gaussian.
Also, if $X,Y$ are independent zero-mean $\sgm_1$ and $\sgm_2$-sub-Gaussians, respectively, we have $\E\exp\brcs{u(X+Y)}\le\exp\brcs{u^2\lrp{{\sgm_1}^2+{\sgm_2}^2}/2}$ for any $u\in\R$. Therefore, $X+Y\sim SG(\sgm_1+\sgm_2)$ as ${\sgm_1}^2+{\sgm_2}^2\le\lrp{\sgm_1+\sgm_2}^2$.
If $Z$ is a Gaussian random variable with variance $\sgm^2$, then $Z$ is $\sgm$-sub-Gaussian.
Now, we state the results for sub-Gaussian random vectors and matrices.
\begin{lemma}
    Suppose we have d-dimensional random vectors $\bx,\by$ such that $\bx\sim SG(\sgm_1)$ and $\by|\bx\sim SG(\sgm_2)$ where $\sgm_2$ is independent with $\bx$. Then, $\bx\pm\by\sim SG\lrp{\sgm_1+\sgm_2}$.
\end{lemma}
\proof
Since for any $u\in\R$ and $\bu\in S^{d-1}$
\[\E\exp\brcs{u\lrp{\bx\pm\by}'\bu}=\E\E\bxs{\exp\brcs{u\lrp{\bx\pm\by}'\bu}|\bx}=\E\bxs{e^{u\bx'\bu}\E\lrp{e^{\pm u\by'\bu}|\bx}}\le\E e^{u\bx'\bu}e^{{\sgm_2}^2u^2/2},\]
we have $\bx\pm\by\sim SG\lrp{\sqrt{{\sgm_1}^2+{\sgm_2}^2}}$. Since $\sqrt{{\sgm_1}^2+{\sgm_2}^2}\le{\sgm_1}+{\sgm_2}$, we have the conclusion.
\qed
\vspace{2mm}

There is a close relationship between the covariance and the source parameter of sub-Gaussian random vectors. 
\begin{lemma}
    \label{lem:var-sv}
    If $\bx\in\R^d$ satisfies $\bx\sim SG(\sgm)$ with $\E\lrp{\bx\bx'}=\Sgm_{d\times d}$, then we have $\lmd_{\max}\lrp{\Sgm}=\nrm{\Sgm}{}\le{\sgm}^2$.
\end{lemma}
\proof
 By definition, the MGF of $\bx$ satisfies $\operatorname{M}_{\bu'\bx}(t)\le\exp\lrp{t^2{\sgm}^2/2}$ for all $t\in\R$ and $\bu\in S^{d-1}$, i.e., its moments are bounded above by corresponding moments of $N(0,\sgm^2)$ distribution.
 Hence we have $\sup_\bu\Var(\bu'\bx)=\sup_\bu\bu'\Sgm\bu\le{\sgm}^2$.
 Noting that $\Sgm$ is a symmetric positive semi-definite matrix, we complete the proof.
\qed
\vspace{2mm}
\begin{lemma}\label{lem:sg-add}
    Suppose we have $n\times d$ sub-Gaussian ensembles $\bX\sim SG(\sgm_1)$ and $\bY$ such that $\bY|\bX\sim SG(\sgm_2)$ with $\sgm_2$ independent of $\bX$. Then, $\bX+\bY\sim SG(\sgm_1+\sgm_2)$.
\end{lemma}
\proof
Pick the first row ${\bx_1}'+{\by_1}'$, for example.
Since for any $u\in\R$ and $\bv\in S^{d-1}$
\[\E\exp\brcs{u\lrp{\bx_1+\by_1}'\bv}=\E\bxs{e^{u{\bx_1}'\bv}\E\lrp{e^{u{\by_1}'\bv}|\bX}}\le\E e^{u{\bx_1}'\bv}e^{{\sgm_2}^2u^2/2},\]
we have $\bx_1+\by_1\sim SG\lrp{\sgm_1+\sgm_2}$. This holds for all independent rows.
\qed

\section{Additional Simulation Results}

\begin{figure}[!htbp]
    \centering
      \includegraphics[width=.49\textwidth]{plots/ee_nt_linear.png}
      \includegraphics[width=.49\textwidth]{plots/ee_K_linear.png}
      \includegraphics[width=.49\textwidth]{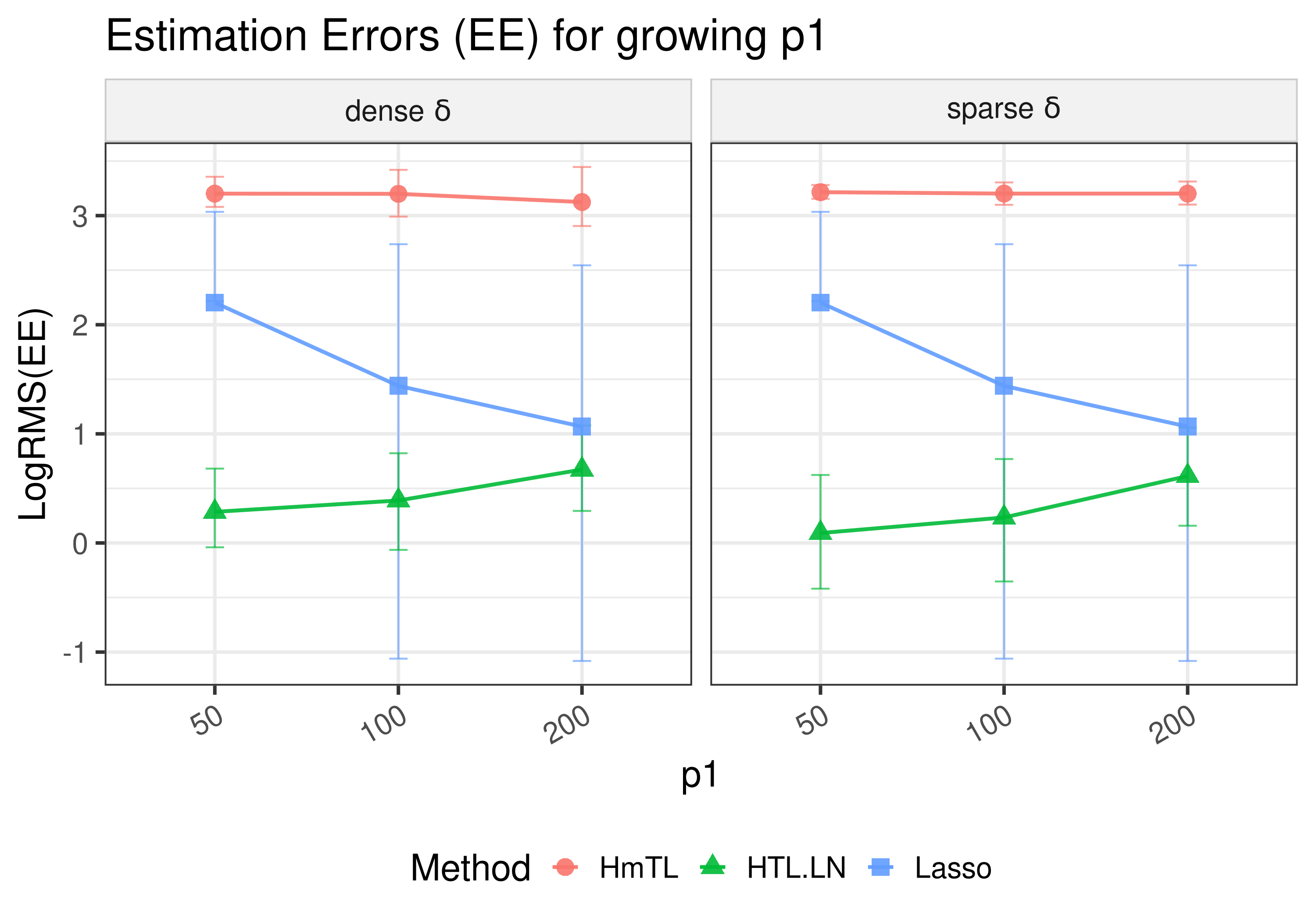}
      \includegraphics[width=.49\textwidth]{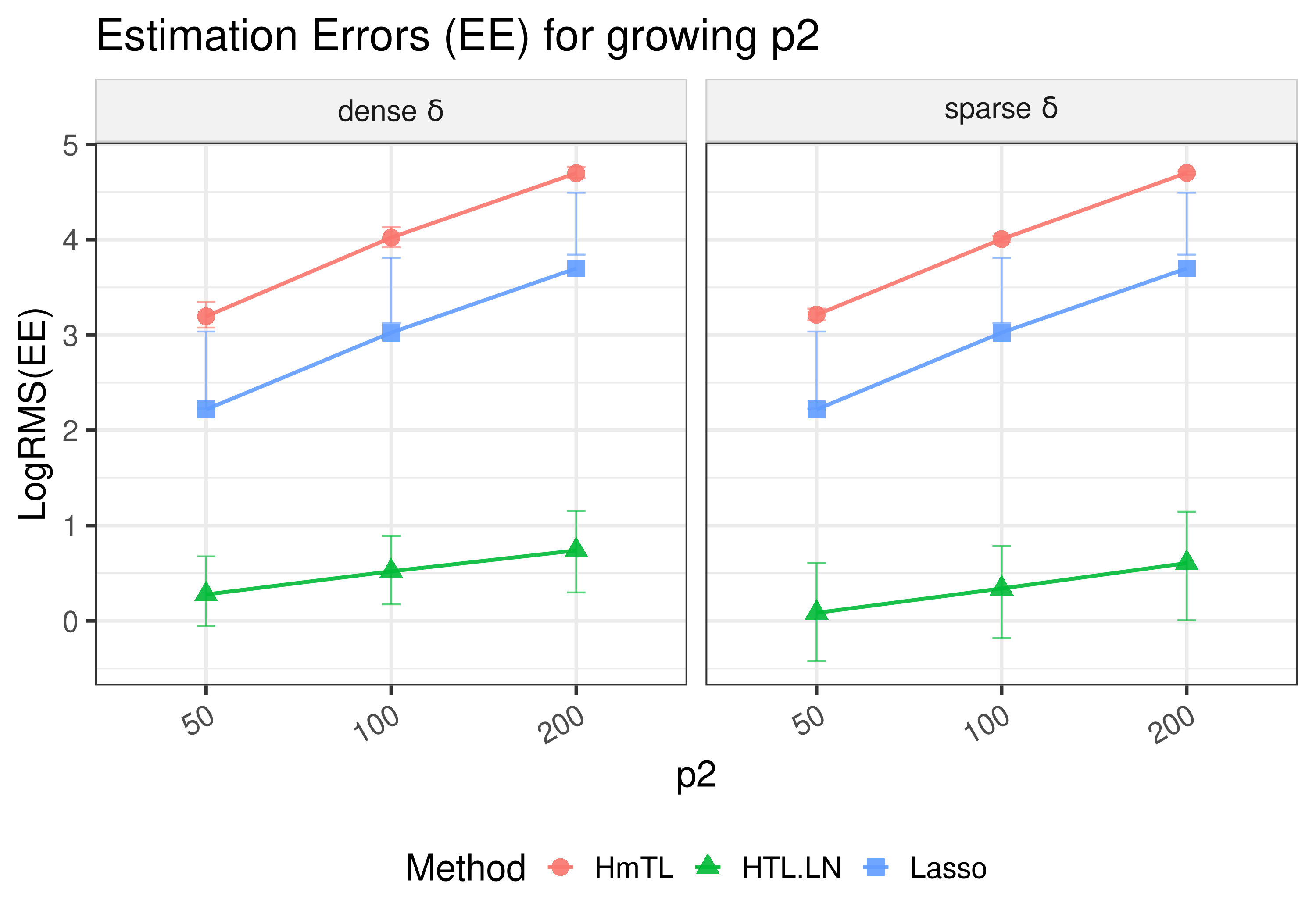}
    \caption{Estimation error of considered methodologies with sparse $\dst$ and non-sparse $\dst$, with increasing $n_t, K,p_1,p_2$ respectively.}
    \label{fig:htl-lfm-sim3}
\end{figure}

\begin{figure}[!h]
    \centering
      \includegraphics[width=.49\textwidth]{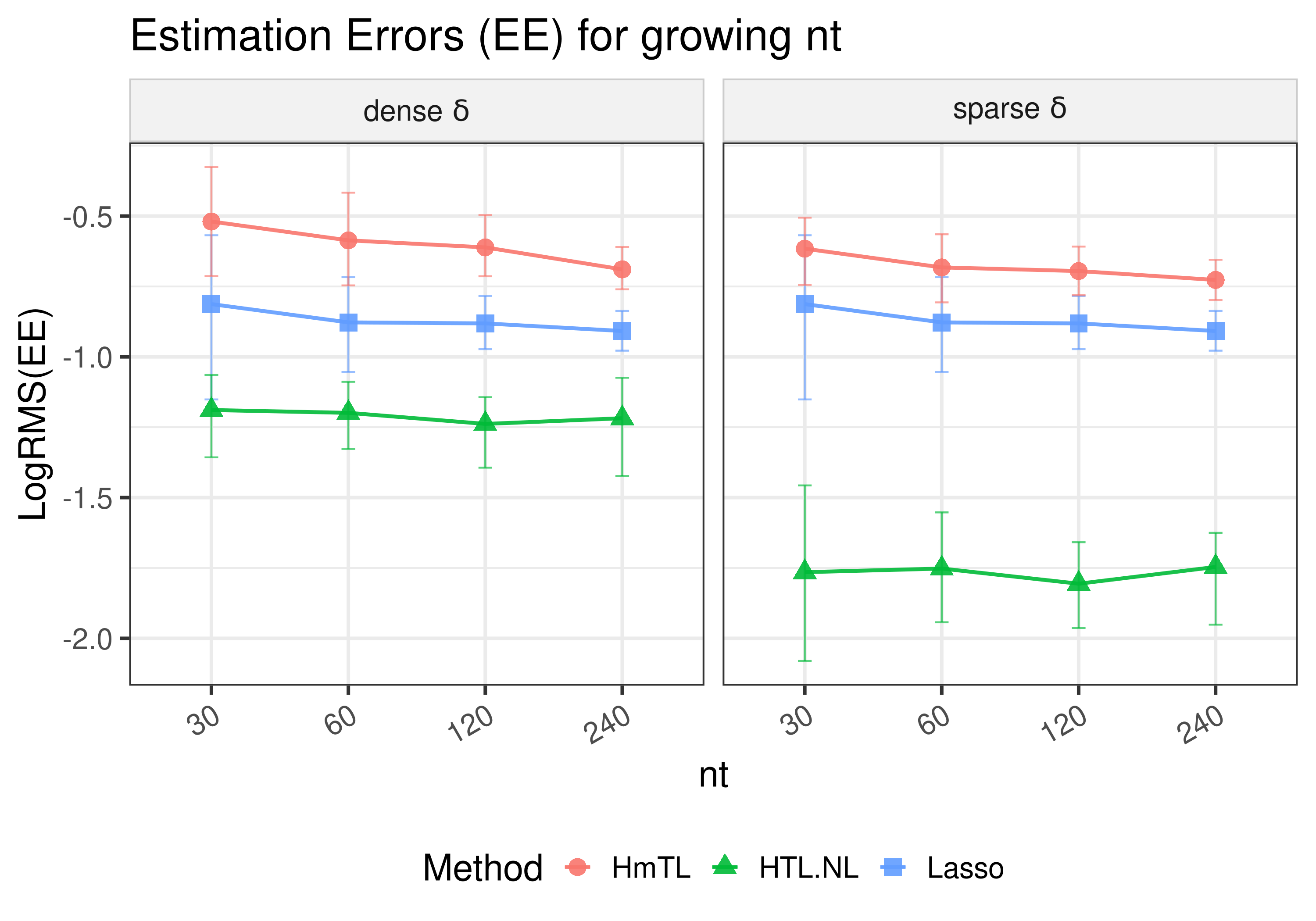}
      \includegraphics[width=.49\textwidth]{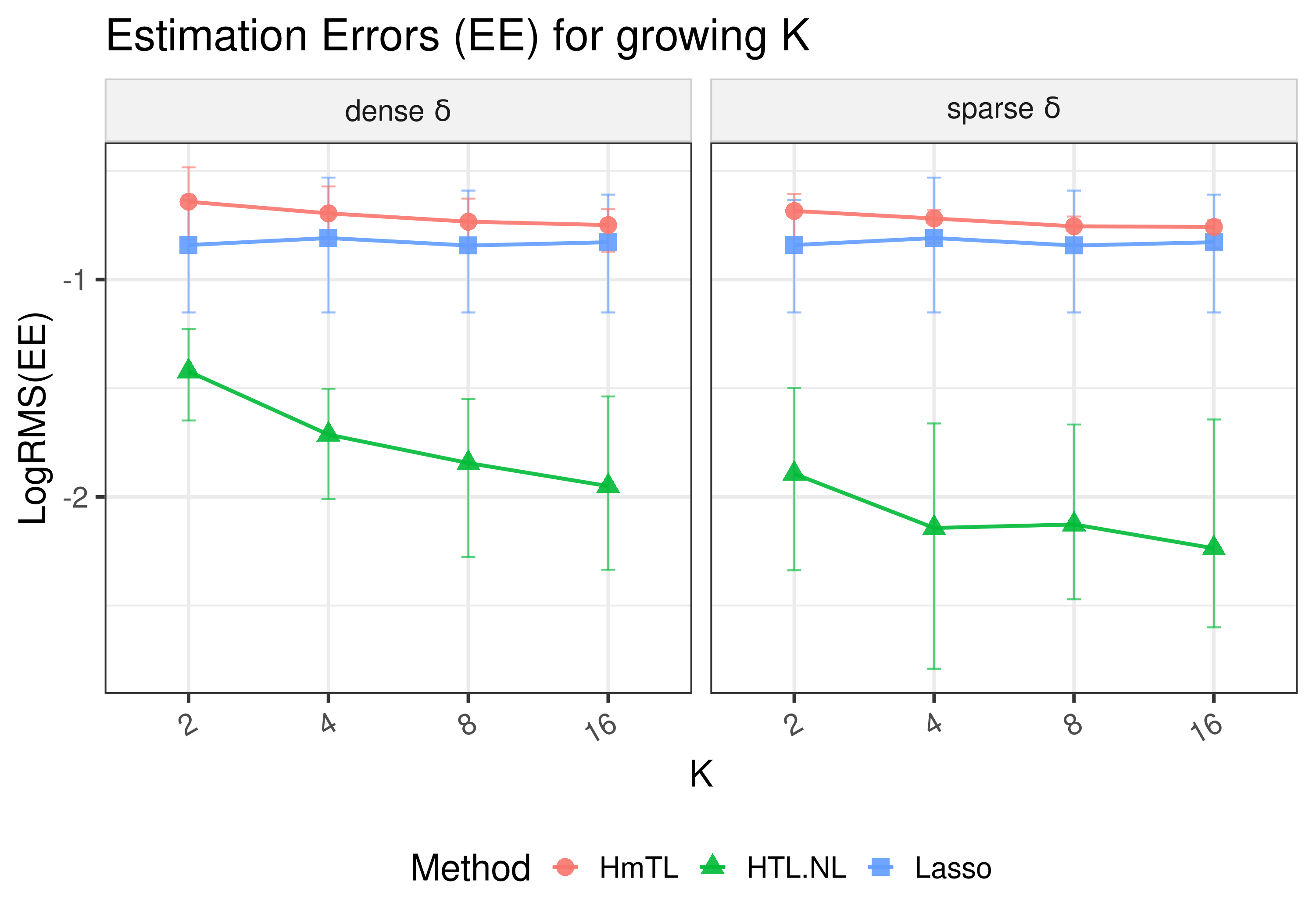}
      \includegraphics[width=.49\textwidth]{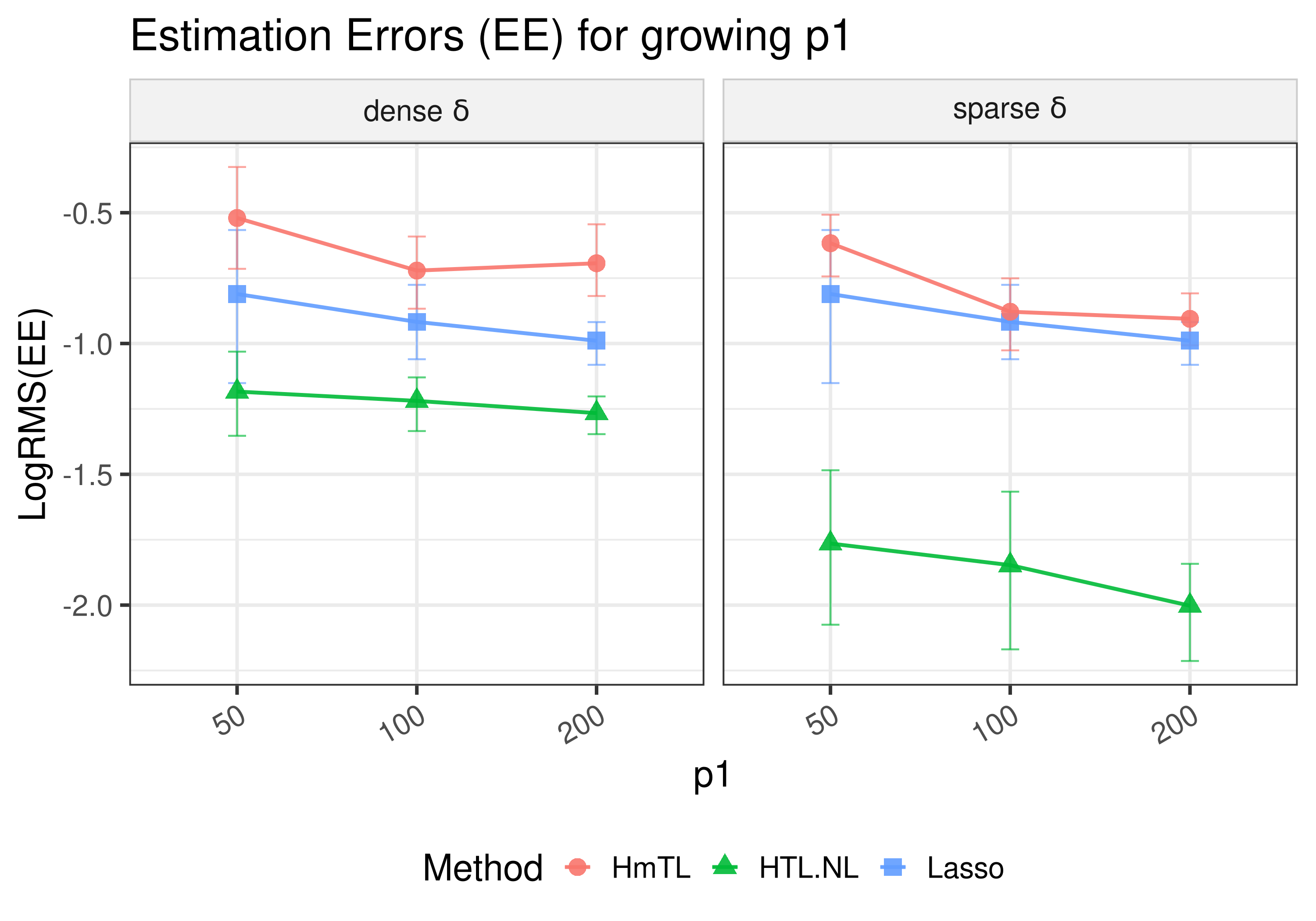}
      \includegraphics[width=.49\textwidth]{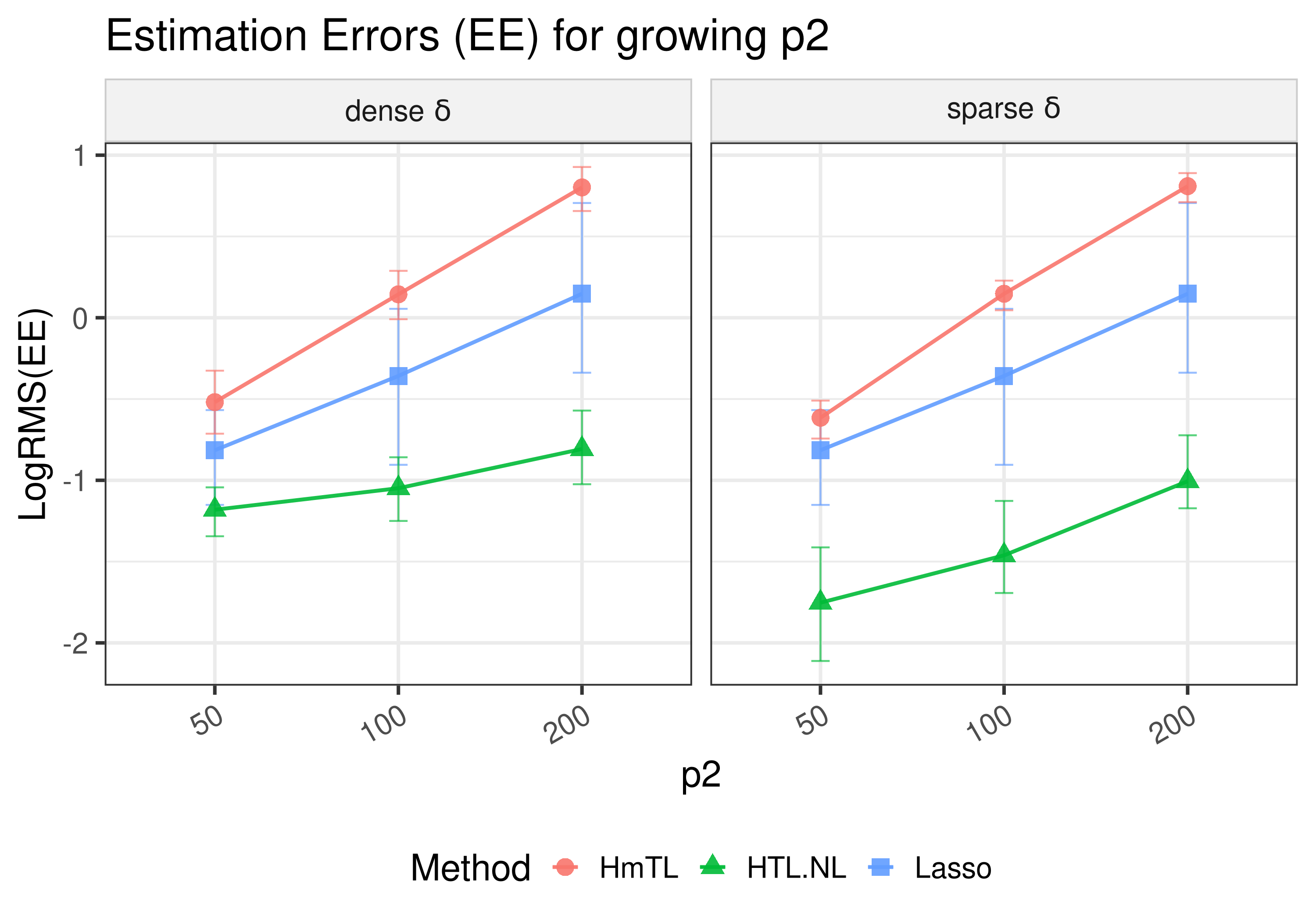}
    \caption{Estimation errors of considered methodologies for estimating $\beta_1\st$ in the non-linear feature map case with increasing $n_t, K,p_1,p_2$ respectively.}
    \label{fig:htl-sim3}
\end{figure}

\subsection{Algorithm for SAND method}
When we have $K$ source datasets, Algorithm \ref{alg:sand} can be applied to construct $\h A$.
\begin{algorithm}[!htbp]
    \caption{Signal-Based Adaptive Negative Transfer Defense (SAND)}\label{alg:sand}
    \begin{algorithmic}[1]
        \small
        \Require $(\Xt, \by_t)$, $\{\h{\omg}\sk\}_{k=1}^K$ and feature mappings $\{\h{\bH}_s\sk\}_{k=1}^K$ from $K$ source domains.
        \Ensure Estimated informative source set $\h{A}$.
        \State Partition target data into $\kappa$ folds $\{\mathcal{V}_j\}_{j=1}^\kappa$; let training sets be $\cT_j := [\nt] \setminus \mathcal{V}_j$.
        \State For each candidate $\lmd > 0$, let
        $\t\beta_{1\fm}^{\scriptscriptstyle(j)}(\lmd) \gets {\arg\min}_{\beta_1} \left\{ \frac{1}{2|\cT_j|} \nrm{\by_{\cT_j} - \bX_{\cT_j} \beta_1}{2}^2 + \lmd\nrm{\beta_1}{1} \right\}$
        \State Let $R_j(\lmd) := \frac{1}{|\mathcal{V}_j|} \nrm{\by_{\mathcal{V}_j} - \bX_{\mathcal{V}_j} \t\beta_{1\fm}^{\scriptscriptstyle(j)}(\lmd)}{2}^2$. Compute CV error and standard error:
        $$\bar R(\lmd) \gets \frac{1}{\kappa}\sum_{j=1}^\kappa R_j(\lmd), \quad \operatorname{SE}(\lmd) \gets \frac{1}{\sqrt{\kappa}} \sqrt{ \frac{1}{\kappa-1} \sum_{j=1}^\kappa \left( R_j(\lmd) - \bar R(\lmd) \right)^2 }$$
        \State Let $\lmd_{\min} \gets {\arg\min}_{\lmd > 0} \bar R(\lmd)$, $\lmd_{\rm1SE} \gets \max \left\{ \lmd : \bar R(\lmd) \le \bar R(\lmd_{\min}) + \operatorname{SE}(\lmd_{\min}) \right\}$
        and $S_t \gets \bar R(\lmd_{\rm1SE})$
        \State Initialize $\h{A} \gets \emptyset$
        \For{each source $k\in[K]$}
            \State $\h{\by}_t\sk \gets \Xt\h{\omg}_1\sk+\h\bH_s\sk(\Xt)\h{\omg}_2\sk$,
            \quad $\t S\sk \gets \frac{1}{\nt} \nrm{\by_t - \h{\by}_t\sk}{2}^2$
            \If{$S_t - \t S\sk \ge\h\eta$}
                \State $\h{A} \gets \h{A} \cup \{k\}$ 
            \EndIf
        \EndFor
        \nobreak
        \State \Return $\h{A}$
    \end{algorithmic}
\end{algorithm}

\subsection{Simulations for Negative Transfer Defense}\label{sec:sim-sand}
For the simulations of the SAND algorithm, we compare the predictions of four methods. \texttt{AllHTL} is a naive HTL method that treats all available source datasets as informative to transfer. \texttt{OracleHTL} is the HTL with a priori knowledge of the true $A$ set, transferring $k\in A$ only. This serves as a benchmark we aim to recover. \texttt{SAND\_HTL} implements the SAND algorithm to estimate $\h A$ and transfer $\h\omg\sk,\h{\Pp\sk}$ for $k\in\h A$ only. \texttt{TLGLM\_org} uses the \texttt{glmtrans} \citep{tian2022transfer} package in R to estimate $\h A$, and then conduct the HmTL with it.

\begin{figure}[h]
    \centering
    \includegraphics[width=.495\textwidth]{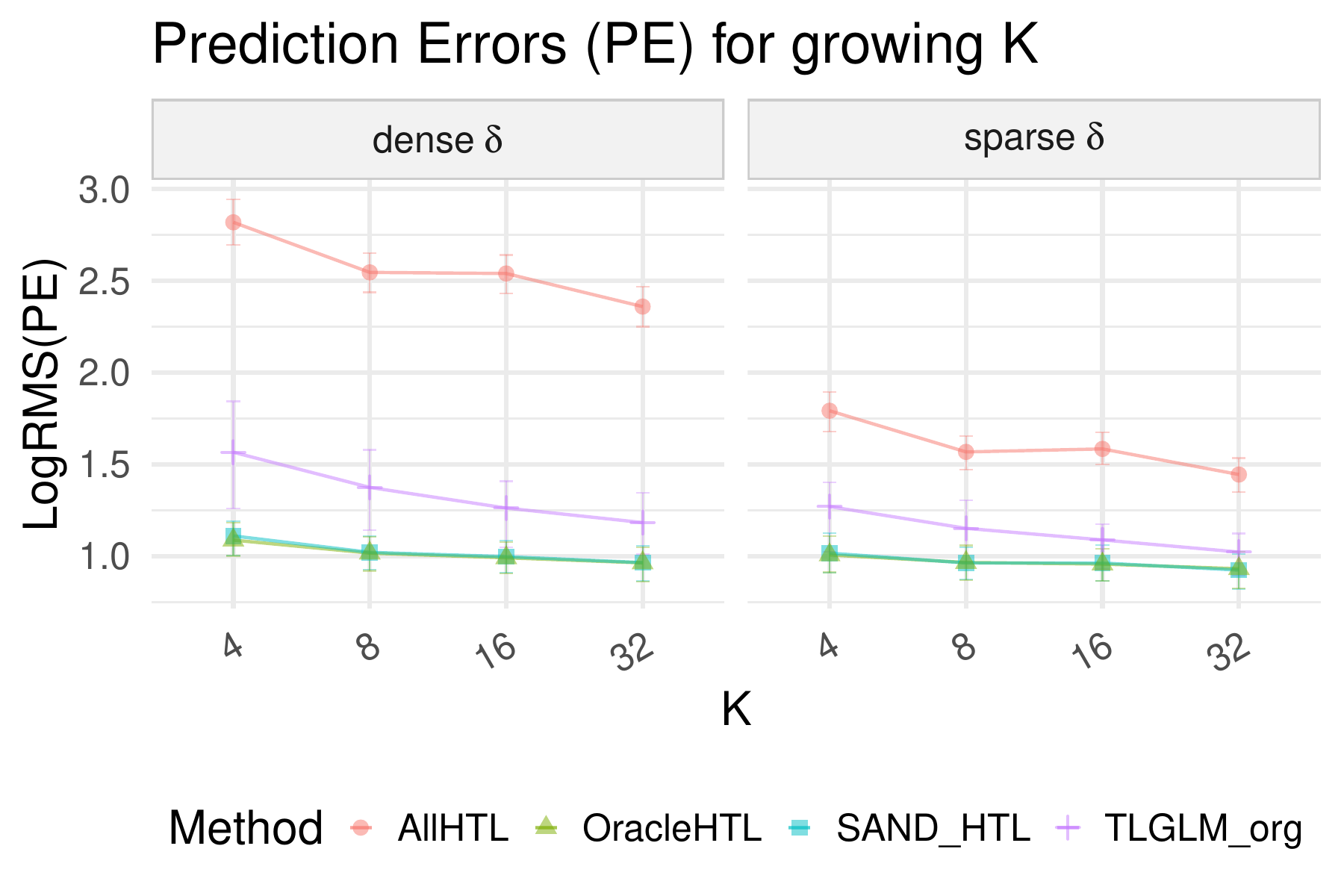}
    \includegraphics[width=.495\textwidth]{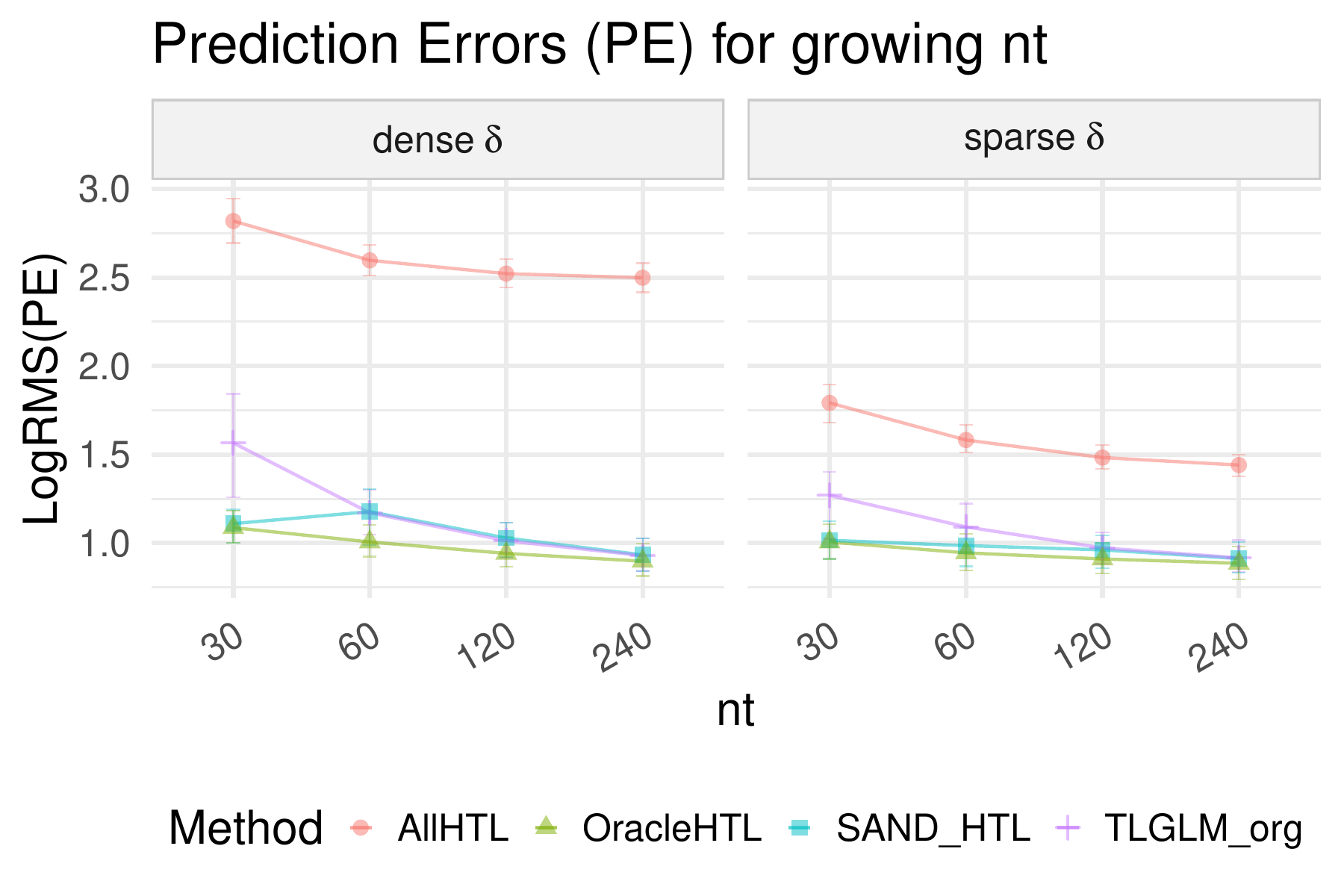}
    \caption{LogRMS(PE) on a held-out test set, faceted by contrast sparsity. \textbf{Left:} $K\in\{4,8,16,32\}$ with $K_{\rm adv}=\lfloor K^{2/3}\rfloor$. \textbf{Right:} $n_t\in\{30,60,120,240\}$ at $K=4$.}
    \label{fig:sand_pred_err}
\end{figure}

\begin{figure}[h]
    \centering
    \includegraphics[width=.495\textwidth]{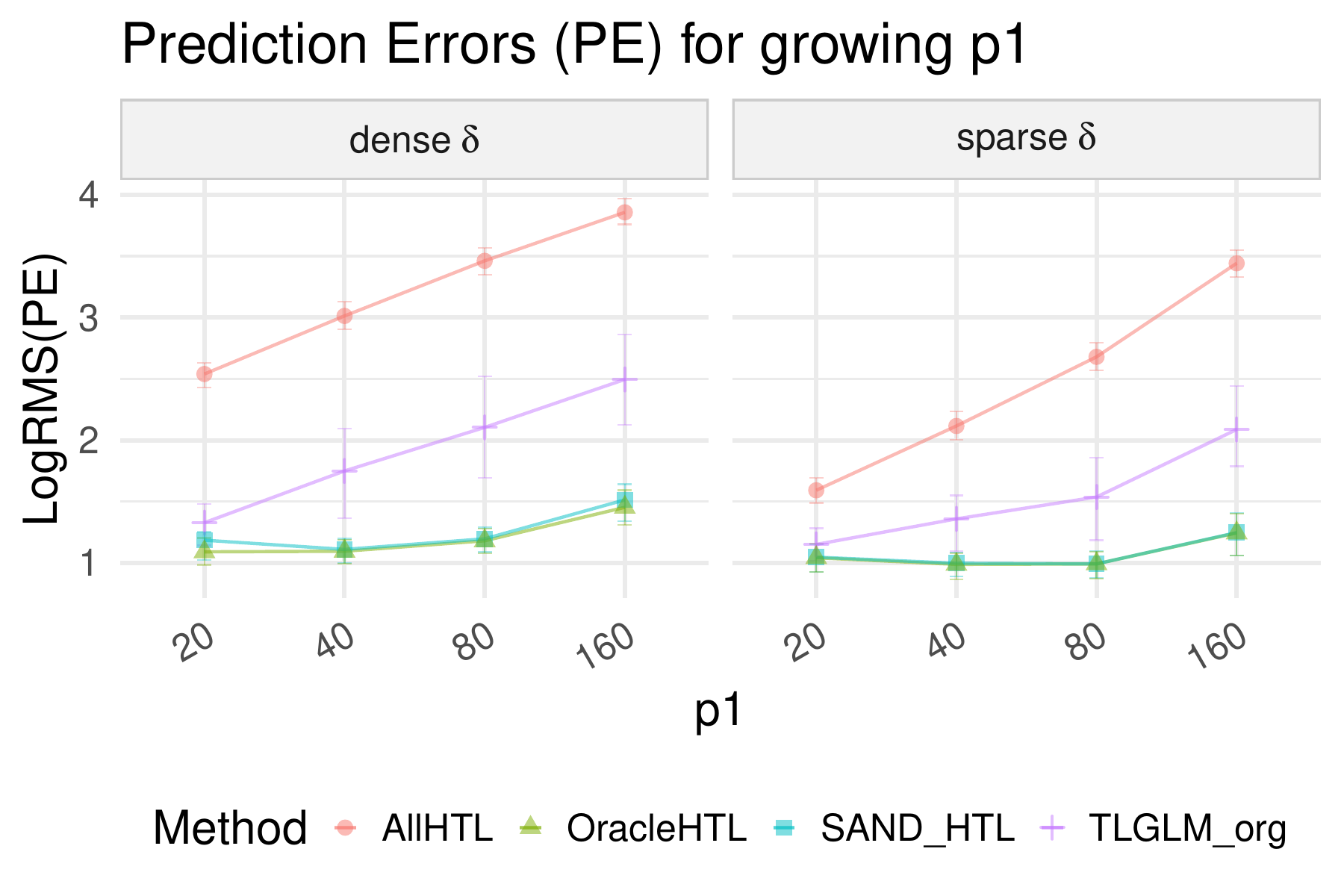}
    \includegraphics[width=.495\textwidth]{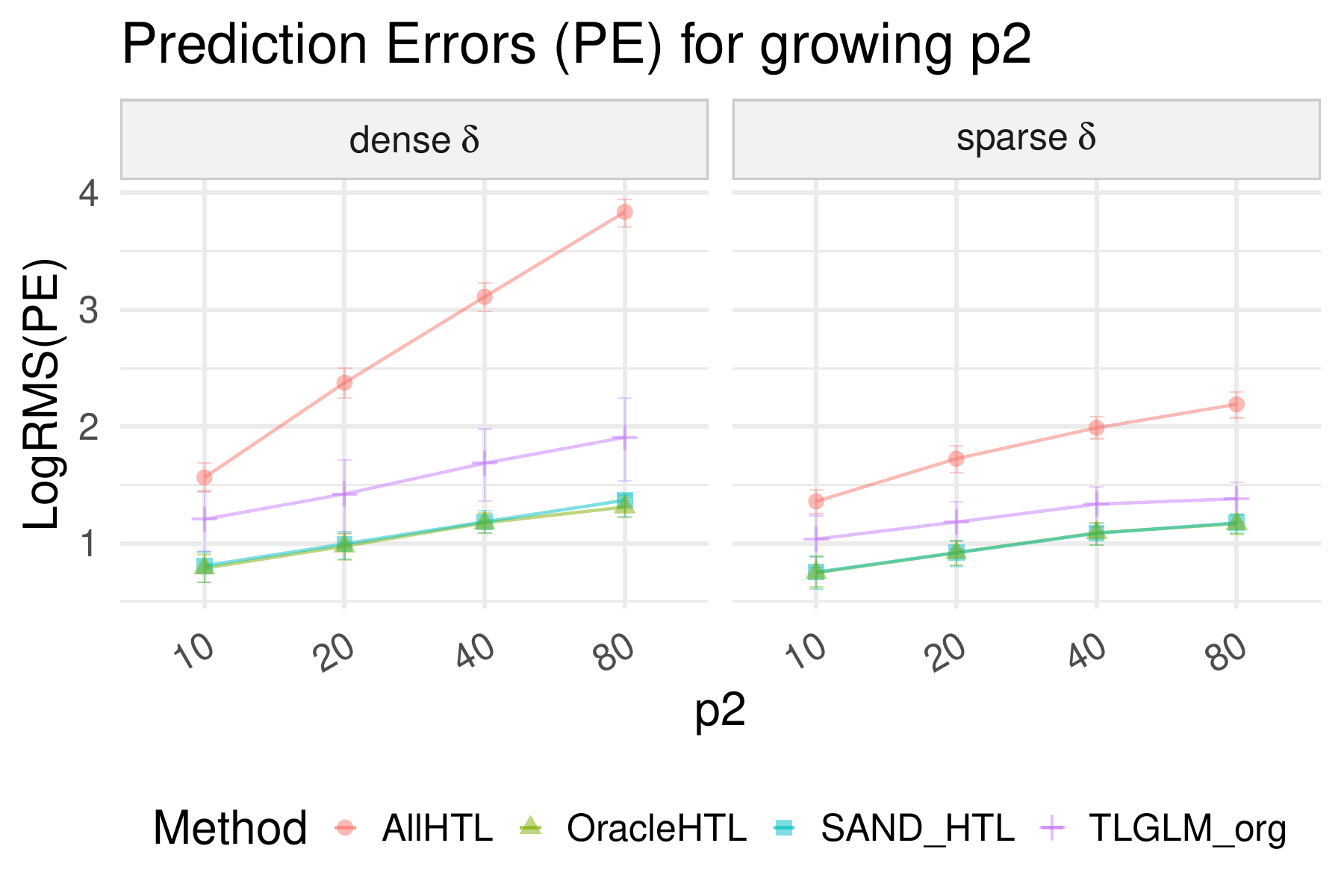}
    \caption{LogRMS(PE) as ${p_1}\in\{20,40,80,160\}$ (left) and ${p_2}\in\{10,20,40,80\}$ (right) vary, with other parameters at baseline. Methods and faceting as in Figure~\ref{fig:sand_pred_err}.}
    \label{fig:sand_pred_err2}
\end{figure}

Figures~\ref{fig:sand_pred_err} and~\ref{fig:sand_pred_err2} plot $\log$ RMS prediction error (LogRMS(PE)) on a held-out test set.
In Figure \ref{fig:sand_pred_err}, as the number of sources $K$ increases, we vary the number of adversarial sources as $\lfloor K^{2/3}\rfloor$ and the remaining $K-\lfloor K^{2/3}\rfloor$ sources are informative ones. Our proposed automatic source selection method
\texttt{SAND\_HTL} tracks \texttt{OracleHTL} closely across all points. \texttt{AllHTL} is uniformly worst and degrades as $K$ or ${p_1}$ grows, reflecting severe negative transfer from retained adversarial sources.
\texttt{TLGLM\_org} lies between \texttt{AllHTL} and \texttt{SAND\_HTL} and improves with larger $n_t$, but does not match \texttt{SAND\_HTL} or \texttt{OracleHTL}.
In right side of Figure \ref{fig:sand_pred_err}, increasing $n_t$ helps all methods except \texttt{AllHTL}, which remains dominated by negative transfer, while \texttt{SAND\_HTL} method preserves the oracle advantage throughout.
Varying ${p_1}$ and ${p_2}$ in Figure \ref{fig:sand_pred_err2}, shows the same behavior: \texttt{SAND\_HTL} is stable and near-oracle, while \texttt{AllHTL} and \texttt{TLGLM\_org} continue to experience larger errors compared to \texttt{SAND\_HTL} as model dimensions grow.

\end{document}